\documentclass{z-image}
\usepackage[utf8]{inputenc} 
\usepackage[T1]{fontenc}    
\usepackage{hyperref}       
\usepackage{url}            
\usepackage{booktabs}       
\usepackage{amsfonts}       
\usepackage{nicefrac}       
\usepackage{microtype}      
\usepackage{float}
\usepackage{mwe}
\usepackage{graphicx}
\usepackage{amssymb}
\usepackage{adjustbox}
\usepackage{colortbl}
\usepackage{array}
\usepackage{multirow}
\usepackage{makecell}
\usepackage{bbm}
\usepackage{collcell,xfp}
\usepackage{pgf}
\usepackage{tikz}
\usepackage[most]{tcolorbox}
\usepackage{csquotes}
\usepackage[noorphans,vskip=1em,leftmargin=1em]{quoting}
\usepackage{pifont} 
\usepackage{enumitem} 
\usepackage{tcolorbox} 
\usepackage{forest}
\usepackage{wrapfig}
\usepackage{caption}
\usepackage{subcaption}
\usepackage{longtable}
\usepackage{algpseudocode}
\usepackage{amsmath}
\usepackage[capitalize]{cleveref}
\usepackage[ruled,vlined,linesnumbered]{algorithm2e} 
\usepackage[percent]{overpic}
\usepackage{xspace}
\usepackage[normalem]{ulem}  
\usepackage{tocloft}  
\usepackage{natbib}  
\usepackage{listings}

\usepackage{CJKutf8}   
\newcommand{\zh}[1]{
  \begin{CJK}{UTF8}{gbsn}#1\end{CJK}}

\usepackage{twemojis}
\usepackage{tikz-dependency}
\usepackage{graphicx}
\usepackage{tikz}
\usepackage{tikz-qtree}
\usepackage{caption}
\usetikzlibrary{arrows.meta}
\definecolor{clearpurple}{RGB}{138, 140, 191}
\definecolor{clearyellow}{HTML}{f2d3bf}
\definecolor{skyblue}{HTML}{a8d8e2}
\definecolor{darkblue}{HTML}{19183B}
\definecolor{tp1}{RGB}{253, 207, 158}

\usepackage{fontawesome}
\newcommand{\huggingface}{\raisebox{-1.5pt}{\includegraphics[height=1.05em]{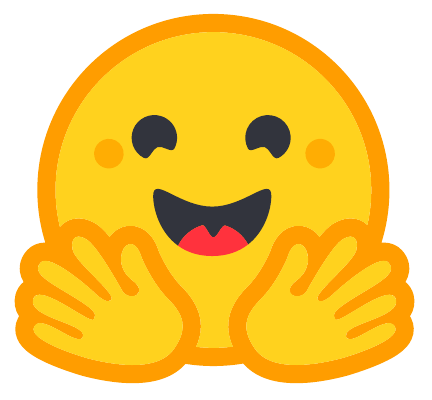}}\xspace}
\newcommand{\modelscope}{\raisebox{-1.5pt}{\includegraphics[height=1.05em]{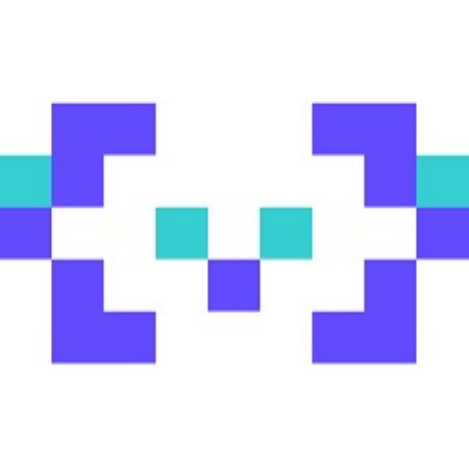}}\xspace}
\newcommand{\github}{\raisebox{-1.5pt}{\includegraphics[height=1.05em]{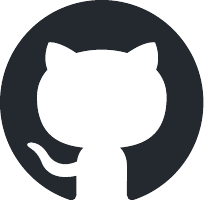}}\xspace}

\definecolor{scholarblue}{rgb}{0.21,0.49,0.74}
\definecolor{bluelink}{RGB}{0,113,188}
\definecolor{greenlink}{RGB}{0,188,113}
\hypersetup{
    colorlinks=true,%
    citecolor=scholarblue,%
    filecolor=red,%
    linkcolor=red!93!black,%
    urlcolor=bluelink
}

\usepackage{titlesec}
\titlespacing*{\paragraph}
    {0pt}     
    {0.25em}  
    {1em}  

\definecolor{navyblue}{HTML}{0071BC}
\iftrue   
\newcommand{\displaytodo}[1]{#1}
\else
\newcommand{\displaytodo}[1]{}
\fi

\def\eg{\emph{e.g.,~}}
\def\ie{\emph{i.e.,~}}

\def\eg{\emph{e.g.}} 
\def\ie{\emph{i.e.}}

\definecolor{blindcolor}{HTML}{AB2AC6}    
\definecolor{chancecolor}{HTML}{F59E0B}   
\definecolor{singlecolor}{HTML}{06B6D4}   
\definecolor{multiplecolor}{HTML}{2563EB} 
\definecolor{captioncolor}{HTML}{22C55E}  

 \newcommand{\culine}[2]{%
    \def\temp@uline{\bgroup\markoverwith
        {\textcolor{#1}{\rule[-0.5ex]{2pt}{1pt}}}\ULon}%
    \temp@uline{#2}%
}
 \newcommand{\cthickuline}[3][0.8pt]{%
    \def\temp@uline{\bgroup\markoverwith
        {\textcolor{#2}{\rule[-0.5ex]{2pt}{#1}}}\ULon}%
    \temp@uline{#3}%
}

\title{\center{Z-Image: An Efficient Image Generation Foundation Model with Single-Stream Diffusion Transformer}}

\renewcommand\Affilfont{\normalfont\fontsize{11}{15}\selectfont\centering}
\author{ 
    Z-Image Team, Alibaba Group
}

\begin{abstract}
The landscape of high-performance image generation models is currently dominated by proprietary systems, such as Nano Banana Pro~\cite{nanopro} and Seedream 4.0~\cite{seedream2025seedream}. Leading open-source alternatives, including Qwen-Image~\cite{qwenimage}, Hunyuan-Image-3.0~\cite{cao2025hunyuanimage} and FLUX.2~\citep{flux-2-2025}, are characterized by massive parameter counts (20B to 80B), making them impractical for inference, and fine-tuning on consumer-grade hardware. To address this gap, we propose \textbf{Z-Image}, an efficient \textit{6B-parameter} foundation generative model built upon a Scalable Single-Stream Diffusion Transformer (S3-DiT) architecture that \textit{challenges the ``scale-at-all-costs" paradigm}.
By systematically optimizing the entire model lifecycle -- from a curated data infrastructure to a streamlined training curriculum -- we complete the full training workflow in \textit{just 314K H800 GPU hours (approx. \$630K)}.
Our few-step distillation scheme with reward post-training further yields \textbf{Z-Image-Turbo}, offering both \textit{sub-second inference latency} on an enterprise-grade H800 GPU and \textit{compatibility with consumer-grade hardware (<16GB VRAM)}. Additionally, our omni-pre-training paradigm also enables efficient training of \textbf{Z-Image-Edit}, an editing model with impressive instruction-following capabilities.
Both qualitative and quantitative experiments demonstrate that our model achieves performance comparable to or surpassing that of leading competitors across various dimensions. 
Most notably, Z-Image exhibits \textit{exceptional capabilities in photorealistic image generation and bilingual text rendering}, delivering results that rival top-tier commercial models, thereby demonstrating that \textit{state-of-the-art results are achievable with significantly reduced computational overhead}.
We publicly release our code, weights, and online demo to foster the development of accessible, budget-friendly, yet state-of-the-art generative models.
\end{abstract}

\begin{document}
\setlength{\parindent}{0pt}

\begin{tikzpicture}[remember picture,overlay]
    
    \node[anchor=north west, xshift=2.5cm, yshift=-1.65cm] at (current page.north west) {
         {\twemoji[width=1.1em]{zap}}
    };
    \node[anchor=north west, xshift=2.9cm, yshift=-1.7cm] at (current page.north west) {
         \large\textbf{\textit{-Image}}
    };
    
    \node[anchor=north east, xshift=-2.5cm, yshift=-1.7cm] at (current page.north east) {
        \large\textcolor{gray}{November 27, 2025}
    };
\end{tikzpicture}

\maketitle



\begin{center}
    \renewcommand{\arraystretch}{1.5}
    \vspace{1em}
    \begin{tabular}{rll}
        \github{} & \textbf{GitHub} & \url{https://github.com/Tongyi-MAI/Z-Image} \\
        \modelscope{} & \textbf{ModelScope Model} & \url{https://modelscope.cn/models/Tongyi-MAI/Z-Image-Turbo} \\
        \huggingface{} & \textbf{HuggingFace Model} & \url{https://huggingface.co/Tongyi-MAI/Z-Image-Turbo} \\
        \modelscope{} & \textbf{ModelScope Demo} & \href{https://www.modelscope.cn/aigc/imageGeneration?tab=advanced&versionId=469191&modelType=Checkpoint&sdVersion=Z_IMAGE_TURBO&modelUrl=modelscope%3A%2F%2FTongyi-MAI%2FZ-Image-Turbo%3Frevision%3Dmaster}{Online Demo (ModelScope)} \\
        \huggingface{} & \textbf{HuggingFace Demo} & \href{https://huggingface.co/spaces/Tongyi-MAI/Z-Image-Turbo}{Online Demo (HuggingFace)} \\
        \twemoji[width=1em]{1f5bc} & \textbf{Image Gallery} & \href{https://modelscope.cn/studios/Tongyi-MAI/Z-Image-Gallery/summary}{Online Gallery} \quad \href{https://www.modelscope.cn/models/Tongyi-MAI/Z-Image-Turbo/file/view/master/assets%2FZ-Image-Gallery.pdf}{Offline Gallery} \\
    \end{tabular}
\end{center}




\newpage
\begin{figure}[H]
  \centering
  \includegraphics[width=\textwidth]{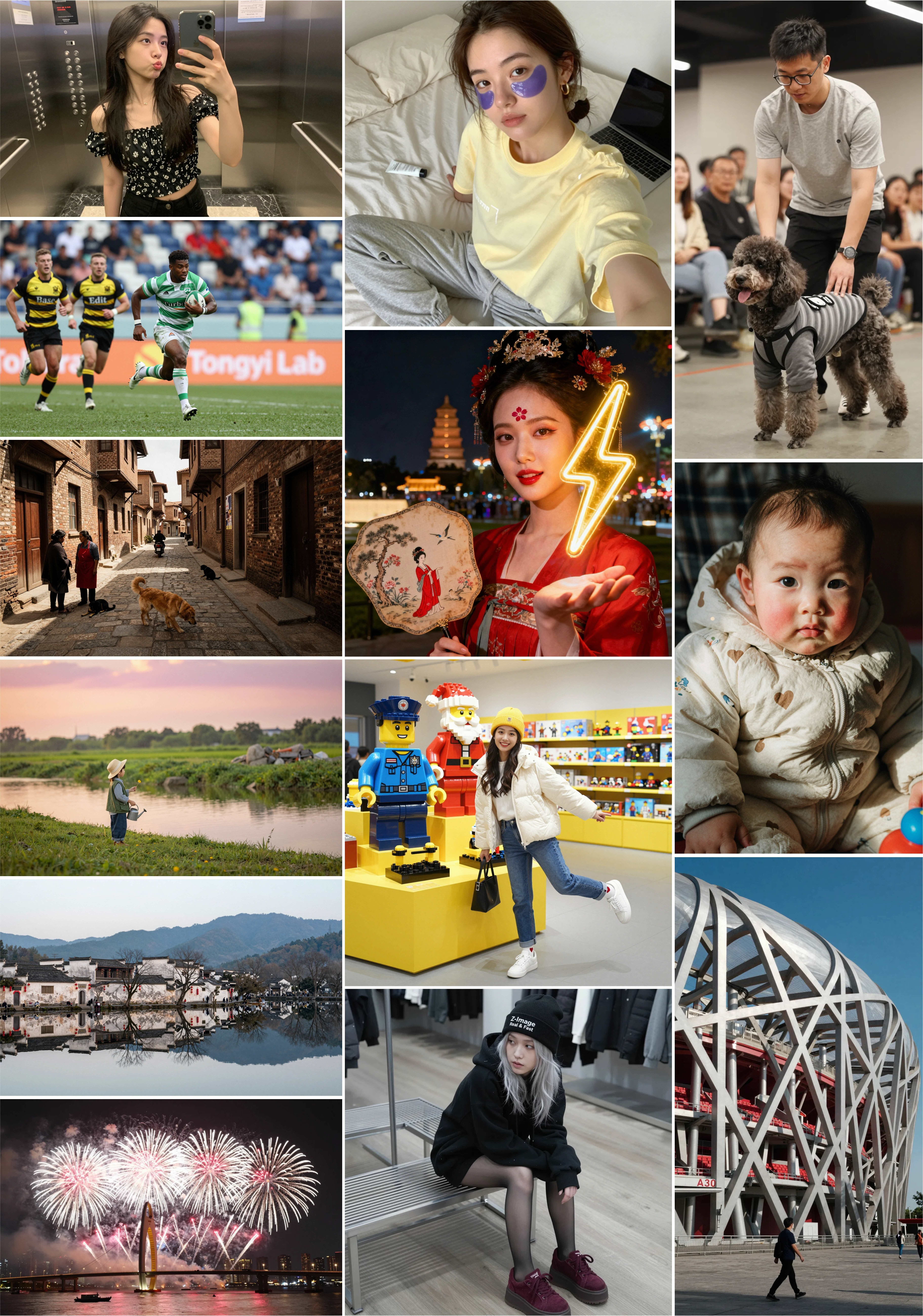}
  \caption{Showcases of Z-Image-Turbo in photo-realistic image generation. All related prompts can be found in Appendix~\ref{sec:fig_1}.}
  \label{fig:showcase_realistic}
\end{figure}
\newpage

\newpage
\begin{figure}[H]
  \centering
  \includegraphics[width=\textwidth]{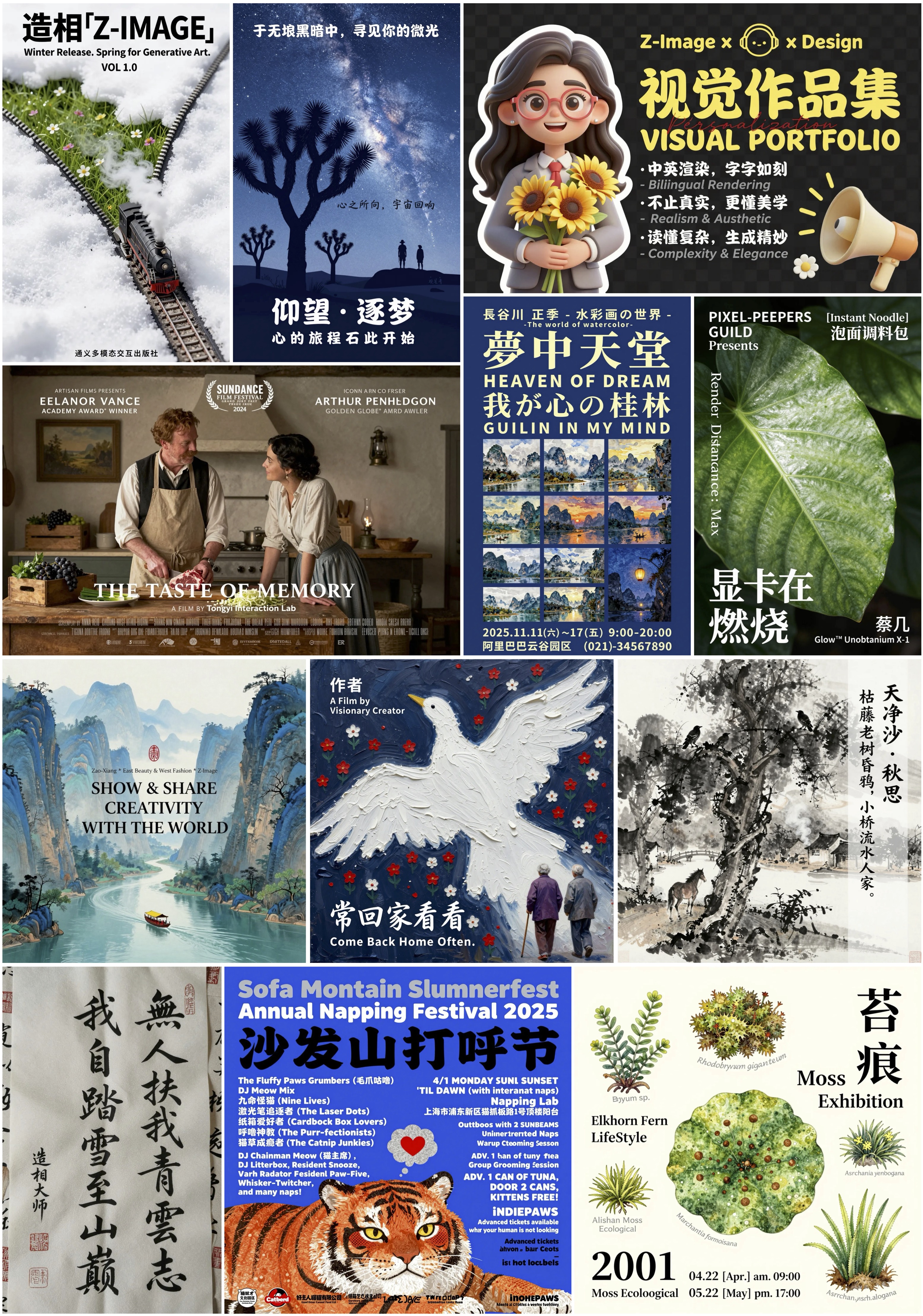}
  \caption{Showcases of Z-Image-Turbo in bilingual text-rendering. All related prompts can be found in Appendix~\ref{sec:fig_2}.}
  \label{fig:showcase_text}
\end{figure}
\newpage


\newpage
\begin{figure}[H]
  \centering
  \includegraphics[width=\textwidth]{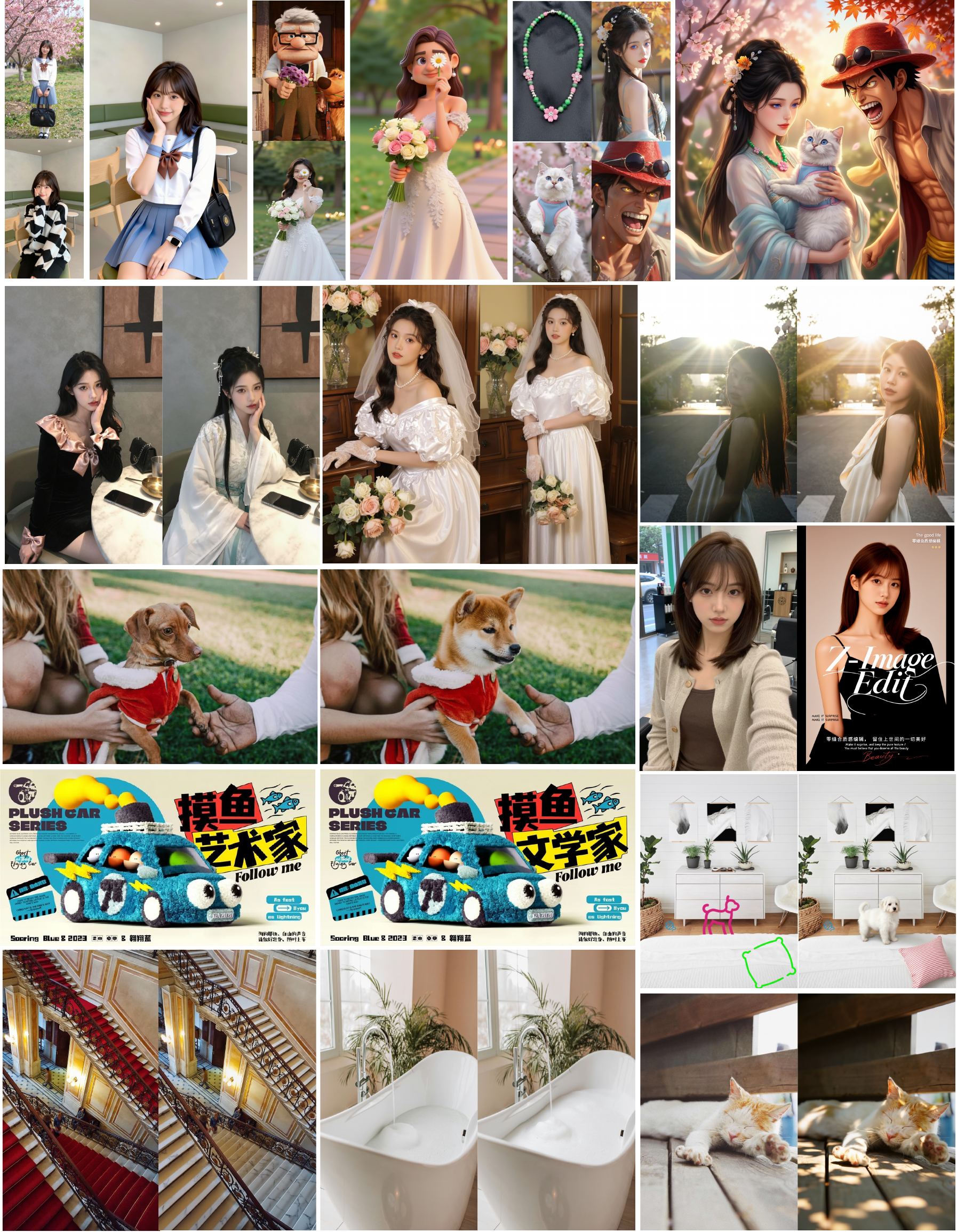}
  \caption{Showcases of Z-Image-Edit in various image-to-image tasks. 
  Each arrow represents an edit from the input to output images. All related prompts can be found in Appendix~\ref{sec:fig_3}.}
  \label{fig:showcase_editing}
\end{figure}
\newpage


\newpage
\begin{figure}[H]
\vspace{-3em}
  \centering
  \includegraphics[width=1\textwidth]{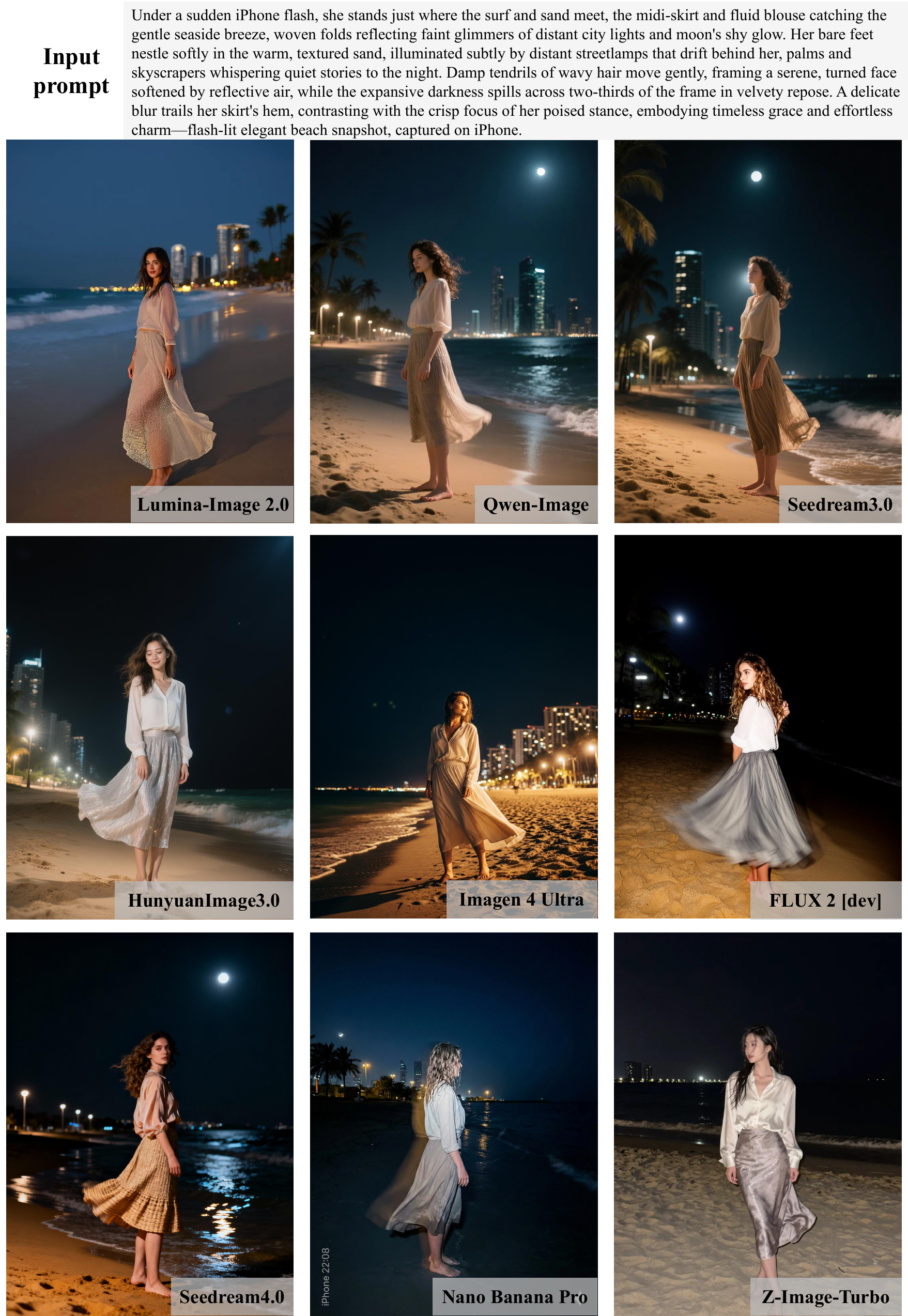}
  \caption{Showcases of comparison between Z-Image-Turbo and currently state-of-the-art models~\cite{qin2025lumina,qwenimage,cao2025hunyuanimage,nanopro,flux-2-2025,seedream2025seedream,gao2025seedream,google2025imagen4}. Z-Image-Turbo shows conspicuous photo-realistic generation capacity.}
  \label{fig:human_s_begin}
\end{figure}
\newpage


\newpage
{
    \hypersetup{linkcolor=black}
    \tableofcontents
}
\newpage


\section{Introduction}

The field of text-to-image (T2I) generation has witnessed remarkable advancements in recent years, evolving from generating rudimentary textures to producing photorealistic imagery with complex semantic adherence~\cite{podell2023sdxl,esser2024scaling,flux2023,qwenimage,seedream2025seedream,cao2025hunyuanimage,betker2023improving}. However, as the capabilities of these models have scaled, their development and accessibility face significant barriers. The current landscape is increasingly characterized by two divergent trends: on one side, state-of-the-art commercial closed-source models -- such as Nano Banana Pro~\cite{nanopro} and Seedream 4.0~\cite{seedream2025seedream} -- remain enclosed within ``black boxes'', offering high performance but limited transparency or reproducibility. On the other side, open-source models, while fostering democratization, often resort to massive parameter scaling -- approaching tens of billions of parameters (\emph{e.g.}, Qwen-Image~\cite{qwenimage} (20B), FLUX.2~\citep{flux-2-2025} (32B) and Hunyuan-Image-3.0~\cite{cao2025hunyuanimage} (80B) -- imposing prohibitive computational costs for both training and inference. In this context, distilling synthetic data from proprietary models has emerged as an appealing shortcut to train high-performing models at lower cost, becoming a prevalent approach for resource-constrained academic research~\cite{chen2023pixart,gao2024lumin-t2x}. However, this strategy risks creating a closed feedback loop that may lead to error accumulation and data homogenization, potentially hindering the emergence of novel visual capabilities beyond those already present in the teacher models.

In this work, we present \textbf{Z-Image}, a powerful diffusion transformer model that challenges both the ``scale-at-all-costs'' paradigm and the reliance on synthetic data distillation. We demonstrate that neither approach is necessary to develop a top-tier image generation model. Instead, we introduce the first comprehensive end-to-end solution that systematically optimizes every stage of the model lifecycle -- from data curation and architecture design to training strategies and inference acceleration -- enabling efficient, low-cost development on \textbf{purely real-world data without distilling results from other models}.

Most notably, this methodological efficiency allows us to complete the entire training workflow with remarkably low computational overhead. As detailed in Table~\ref{tab:training_costs}, the complete training pipeline for Z-Image requires only \textbf{314K H800 GPU hours}, translating to approximately \textbf{\$628K} at current market rates (about \$2 per GPU hour~\cite{leadergpu2025pricing}). In a landscape where leading models often demand orders of magnitude more resources, this modest investment demonstrates that principled design can effectively rival brute-force scaling.

\begin{table}[h!]
\label{tab:training_costs}
\centering
\caption{Training costs of Z-Image, assuming the rental price of H800 is about \$2 per GPU hour. The rental price refers from~\cite{leadergpu2025pricing}.}
\label{tab:training_costs}
\begin{tabular*}{\textwidth}{l @{\extracolsep{\fill}} cccc}
\toprule
\textbf{Training Costs} & \textbf{Low-res. Pre-Training} & \textbf{Omni-Pre-Training} & \textbf{Post-Training} & \textbf{Total} \\
\midrule
in H800 GPU Hours & 147.5K & 142.5K & 24K& 314K \\
in USD & \$295K & \$285K & \$48K & \$628K \\
\bottomrule
\end{tabular*}
\end{table}

This breakthrough in cost-efficiency is underpinned by a systematic methodology built on four pillars:
\begin{itemize}
\item \textbf{Efficient Data Infrastructure:} In resource-constrained scenarios, an efficient data infrastructure is pivotal; it serves to maximize the rate of knowledge acquisition per unit of time -- thereby accelerating training efficiency -- while simultaneously establishing the upper bound of model capabilities. To achieve this, we introduce a comprehensive Data Infrastructure composed of four synergistic modules: a \textit{Data Profiling Engine} for multi-dimensional feature extraction, a \textit{Cross-modal Vector Engine} for semantic deduplication and targeted retrieval, a \textit{World Knowledge Topological Graph} for structured concept organization, and an \textit{Active Curation Engine} for closed-loop refinement. By granularly profiling data attributes and orchestrating the training distribution, we ensure that the ``right data'' is aligned with the ``right stage'' of model development. This infrastructure maximizes the utility of real-world data streams, effectively eliminating computational waste arising from redundant or low-quality samples.

\item \textbf{Efficient Architecture:} Inspired by the remarkable scalability of decoder-only architectures in large language models~\cite{brown2020language}, we propose a \textbf{Scalable Single-Stream Multi-Modal Diffusion Transformer (S3-DiT)}. Unlike dual-stream architectures that process text and image modalities in isolation, our design facilitates dense cross-modal interaction at every layer. This high parameter efficiency enables Z-Image to achieve superior performance within a compact 6B parameter size, significantly lowering the hardware requirements for both training and deployment. The compact model size is also made possible in part by our use of a prompt enhancer (PE) to augment the model's complex world knowledge comprehension and prompt understanding capabilities, further mitigating the limitations of the modest parameter count. Furthermore, this early-fusion transformer architecture ensures superior versatility by treating tokens from different modalities uniformly -- including text tokens, image VAE tokens, and image semantic tokens -- enabling seamless handling of diverse tasks such as text-to-image generation and image-to-image editing within a unified framework.

\item \textbf{Efficient Training Strategy:} We design a progressive training curriculum composed of three strategic phases: (1) \textbf{Low-resolution Pre-training}, which bootstraps the model to acquire foundational visual-semantic alignment and synthesis knowledge at a fixed $256^2$ resolution. (2) \textbf{Omni-pre-training}, a unified multi-task stage that consolidates arbitrary-resolution generation, text-to-image synthesis, and image-to-image manipulation. By amortizing the heavy pre-training budget across these diverse capabilities, we eliminate the need for separate, resource-intensive stages. (3) \textbf{PE-aware Supervised Fine-tuning}, a joint optimization paradigm where Z-Image is fine-tuned using PE-enhanced captions. This ensures seamless synergy between the Prompt Enhancement module and the diffusion backbone without incurring additional LLM training costs, thereby maximizing the overall development efficiency of the Z-Image system.

\item \textbf{Efficient Inference:} We present \textbf{Z-Image-Turbo}, which delivers exceptional aesthetic alignment and high-fidelity visual quality in only 8 Number of Function Evaluations (NFEs). This performance is unlocked by the synergy of two key innovations: \textit{Decoupled DMD}~\cite{liu2025decoupled}, which explicitly disentangles the quality-enhancing and training-stabilizing roles of the distillation process, and \textit{DMDR}~\cite{jiang2025dmdr}, which integrates Reinforcement Learning by employing the distribution matching term as an intrinsic regularizer. Together, these techniques enable highly efficient generation without the typical trade-off between speed and quality.
\end{itemize}
Building upon this robust foundation and efficient workflow, we have successfully derived two specialized variants that address distinct application needs. First, our few-shot distillation scheme with reinforcement learning yields \textbf{Z-Image-Turbo}, an accelerated model that achieves exceptional aesthetic alignment in just 8 NFEs. It offers \textbf{sub-second inference\footnote{FlashAttention-3~\cite{shah2024flashattention} and \texttt{torch.compile}~\cite{ansel2024pytorch} is necessary for achieving sub-second inference latency.} latency on enterprise GPUs} and \textbf{fits within the memory constraints of consumer-grade hardware (<16GB VRAM)}. Second, leveraging the multi-task nature of our omni-pre-training, we introduce \textbf{Z-Image-Edit}, a model specialized for precise instruction-following image editing.

Extensive qualitative and quantitative experiments demonstrate the superiority of the Z-Image family. As illustrated in Figure~\ref{fig:showcase_realistic} and Figure~\ref{fig:showcase_text}, Z-Image delivers strong capabilities of photorealistic generation and exceptional bilingual (Chinese/English) text rendering, matching the visual fidelity of much larger models. Figure~\ref{fig:showcase_editing} showcases the capabilities of Z-Image-Edit, highlighting its precise adherence to editing instructions. Furthermore, qualitative comparisons in Figure~\ref{fig:human_s_begin} and Section~\ref{sec:qualitative_evaluation} reveal that our model rivals top-tier commercial systems, proving that \textbf{state-of-the-art results are achievable with significantly reduced computational overhead}. We publicly release our code, weights, and online demo to foster the development of accessible, budget-friendly generative models.

\section{Data Infrastructure}
\label{sec:data-infrastructure}
While the remarkable capabilities of state-of-the-art text-to-image models are underpinned by large-scale training data, achieving optimal performance under constrained computational resources necessitates a paradigm shift from data quantity to data efficiency. Simply scaling the dataset size often leads to diminishing returns; instead, an efficient training pipeline requires a data infrastructure that maximizes the information gain per computing unit. To this end, an ideal data system must be strictly curated to be \emph{conceptually broad} yet \emph{non-redundant}, exhibit \emph{robust multilingual text-image alignment}, and crucially, be \emph{structured for dynamic curriculum learning}, ensuring that the data composition evolves to match the model's training stages. To realize this, we have designed and implemented an integrated Efficient Data Infrastructure. Far from a static repository, this system operates as a dynamic engine architected to maximize the rate of knowledge acquisition within a fixed training budget. As the cornerstone of our pipeline, this infrastructure is composed of four core, synergistic modules:

\begin{enumerate}
    \item \textbf{Data Profiling Engine:} This module serves as the quantitative foundation for our data strategy.
    It extracts and computes a rich set of multi-dimensional features from raw data, spanning low-level physical attributes (\eg image metadata, clarity metrics) to high-level semantic properties (\eg, anomaly detection, textual description). These computed profiles are not merely for basic filtering; they are the essential signals used to quantify data complexity and quality, enabling the programmatic construction of curricula for our dynamic learning stages.

    \item \textbf{Cross-modal Vector Engine:} Built on billions of embeddings, this module is the engine for ensuring efficiency and diversity. It directly supports our goal of a \emph{non-redundant} dataset through large-scale semantic deduplication. Furthermore, its cross-modal search capabilities are critical for diagnosing and remediating model failures. This allows us to pinpoint and prune data responsible for specific failure cases and strategically sample to fill conceptual gaps.
    
    \item \textbf{World Knowledge Topological Graph:} This structured knowledge graph provides the semantic backbone for the entire infrastructure. It directly underpins our goal of \emph{conceptual breadth} by organizing knowledge hierarchically. Crucially, this topology functions as a semantic compass for data curation. It allows us to identify and fill conceptual voids in our dataset by traversing the graph to find underrepresented entities. Furthermore, it provides the structured framework needed to precisely rebalance the data distribution across different concepts during training, ensuring a more efficient and comprehensive learning process.

    \item \textbf{Active Curation Engine:} This module operationalizes our infrastructure into a truly dynamic, self-improving system. It serves two primary, synergistic functions. First, it acts as a frontier exploration engine, employing automatic sampling to identify concepts on which the model performs poorly or lacks knowledge (``hard cases"). Second, it drives a closed-loop data annotation pipeline. This ensures that every iteration not only expands \emph{conceptual breadth} of the dataset with high-value knowledge but also continuously refines the data quality, maximizing the learning efficiency of the entire training process.

\end{enumerate}

Collectively, these components forge a robust data infrastructure that not only fuels the training of text-to-image models but also establishes a versatile infrastructure for broader multimodal model training. Leveraging this system, we successfully facilitate the training of various critical components, including captioners, reward models, and our image editing model (\ie, Z-Image-Edit). In particular, we construct a dedicated data pipeline specifically for Z-Image-Edit upon this infrastructure, the details of which are elaborated in Section~\ref{sec:editing-data}.

\subsection{Data Profiling Engine}
\label{sec:data_profiling_engine}

The Data Profiling Engine is engineered to systematically process a massive, uncurated data pool, comprising large-scale internal copyrighted collections. It computes a comprehensive suite of multi-dimensional features for each image-text pair, enabling principled data curation. Recognizing that different data sources exhibit unique biases, our engine supports source-specific heuristics and sampling strategies to ensure a balanced and high-quality training corpus. The profiling process is structured across several key dimensions:

\paragraph{Image Metadata.} We begin by caching fundamental properties for each image. This includes elementary metadata like resolution (width and height) and file size, which facilitate efficient filtering based on resolution and aspect ratio. Simultaneously, we compute a perceptual hash (pHash) from the image's byte stream. This hash acts as a compact visual fingerprint, enabling rapid and effective low-level deduplication to remove identical or near-identical images. Together, these pre-computed attributes form the first layer of data selection.

\paragraph{Technical Quality Assessment.} The technical quality of an image is a critical determinant of model performance. Our engine employs a multi-faceted approach to quantify and filter out low-quality assets:
\begin{itemize}
    \item \textbf{Compression Artifacts:} To identify over-compressed images, we calculate the ratio of the ideal uncompressed file size (derived from resolution and bit depth) to the actual file size. A low ratio indicates potential quality degradation due to excessive compression.
    
    \item \textbf{Visual Degradations:} We utilize an in-house trained quality assessment model to score images on a range of degradation factors, including color cast, blurriness, perceptible watermarks, and excessive noise.
    
    \item \textbf{Information Entropy:} To maximize the density of meaningful content seen during training, we filter out low-entropy images. This is achieved through two complementary methods: (1) analyzing the variance of border pixels to detect images with large, uniform-color backgrounds or frames, and (2) performing a transient JPEG re-encoding and using the resulting bytes-per-pixel (BPP) as a proxy for image complexity.
\end{itemize}

\paragraph{Semantic and Aesthetic Content.} Beyond technical quality, we profile the high-level semantic and aesthetic properties of images:
\begin{itemize}
    \item \textbf{Aesthetic Quality:} We leverage an aesthetics scoring model, trained on labels from professional annotators, to quantify the visual appeal of each image.
    
    \item \textbf{AIGC Content Detection:} Following the findings of Imagen 3~\cite{baldridge2024imagen}, we trained a dedicated classifier to detect and filter out AI-generated content. This step is crucial for preventing degradation in the model's output quality and physical realism.
    
    \item \textbf{High-Level Semantic Tagging:} We have trained a specialized Vision-Language Model (VLM) to generate rich semantic tags. These tags include general object categories, human-centric attributes (\eg number of people), and culturally specific concepts, with a particular focus on elements relevant to Chinese culture. The same model also performs safety assessment by assigning Not-Safe-for-Work (NSFW) scores, allowing for the unified filtering of both semantically irrelevant and inappropriate content.
\end{itemize}

\paragraph{Cross-Modal Consistency and Captioning.} The alignment between an image and its textual description is paramount.
\begin{itemize}
    \item \textbf{Text-Image Correlation:} We use CN-CLIP~\cite{yang2022chinese} to compute the alignment score between an image and its associated alt caption. Pairs with low correlation scores are discarded to ensure the relevance of textual supervision.
    
    \item \textbf{Multi-Level Captioning:} For all images selected for pre-training, we generate a structured set of captions, including concise tags, short phrases, and detailed long-form descriptions. Notably, diverging from prior works~\cite{gao2025seedream, seedream2025seedream, qwenimage} that use separate modules for Optical Character Recognition (OCR) and watermark detection, our approach leverages the powerful inherent capabilities of our VLM. We explicitly prompt the VLM to describe any visible text or watermarks within the image, seamlessly integrating this information into the final caption. This unified strategy not only streamlines the data processing pipeline but also enriches the textual descriptions with critical visual details, as further elaborated in Section~\ref{sec:image_captioner}.
\end{itemize}

\subsection{Cross-modal Vector Engine}

We enhance the de-duplication method proposed in Stable Diffusion 3 \cite{esser2024scaling}, reformulating it as a scalable, graph-based community detection task. Addressing the severe scalability bottleneck of the original $range\_search$ function, we substitute it with a highly efficient k-nearest neighbor (k-NN) $search$ function. We construct a proximity graph from the k-NN distances and subsequently apply the community detection algorithm \cite{traag2019louvain}. This methodology closely approximates the original algorithm's output for a sufficiently large k while drastically reducing time complexity. Our fully GPU-accelerated \cite{cugraph_rapidsai} pipeline achieves a processing rate of approximately 8 hours per 1 billion items on 8 H800s, encompassing index construction and 100-NN querying. This approach not only ensures a non-redundant dataset by identifying dense clusters for effective de-duplication but also extracts semantic structures via modularity levels, facilitating fine-grained data balancing.

Furthermore, we constructed an efficient retrieval pipeline leveraging multimodal features \cite{yang2022chinese} combined with a state-of-the-art index algorithm \cite{ootomo2024cagra}. This system's cross-modal search capabilities are critical for both data curation and active model remediation. Beyond identifying distributional voids for strategically sampling to fill conceptual gaps -- thereby enabling targeted augmentation for a balanced pre-training distribution -- this engine is instrumental in diagnosing model failures. By querying the system with failure cases (\eg, problematic generated images or text prompts), we can pinpoint and prune the underlying data clusters responsible for the erroneous behavior. This iterative refinement process, targeting both data gaps and model failures, ensures dataset robustness and is pivotal for sourcing high-quality candidates for complex downstream tasks.

\subsection{World Knowledge Topological Graph}
\label{sec:wordknowlegde_topology_graph}

The construction of our knowledge graph follows a three-stage process. Initially, we build a comprehensive but redundant knowledge graph from all Wikipedia entities and their hyperlink structures. To refine this graph, we employ a two-pronged pruning strategy: first, centrality-based filtering removes nodes with exceptionally low PageRank~\cite{page1999pagerank} scores, which represent isolated or seldom-referenced concepts; second, visual generatability filtering uses a  VLM to discard abstract or ambiguous concepts that cannot be coherently visualized. Subsequently, to address the limited conceptual coverage of the pruned graph, we augment it by leveraging a large-scale internal dataset of captioned images. We extract tags and corresponding text embeddings from all available captions. Inspired by ~\cite{vo2024automatic}, we then perform an automatic hierarchical strategy on these embeddings. Each parent node is named by using a VLM to summarize its child nodes. This not only supplements the graph with new concept nodes but also organizes them into a structured taxonomic tree, significantly enhancing the structural integrity of the graph. In the final stage, we perform weight assignment and dynamic expansion to align the graph with practical applications. This involves manually curating and up-weighting high-frequency concepts from user prompts, and proactively integrating novel, trending concepts not yet present in our data pool to maintain the relevance and timeliness of the graph.

In application, this graph underpins our semantic-level balanced sampling strategy. We map the tags within each training caption to their corresponding nodes in the knowledge graph. By considering both the BM25~\cite{robertson2009probabilistic} score of a tag and its hierarchical relationships (\ie, parent-child links) within the graph, we compute a semantic-level sampling weight for each data point. This weight then guides our data engine to perform principled, staged sampling from the data pool, enabling fine-grained control over the training data distribution.
\begin{figure}[h!]
  \centering
  \includegraphics[width=1\textwidth]{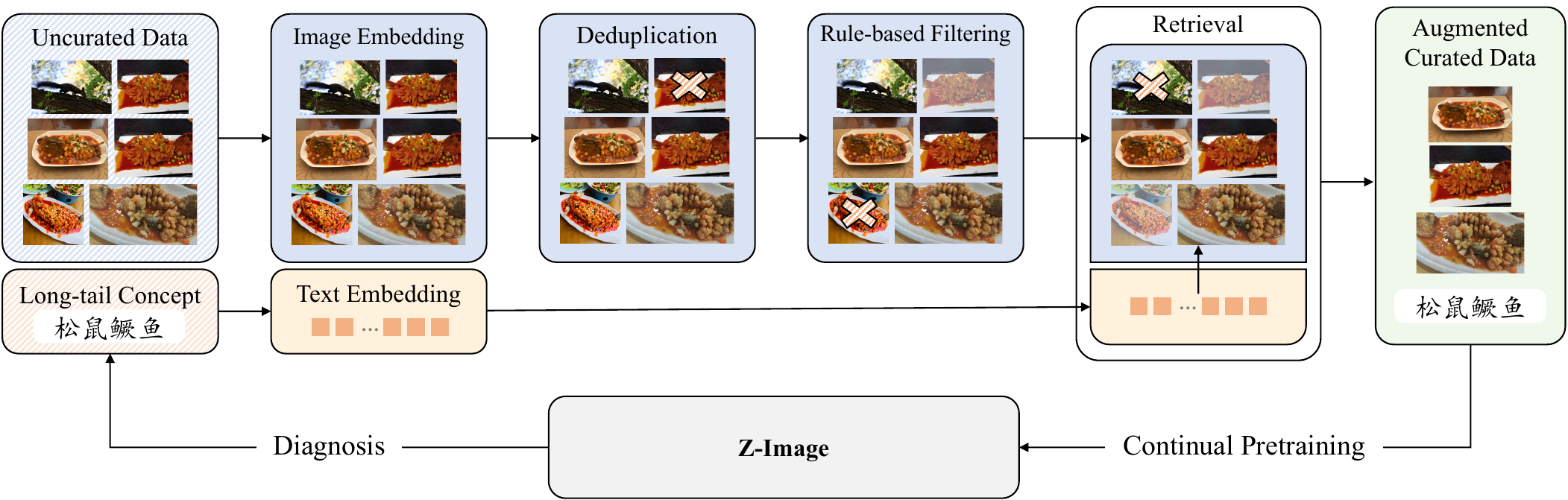}
  \caption{Overview of the Active Curation Engine. The pipeline refines uncurated data through cross-modal embedding, deduplication, and rule-based filtering to construct a high-quality augmented dataset. A feedback mechanism leverages the Z-Image model to diagnose long-tail distribution deficiencies, dynamically guiding cross-modal retrieval to reinforce the data collection process. The ``Squirrel Fish'' (\zh{松鼠鳜鱼}) case illustrates a classic long-tail challenge: it is actually the name of a Chinese cuisine but the model lacks the specific concept for this dish and may rely on compositional reasoning (combining ``Squirrel'' (\zh{松鼠}) and ``Fish'' (\zh{鳜鱼})), leading to erroneous generations absent of domain-specific training data.}
  \label{fig:retrieval}
\end{figure}

\subsection{Active Curation Engine}

\begin{figure}[h!]
  \centering
  \includegraphics[width=1\textwidth]{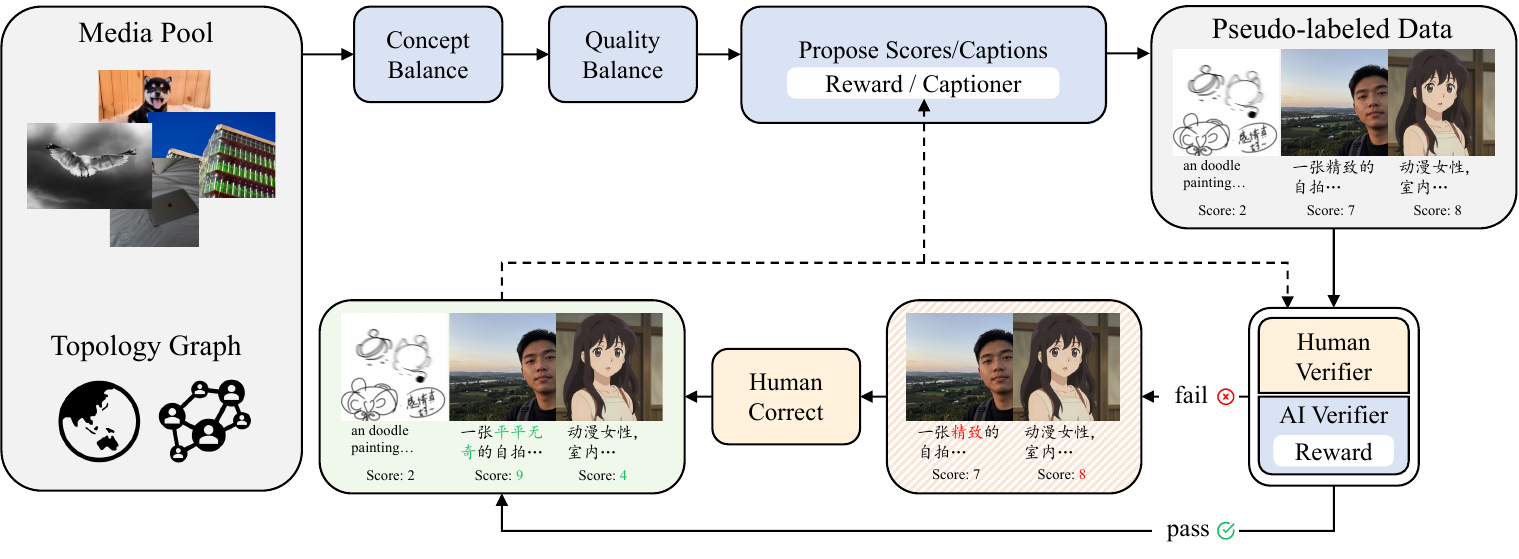}
  \caption{Illustration of the Human-in-the-Loop Active Learning Cycle. Data sampled from the media pool undergoes concept and quality balancing before being assigned pseudo-labels . A dual-verifier system (Human and AI) filters these proposals: approved samples pass directly, while rejected cases trigger a manual correction phase . This feedback loop iteratively refines the annotations and updates the topology graph to ensure high-precision alignment.}
  \label{fig:active_learning}
\end{figure}

To systematically elevate data quality and address long-tail distribution challenges, we deploy a comprehensive Active Curation Engine (Figure \ref{fig:retrieval}). This framework incorporates a filtering tool and Z-Image as a diagnostic generative prior. The pipeline begins by processing uncurated data through cross-modal embedding and deduplication, followed by rule-based filtering to eliminate low-quality samples.

To support the continuous evolution of Z-Image, we establish a human-in-the-loop active learning cycle (Figure~\ref{fig:active_learning}) where the reward model and captioner are progressively optimized. In this pipeline, we first employ the topology graph (Section~\ref{sec:wordknowlegde_topology_graph}) and the initial reward model to curate a balanced subset from the unlabeled media pool. The current captioner and reward model then assign pseudo-labels to these samples. A hybrid verification mechanism -- comprising both human and AI verifiers -- verifies these proposals; rejected samples trigger a manual correction phase by human experts to refine captions or scores. This high-quality annotated data is then used to retrain the captioner and reward model, thereby creating a virtuous cycle of our whole data infrastructure enhancement.



\subsection{Efficient Construction of Editing Pairs with Graphical Representation}
\label{sec:editing-data}

\begin{figure}[h!]
  \centering
  \begin{overpic}[width=0.95\linewidth]{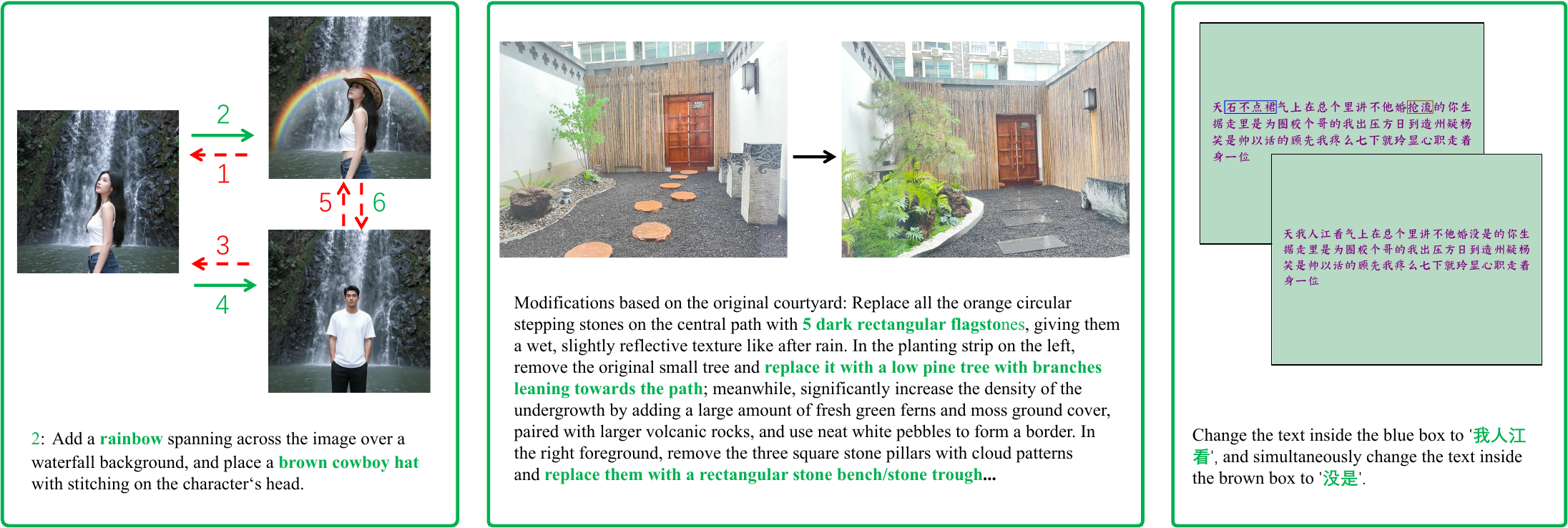}
    \put(0.2, -2.2){(a) Graphical Representation}
    \put(38.5, -2.2){(b) Paired Image from Videos}
    \put(72.0, -2.2){(c) Rendering for Text Editing}
    \put(1.0, 14.2){{\footnotesize {Input Image}}}
    \put(16.2, 6.5){{\footnotesize {Edited Images}}}
  \end{overpic}
  \vspace{2pt}
  \caption{Data construction for image editing using different strategies: 
  (a) arbitrarily permuting and combining different edited versions of the same input image 
  where the \textcolor{green}{green arrow} represents the pair constructed by task-specfic expert models and 
  the \textcolor{red}{red arrow} denotes the pair generated by combination and permutation, 
  (b) collecting images with inherent relationship from video frames, and 
  (c) controllable text rendering system for text editing.}
  \label{fig:i2i_data_pipeline}
\end{figure}
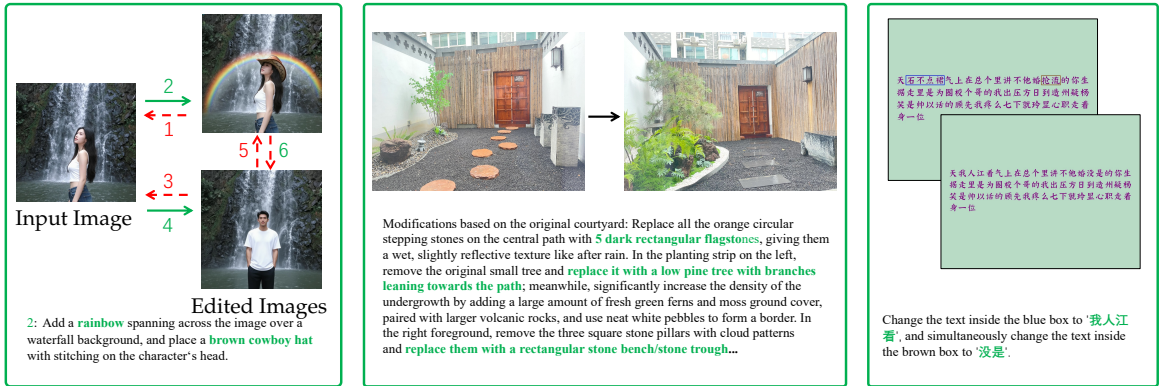

Collecting editing pairs that exhibits 
precise instruction following 
is challenging, 
owing to the requirement of consistency maintaining and the diverse and complex nature of editing operations.
Through scalable and controllable strategies as shown in Figure \ref{fig:i2i_data_pipeline}, 
we construct a large-scale training corpus from diverse sources.

\textbf{Mixed Editing with Expert Models.} 
To guarantee broad task coverage, 
we begin by curating a diverse taxonomy of editing tasks, 
and then leverage task-specific expert models to synthesize high-quality training data for each category.
To improve the training efficiency, 
we construct mixed-editing data, 
where multiple editing actions are integrated into one editing pair. 
Thus, 
the model can enhance its ability in multiple editing tasks from only a single composite pair, 
instead of relying on multiple ones.

\textbf{Efficient Graphical Representation.}
For an input image, we synthesize multiple edited versions corresponding to different editing tasks, 
enabling us to further scale the training data at zero cost through 
arbitrary pairwise combination~\cite{li2025visualcloze}~(\eg, 2$\tbinom{N+1}{2}$ pairs are constructed from one input image and its $N$ edited versions). 
Apart from scaling the quantity, 
this strategy 1) creates mixed-editing training data by combining two edited versions to enhance the training efficiency, 
and 2) yields inverse pairs to improve data quality, \ie, transforming a real, undistorted input image to an output image.

\textbf{Paired Images from Videos.}
Constructing image editing pairs from predefined tasks suffers from limited diversity. 
To overcome this issue, 
we leverage naturally grouped images collected from a large scale video frames in our media pool. 
These images, by sharing inherent relatedness (\emph{e.g.}, common subjects, scenes, or styles), implicitly define complex editing relationships among themselves. Building on this, we refine the data by calculating the cosine similarity between image embeddings using CN-CLIP~\cite{yang2022chinese}, allowing us to filter for pairs with high semantic relevance within each image group. 
The resulting dataset of video frame pairs offers three key advantages: 1) high task diversity, 2) inherent coupling of multiple edit types (\emph{e.g.}, simultaneous changes in human pose and background), and 3) superior scalability.

\textbf{Rendering for Text Editing.}
The acquisition of high-quality training data for text editing presents substantial challenges, 
where natural images suffer from the scarcity and imbalance of textual content, 
and text editing requires paired samples with precise operation annotations. 
To address these challenges, we develop a controllable text rendering system~\cite{qwenimage} that 
grants us precise control over not only the textual content but also its visual attributes, 
such as font, color, size, and position. 
This approach enables us to systematically generate a large-scale dataset of paired images, where the ground-truth editing instruction are known by the rendering operation, 
thereby directly overcoming the aforementioned data limitations.

\section{Image Captioner}
\label{sec:image_captioner}

\begin{figure}[htb]
    \centering
    \scalebox{0.6}{
        \begin{tikzpicture}[
        connect/.style={
                rounded corners=4pt,
                very thick,
                draw=black!80
            },
        arrow/.style={
                arrows = {-Straight Barb[length=1mm]},
                shorten >= 2pt,
                shorten <= 1.5pt,
                very thick
            },
        doublearrow/.style={
                arrows={Straight Barb[length=1mm]-Straight Barb[length=0.8mm]},
                shorten >=2pt,
                shorten <=2pt,
                very thick
            },
        dashedarrow/.style={
                arrows = {-Straight Barb[length=1mm]},
                dashed,
                shorten >= 2pt,
                shorten <= 1.5pt,
                very thick
            }
        ]
            \pgfdeclarelayer{background}
            \pgfsetlayers{background,main}

            \node[anchor=west] (image1) at (0, 0) {\includegraphics[width=0.12\textwidth]{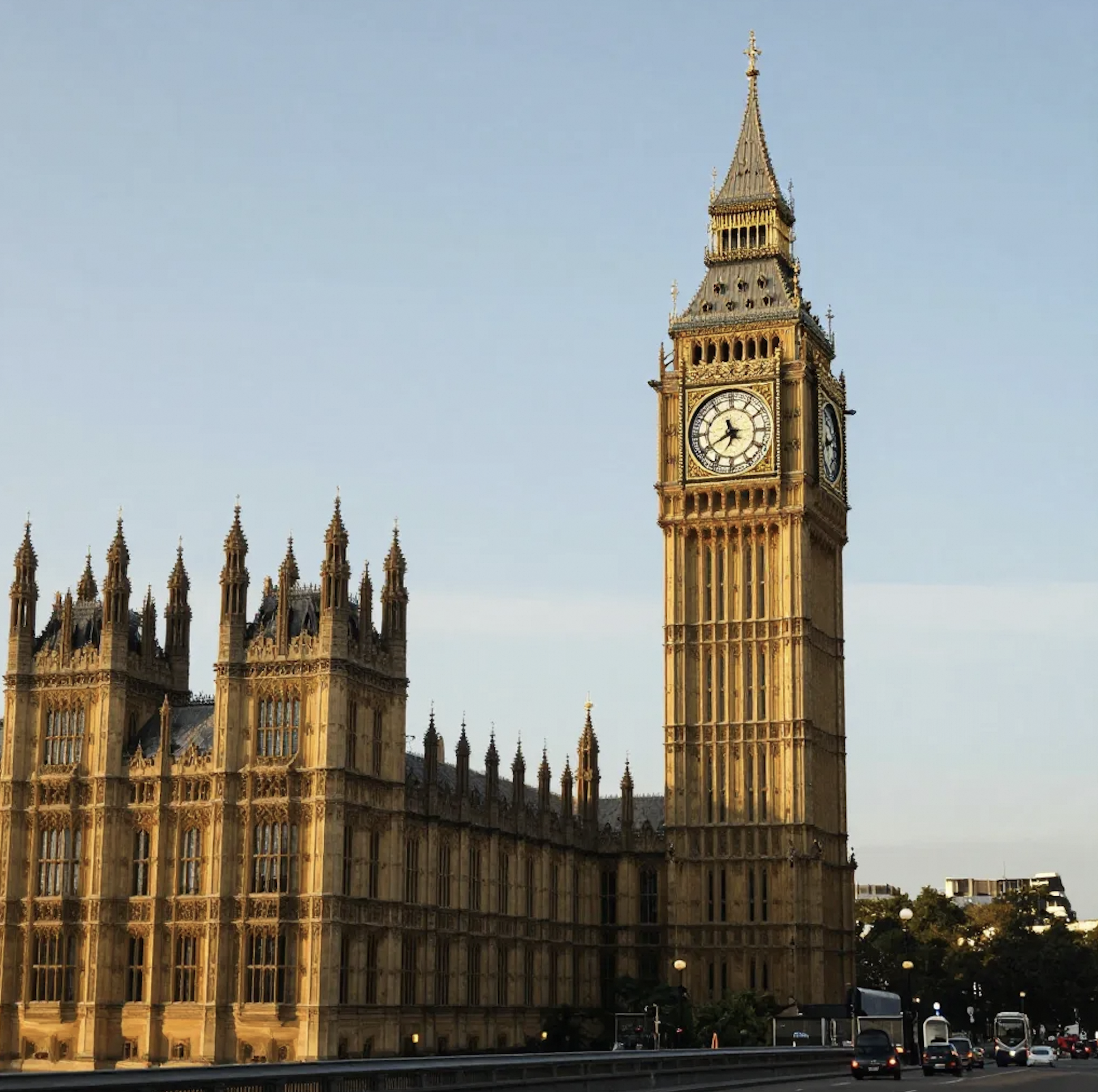}};
            \node[anchor=center] (image2) at ($(image1.south) + (0cm, -1.5cm)$) {\includegraphics[width=0.12\textwidth]{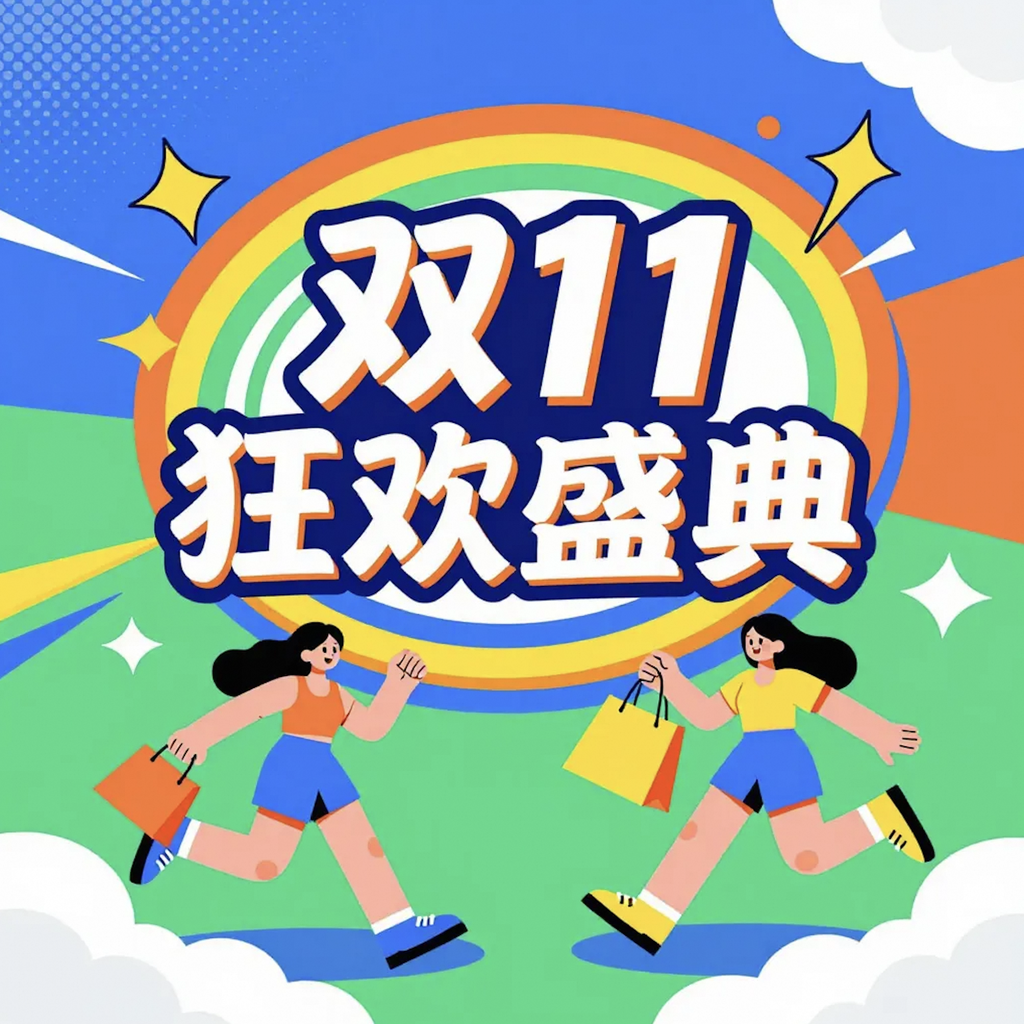}};
            \node[anchor=center] (image3) at ($(image1.south) + (-2.5cm, -1.5cm)$) {\includegraphics[width=0.12\textwidth]{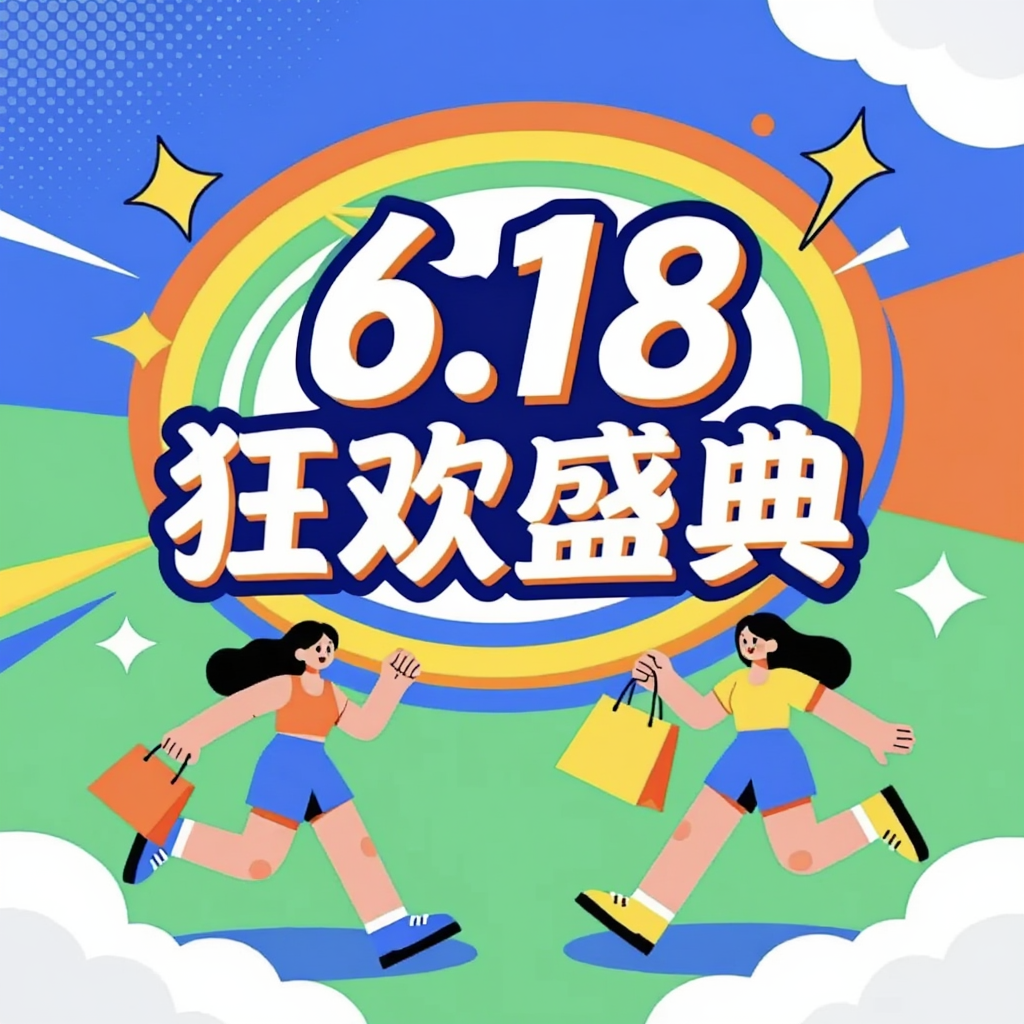}};

            \node[draw, thick, rounded corners, dashed, inner sep=2pt, fit=(image1), label={[xshift=-0.cm, yshift=0.2cm, anchor=south]\textbf{Single Image}}] (image_input) {};
            \node[draw, thick, rounded corners, dashed, inner sep=2pt, fit=(image2) (image3), label={[xshift=-0.cm, yshift=-2.6cm, anchor=north]\textbf{Image Pair}}] (image_pair_input) {};

            \node[draw, rounded corners, thick, fill=orange!10, minimum width=3.5cm, minimum height=4.5cm, text centered, anchor=center, align=center] (caption_model) at ($(image1)!0.5!(image3) + (6.5cm, 0.0cm)$) {\large{Z-Captioner}\\ \large{Model}};
            \node[anchor=center] (world) at ($(caption_model.north) + (0cm, +1.5cm)$) {\includegraphics[width=0.12\textwidth]{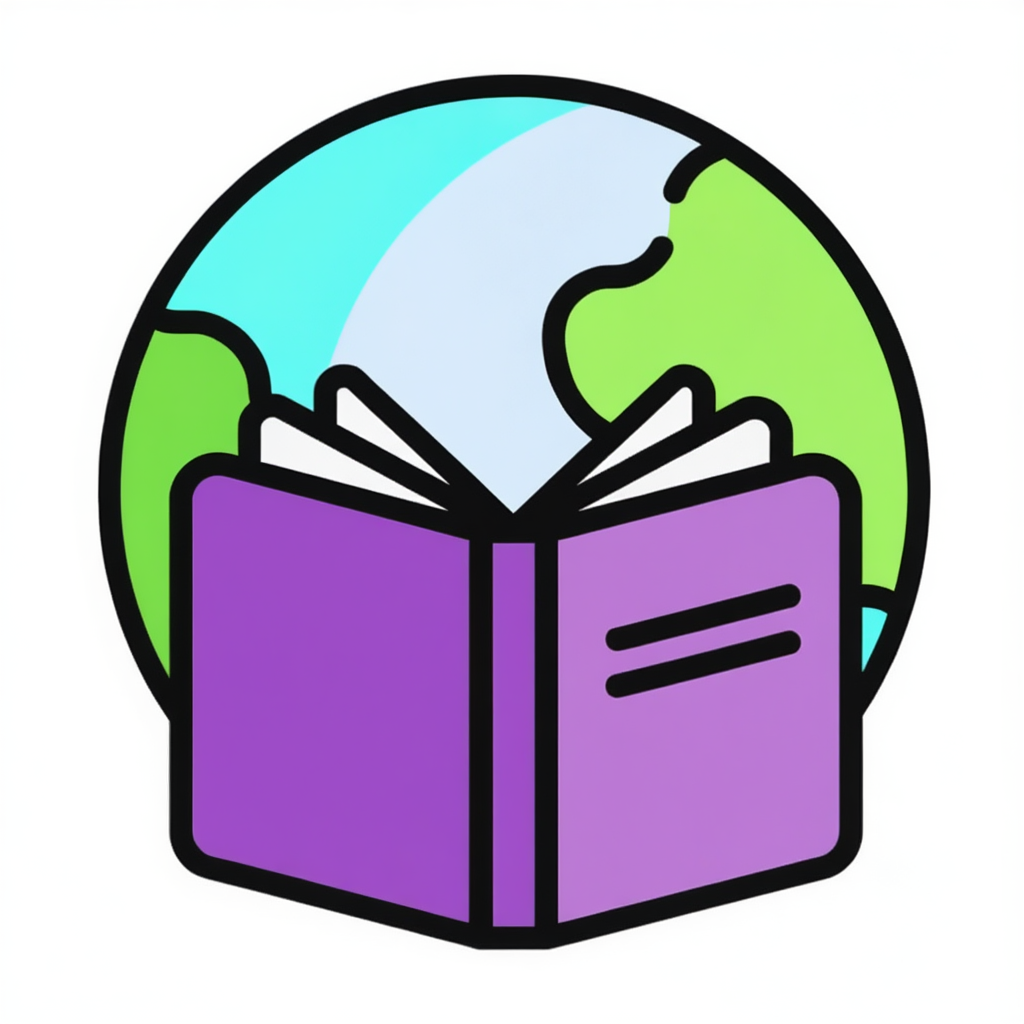}};
            \draw[arrow, connect] ($(world.south) + (0.0cm, +0.3cm)$) -- ($(caption_model.north) + (0.0cm, -0.0cm)$);
            \node[anchor=center] (world_knowledge) at ($(world.north) + (0cm, +0.0cm)$) {\textbf{World Knowledge}};
            \node[anchor=center] (ocr) at ($(caption_model.south) + (0cm, -1.5cm)$) {\includegraphics[width=0.15\textwidth]{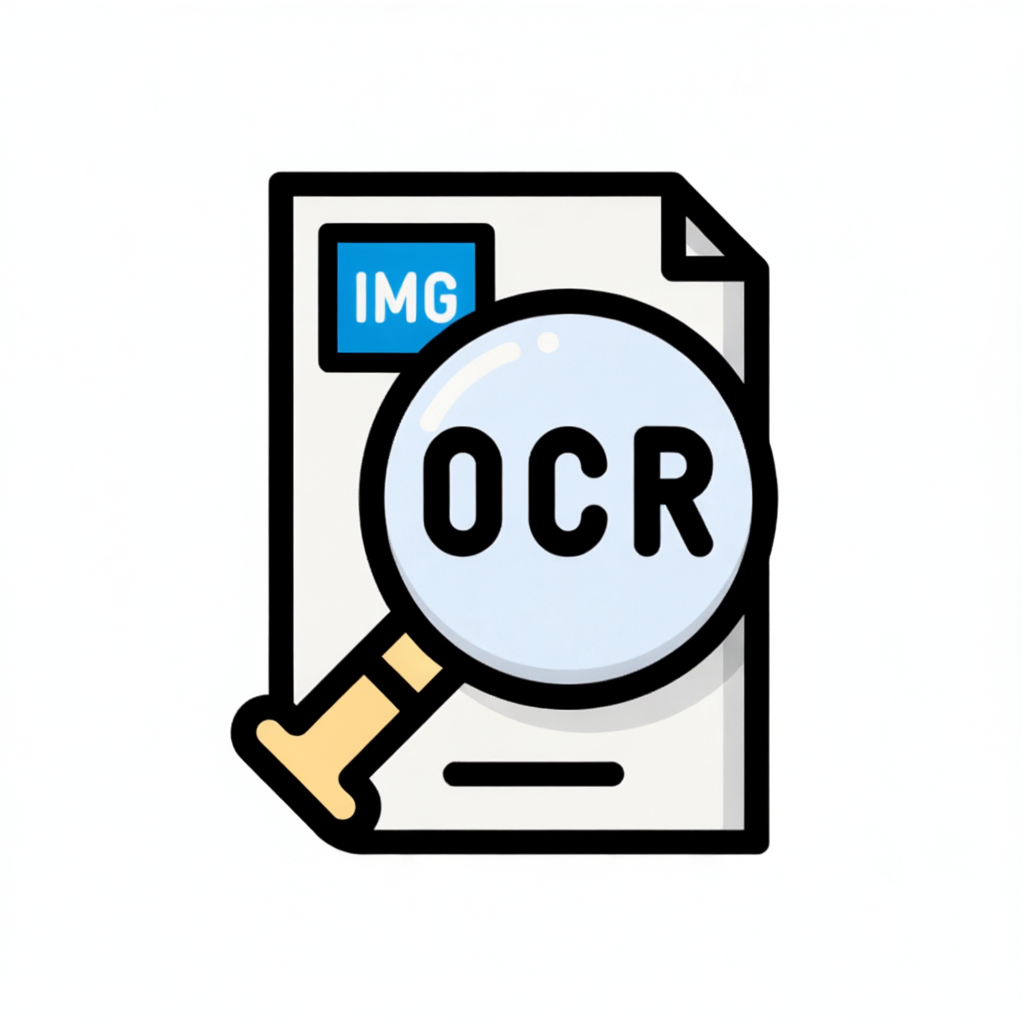}};
            \draw[arrow, connect] ($(ocr.north) + (0.0cm, -0.5cm)$) -- ($(caption_model.south) + (0.0cm, -0.0cm)$);
            \node[anchor=center] (ocr_aug) at ($(ocr.south) + (0cm, +0.0cm)$) {\textbf{OCR Augmentation}};

            \node[draw, thick, fill=gray!5, anchor=north, minimum width=1.8cm, minimum height=1.8cm, align=left, inner sep=6pt, anchor=west] (tag_caption) at ($(caption_model.east) + (1.4cm, +1.5cm)$) {Tagging \\ Caption};

            \node[draw, thick, fill=gray!5, anchor=north, minimum width=1.8cm, minimum height=1.8cm, align=left, inner sep=6pt, anchor=west] (short_caption) at ($(tag_caption.east) + (0.8cm, +0cm)$) {Short \\Caption};

            \node[draw, thick, fill=gray!5, anchor=north, minimum width=1.8cm, minimum height=1.8cm, align=left, inner sep=6pt, anchor=west] (long_caption) at ($(short_caption.east) + (0.8cm, +0cm)$) {Long \\Caption};

            \node[draw, thick, rounded corners, dashed, inner sep=10pt, fit=(tag_caption) (short_caption) (long_caption), label={[xshift=-0.cm, yshift=0.2cm, anchor=south]\textbf{Text-to-Image Captions}}] (text_to_image_group) {};

            \node[draw, thick, fill=gray!5, anchor=north, minimum width=1.8cm, minimum height=1.8cm, align=left, inner sep=6pt, anchor=north] (step1) at ($(tag_caption) + (0.cm, -2.2cm)$) {Step1: \\Caption};

            \node[draw, thick, fill=gray!5, anchor=north, minimum width=1.8cm, minimum height=1.8cm, align=left, inner sep=6pt, anchor=west] (step2) at ($(step1.east) + (0.8cm, +0cm)$) {Step2: \\Analysis};

            \node[draw, thick, fill=gray!5, anchor=north, minimum width=1.8cm, minimum height=1.8cm, align=left, inner sep=6pt, anchor=west] (step3) at ($(step2.east) + (0.8cm, +0cm)$) {Step3:\\Instruction};

            \node[draw, thick, rounded corners, dashed, inner sep=10pt, fit=(step1) (step2) (step3), label={[xshift=-0.cm, yshift=-3.3cm, anchor=south]\textbf{Image Editing Instructions}}] (image_editing_group) {};


            \draw[arrow, connect] ($(caption_model.east) + (0.0cm, +1.25cm)$) -- ($(text_to_image_group.west) + (0.0cm, -0.25cm)$);

            \draw[arrow, connect] ($(caption_model.east) + (0.0cm, -1.25cm)$) -- ($(image_editing_group.west) + (0.0cm, +0.37cm)$);

            \draw[arrow, connect] ($(image_input.east) + (0.0cm, +0cm)$) -- ($(caption_model.west) + (0.0cm, +1.3cm)$);
            \draw[arrow, connect] ($(image_pair_input.east) + (0.0cm, 0cm)$) -- ($(caption_model.west) + (0.0cm, -1.3cm)$);
        \end{tikzpicture}
    }
    \caption{Pipeline for generating text-to-image captions and image editing instructions. OCR results (obtained through CoT) and world knowledge (from meta information) are explicitly included into the captions.}
    \label{fig:caption}
\end{figure}
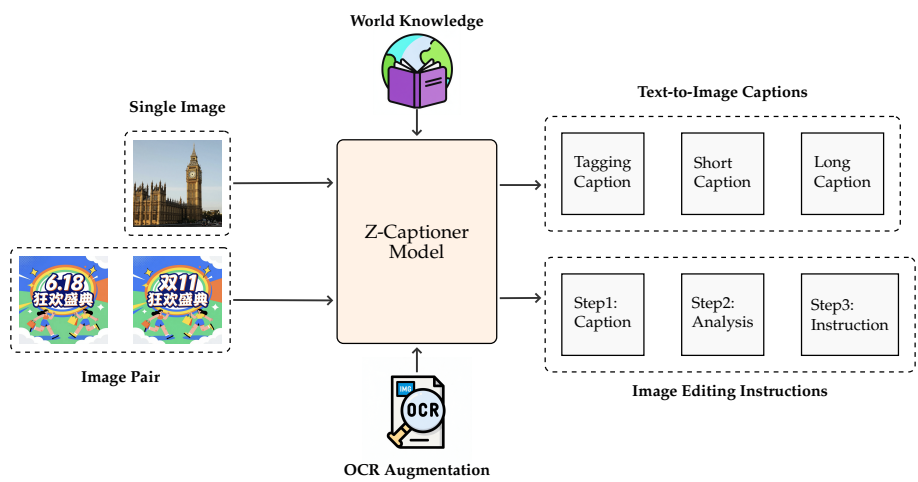

\begin{figure}[h!]
  \centering
  \includegraphics[width=1\textwidth]{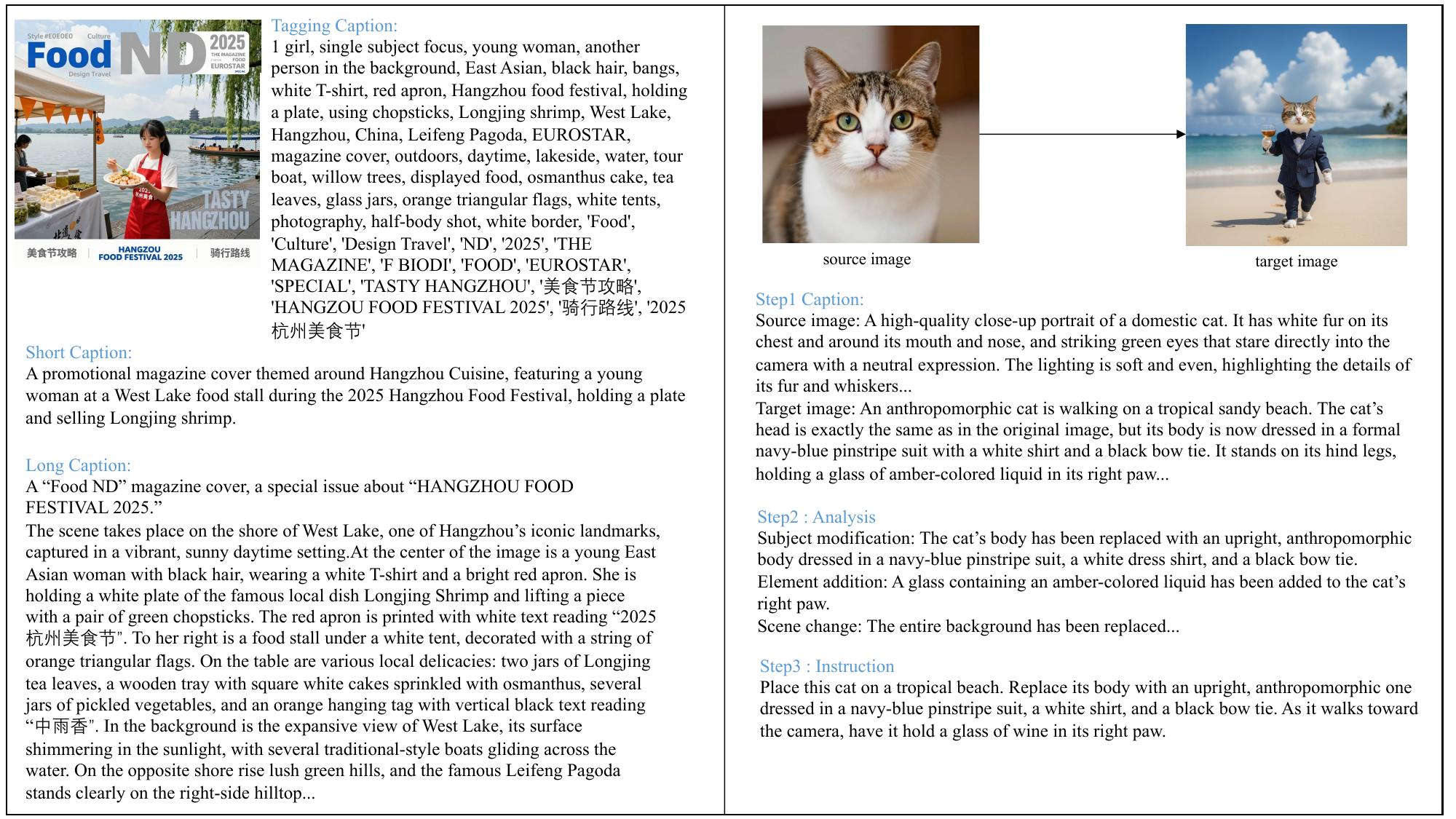}
  \caption{Single image caption and difference caption examples. Left: for single image, we have captions of different types and lengths, and notably, OCR results (all the texts transcribed in their original languages) and world knowledge (explicitly and correctly recognizing the famous beauty spot, West Lake, Hangzhou, China, in this example) is included. Right: difference captions are composed step-by-step.}
  \label{fig:caption_show}
\end{figure}

We build an all-in-one image captioner, Z-Captioner, by incorporating multiple types of image caption. As revealed in previous works \cite{lu2025omnicaptioner}, different captioning tasks can benefit each other as they share the same goal of understanding and depicting images. 
Our model is designed not only to describe visual elements, but also to leverage extensive world knowledge to interpret the semantic context of the image.
The integration of world knowledge is particularly critical for the downstream text-to-image synthesis task, as it enables the model to accurately render images involving specific named entities. 
Figure \ref{fig:caption} shows our pipeline for generating text-to-image captions and image editing instructions.

\subsection{Detailed Caption with OCR Information}
First, we specially emphasize that according to our experiments, including explicit OCR information in image captions is inextricably bound with accurate text rendering in the generated images. Therefore, we employ a way that shares the same spirit as Chain-of-Thought (CoT) \cite{wei2022chain}, by first explicitly recognizing all optical characters in the image and then generating a caption based on the OCR results. This effectively mitigates missing texts compared to directly generating a caption that encapsulates everything, especially for the cases where texts are very long/dense. In addition, we force the OCR results to remain in their original languages without any translation, avoiding them being falsely rendered in their translated languages. 

\subsection{Multi-Level Caption with World Knowledge}
We design five different types of image captions in total, including long, medium and short captions, as well as tags and simulated user prompts. With the data infrastructure in Section \ref{sec:data-infrastructure}, we include world knowledge in all five types of captions by performing image captioning conditioned on meta information. This significantly alleviates hallucinations when our captioner identifies and names specific entities such as public figures, famous landmarks, or known events.

To be specific, for relatively long captions, we include very dense information of the images, in order that the model could learn a mapping from the text to the image as accurate as possible. These captions contain full OCR results as mentioned above, along with subjects, objects, background, location information, et al. 
We deliberately adopt a plain and objective linguistic style for our descriptions, strictly confining them to factual information observable in the image. By inhibiting subjective interpretations and imaginative associations, our purpose is to enhance data efficiency for the image generation task by eliminating non-essential information.

On the other hand, short captions, tags and simulated user prompts are designed for the model to adapt to real user prompts (which are usually short and unspecific) for better user experience. Notably, most of the simulated user instructions are incomplete prompts. They differ from short captions in that a short caption provides a relatively complete and comprehensive description of the entire image. In contrast, a short simulated prompt may mimic user behavior by focusing only on specific parts of interest to the user, while making no mention of the rest of the image.

\subsection{Difference Caption for Image Editing}
Difference caption is a concise editing instruction specifying the transformation from a source to a target image. 
To generate this, we employ a three-step CoT process that systematically breaks down the comparative task \cite{zhuo2025factuality}.
\begin{enumerate}
\item \textbf{Step1: Detailed Captioning.} We first generate a comprehensive, OCR-inclusive caption for both the source and target images respectively. This step provides a structured, detailed representation of each image's content.

\item \textbf{Step2: Difference Analysis.} The model then performs a comparative analysis, leveraging both the raw images and their generated captions, to tell all discrepancies from visual and textual  perspectives.

\item \textbf{Step3: Instruction Synthesis.} Finally, the model generates a concise editing instruction based on the identified differences.
\end{enumerate}
This step‑by‑step process helps the model create clear and useful instructions by moving from understanding, to comparing, and finally to generating the instructions.

\section{Model Training}

This section presents the complete training pipeline of Z-Image and Z-Image-Edit. We begin by introducing our Scalable Single-Stream Diffusion Transformer (S3-DiT) architecture (Section~\ref{sec:architecture}) and training efficiency optimizations (Section~\ref{sec:training_efficiency}), followed by a multi-stage training process: pre-training (Section~\ref{sec:pre_training}), supervised fine-tuning (Section~\ref{sec:sft}), few-step distillation (Section~\ref{sec:distillation}), and reinforcement learning with human feedback (Section~\ref{sec:rl}). Finally, we describe the continued training strategy for image editing capabilities (Section~\ref{sec:editing}) and our reasoning-enhanced prompt enhancer (Section~\ref{sec:pe}). The overall training pipeline is summarized in Figure~\ref{fig:pipeline}. And in Figure~\ref{fig:stage_showcase}, we present intermediate generation results throughout Z-Image's training process to demonstrate the benefits contributed by each stage.

\begin{figure}[h!]
  \centering
  \includegraphics[width=1\textwidth]{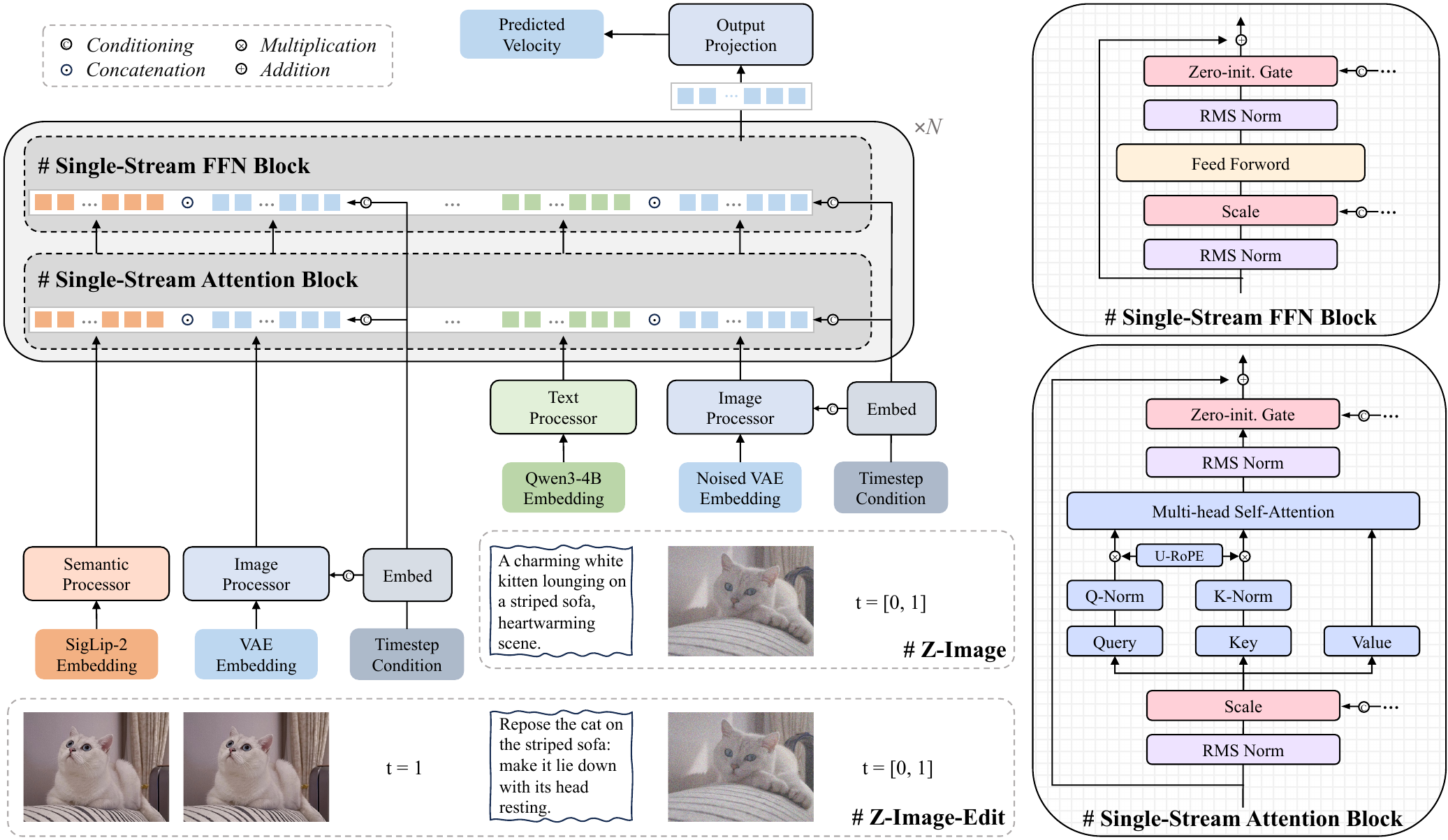}
  \caption{Architecture overview of the Z-Image series. The S3-DiT consists of single-stream FFN blocks and single-stream attention blocks. It processes inputs from different modalities through lightweight modality-specific processors, then concatenates them into a unified input sequence. This modality-agnostic architecture maximizes cross-modal parameter reuse to ensure parameter efficiency, while providing flexible compatibility for varying input configurations in both Z-Image and Z-Image-Edit.}
  \label{fig:model_arch}
\end{figure}

\subsection{Architecture}
\label{sec:architecture}
Efficiency and stability are the core objectives guiding the design of Z-Image. To achieve this, we employ the lightweight Qwen3-4B~\cite{yang2025qwen3} as the text encoder, leveraging its bilingual proficiency to align complex instructions with visual content. For image tokenization, we utilize the Flux VAE~\cite{flux2023} selected for its proven reconstruction quality. Exclusively for editing tasks, we augment the architecture with SigLIP 2~\cite{tschannen2025siglip} to capture abstract visual semantics from reference images. Inspired by the scaling success of decoder-only models, we adopt a Single-Stream Multi-Modal Diffusion Transformer (MM-DiT) paradigm~\cite{esser2024scaling}. In this setup, text, visual semantic tokens, and VAE image tokens are concatenated at the sequence level to serve as a unified input stream, maximizing parameter efficiency compared to dual-stream approaches~\cite{esser2024scaling,qwenimage}. We employ 3D Unified RoPE~\cite{qin2025lumina,wu2025omnigen2} to model this mixed sequence, wherein image tokens expand across spatial dimensions and text tokens increment along the temporal dimension. Crucially, for editing tasks, the reference image tokens and target image tokens are assigned aligned spatial RoPE coordinates but are separated by a unit interval offset in the temporal dimension. Additionally, different time-conditioning values are applied to the reference and target images to distinguish between clean and noisy images.

As illustrated in Figure~\ref{fig:model_arch}, the specific architecture of our S3-DiT (Scalable Single-Stream DiT) commences with lightweight modality-specific processors, each composed of two transformer blocks for initial modal alignment. Subsequently, tokens enter the unified single-stream backbone. To ensure training stability, we implement QK-Norm to regulate attention activations~\cite{karras2024analyzing, luo2018cosine, gidaris2018dynamic, nguyen2023enhancing} and Sandwich-Norm to constrain signal amplitudes at the input and output of each attention / FFN blocks~\cite{ding2021cogview,gao2024lumina-next}. For conditional information injection, input condition vectors are projected into scale and gate parameters to modulate the normalized inputs and outputs of both Attention and FFN layers. To reduce parameter overhead, this projection is decomposed into a low-rank pair: a shared, layer-agnostic down-projection layer followed by layer-specific up-projection layers. Finally, RMSNorm~\cite{zhang2019root} is uniformly utilized for all the aforementioned normalization operations.

\begin{table}[thbp]
\centering
\caption{Architecture Configurations of S3-DiT.}
\label{tab:model_architectures_transposed}
\begin{tabular}{c|ccc}
\toprule
\textbf{Configuration} & \quad \quad \quad & \textbf{S3-DiT} & \quad \quad \quad \\
\midrule
Total Parameters & \quad \quad \quad & 6.15B &\quad \quad \quad \\
Number of Layers & \quad \quad \quad & 30 & \quad \quad \quad \\
Hidden Dimension & \quad \quad \quad & 3840 & \quad \quad \quad \\
Number of Attention Heads & \quad \quad \quad & 32 & \quad \quad \quad \\
FFN Intermediate Dimension & \quad \quad \quad & 10240 & \quad \quad \quad \\
$(d_t, d_h, d_w)$ & \quad \quad \quad & $(32, 48, 48)$ & \quad \quad \quad \\
\bottomrule
\end{tabular}
\vspace{0.5em}
\footnotesize
\end{table}

\begin{figure}[thbp]
  \centering
  \includegraphics[width=0.8\textwidth]{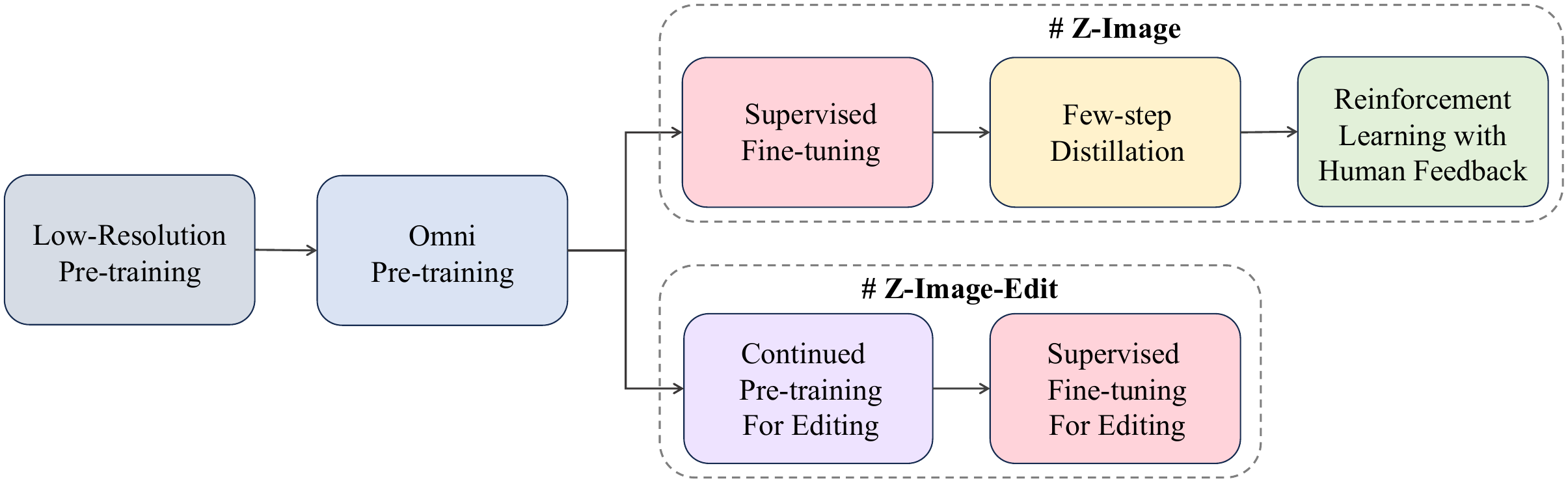}
  \caption{The training pipeline of Z-Image and Z-Image-Edit. The low-resolution pre-training and omni-pre-training stages provide a suitable initialization for image generation and editing tasks, after which separate post-training processes yield the Z-Image and Z-Image-Edit models respectively.}
  \label{fig:pipeline}
\end{figure}

\subsection{Training Efficiency Optimization}
\label{sec:training_efficiency}
To optimize training efficiency, we implemented a multi-faceted strategy targeting both computational and memory overheads.

For distributed training, we employed a hybrid parallelization strategy. We applied standard Data Parallelism (DP) to the VAE and Text Encoder, as they remain frozen and incur minimal memory footprint. In contrast, for the large DiT model, where optimizer states and gradients consume substantial memory, we utilized FSDP2~\cite{zhao2023pytorch} to effectively shard these overheads across GPUs. Furthermore, we implemented gradient checkpointing across all DiT layers. This technique trades an acceptable increase in computational cost for significant memory savings, enabling larger batch sizes and improved overall throughput. To further accelerate computation and optimize memory usage, the DiT blocks were compiled using \texttt{torch.compile}, a just-in-time (JIT) compiler~\cite{ansel2024pytorch}.

In addition to system-level optimizations, we addressed inefficiencies arising from mixed-resolution training. Grouping samples with significantly different sequence lengths into a single batch typically results in excessive padding, which significantly impedes overall training speed. To mitigate this, we designed a sequence length-aware batch construction strategy. Prior to training, we estimate the sequence length of each sample based on the resolution (height and width) recorded in the metadata. The sampler then groups samples with similar sequence lengths into the same batch, thereby minimizing computational waste. Crucially, we additionally employ a dynamic batch sizing mechanism: smaller batch sizes are assigned to long-sequence batches to prevent Out-Of-Memory (OOM) errors, while larger batch sizes are used for short sequences to avoid resource vacancy. This approach ensures maximal hardware utilization across varying resolutions.

\subsection{Pre-training}
\label{sec:pre_training}

Z-Image is trained using the flow matching objective~\cite{lipman2022flow,liu2022flow}, where noised inputs are first constructed through linear interpolation between Gaussian noise $x_0$ and the original image $x_1$, \emph{i.e.}, $x_t = t \cdot x_1 + (1 - t) \cdot x_0$. The model is then trained to predict the velocity of the vector field that defines the path between them, \emph{i.e.}, $v_t = x_1 - x_0$. The training objective can be formulated as:
\begin{equation}
    \mathcal{L} = \mathbb{E}_{t,x_0,x_1,y}[\|u(x_t,y,t;\theta)-(x_1 - x_0)\|^2],
\end{equation}
Where $\theta$ as the learnable parameters and $y$ as the conditional embedding. Following SD3~\cite{esser2024scaling}, we employ the logit-normal noise sampler to concentrate the training process on intermediate timesteps. Additionally, to account for the variations in Signal-to-Noise Ratio (SNR) arising from our multi-resolution training setup, we adopt the dynamic time shifting strategy as used in Flux~\cite{flux2023}. This ensures that the noise level is appropriately scaled for different image resolutions, leading to more effective training.

The pre-training of Z-Image can be broadly divided into two phases: low-resolution pre-training and omni-pre-training.

\textbf{Low-resolution Pre-training}. This phase consists of a single stage, conducted exclusively at a $256^2$ resolution on the text-to-image generation task. The primary emphasis of this stage is on efficient cross-modal alignment and knowledge injection -- equipping the model with the capability to generate a diverse range of concepts, styles, and compositions, which is consistent with the initial stage of conventional multi-stage training protocols. As shown in Figure~\ref{tab:training_costs}, this phase accounts for over half of our total pre-training compute. This allocation is based on the rationale that the majority of the model's foundational visual knowledge (\emph{e.g.}, Chinese text rendering) is acquired during this low-resolution training stage.

\textbf{Omni-pre-training}. The ``omni'' here signifies three key aspects:

\begin{itemize}

\item \textbf{Arbitrary-Resolution Training}: We design an arbitrary-resolution training strategy in which the original image resolution is mapped to a predefined training resolution range through a resolution-mapping function. The model is then trained on images with diverse resolutions and aspect ratios. This enables the learning of cross-scale visual information, mitigates information loss caused by downsampling to a fixed resolution, and improves overall data efficiency.

\item \textbf{Joint Text-to-Image and Image-to-Image Training}: We integrate the image-to-image task into the pre-training framework. By leveraging the substantial compute budget available during pre-training, we can effectively exploit large-scale, naturally occurring, and weakly aligned image pairs, as discussed in Section~\ref{sec:editing-data}. Learning the relationships between natural image pairs provides a strong initialization for downstream tasks such as image editing. Importantly, we observe that this joint pre-training scheme does not introduce any noticeable performance degradation on the text-to-image task.

\item \textbf{Multi-level and Bilingual Caption Training}: It is widely recognized that high-quality captions are crucial for training text-to-image models~\cite{betker2023improving}. To ensure both bilingual understanding and strong native prompt-following capability, we employ Z-Captioner to generate bilingual, multi-level synthetic captions (including long, medium, and short descriptions, as well as tags and simulated user prompts). In addition, the original textual metadata associated with each image is incorporated with a small probability to further enhance the model’s acquisition of world knowledge. The use of captions at different granularities and from diverse perspectives provides broad mode coverage, which is beneficial for subsequent stages of training. Moreover, for image-to-image tasks, we randomly sample either the target image's caption or the pairwise difference caption with a certain probability, corresponding to reference-guided image generation and multi-task image editing, respectively.

\end{itemize}

Working with our data infrastructure, the omni-pre-training phase is conducted in multiple stages.
Upon completion of the final stage, the model becomes capable of generating images at arbitrary resolutions up to the 1k-1.5k range and can condition its output on both image and text inputs. This provides a suitable starting point for the subsequent training of Z-Image and Z-Image-Edit.

\begin{figure}[h!]
  \centering
  \includegraphics[width=1\textwidth]{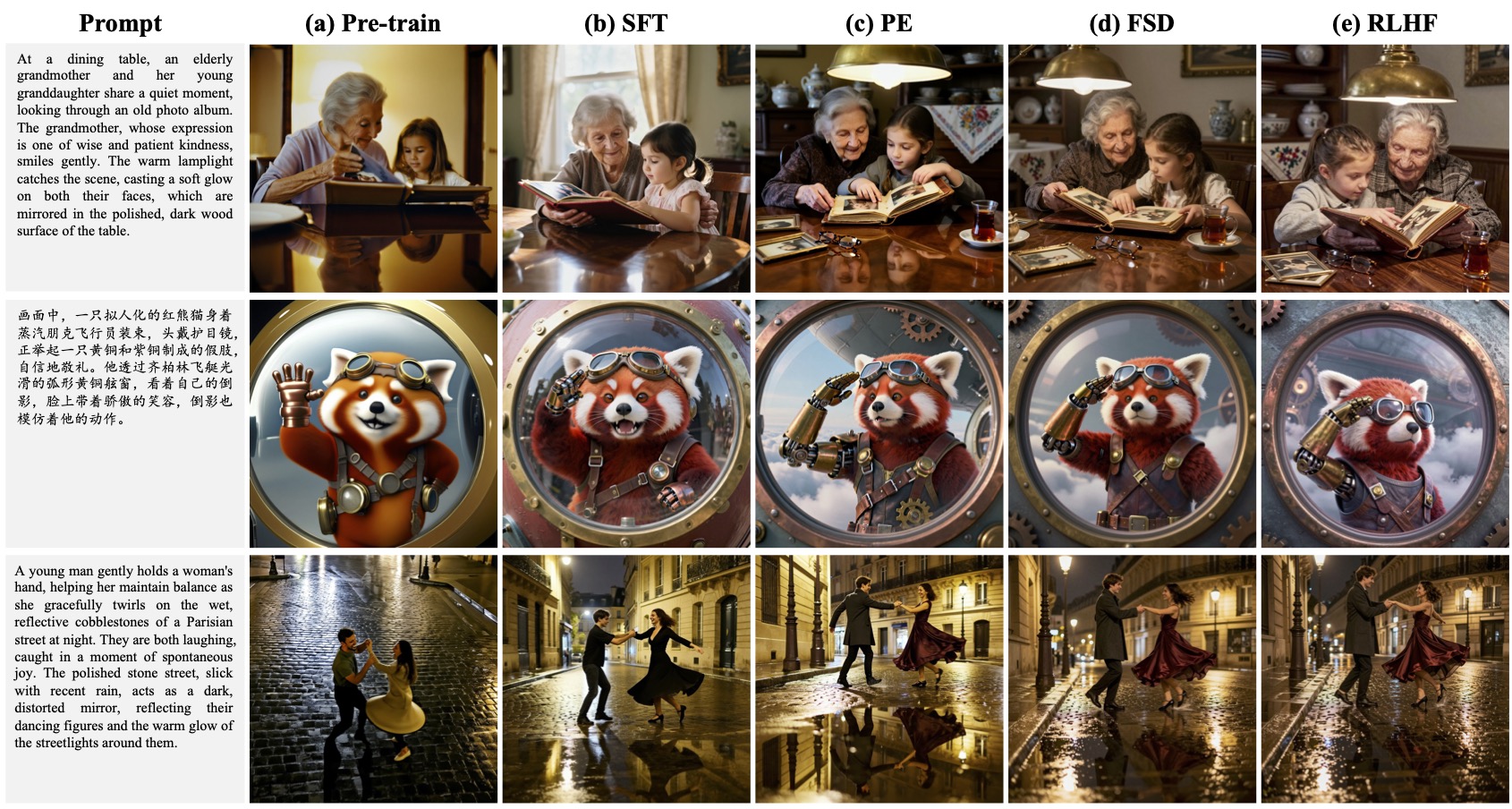}
  \caption{Intermediate generation results throughout Z-Image-Turbo's training process, echoing our analysis of each stage's contribution.}
  \label{fig:stage_showcase}
\end{figure}

\subsection{Supervised Fine-Tuning (SFT)}
\label{sec:sft}
\paragraph{Distribution Narrowing via High-Quality Alignment.}
While the omni-pre-training stage establishes broad world understanding and mode coverage, the resulting distribution inevitably exhibits high variance, reflecting the noisy nature of web-scale data. Consequently, the primary objective of Supervised Fine-Tuning (SFT) is not merely to correct local artifacts, but to narrow the generation distribution \textbf{towards a focused, high-fidelity sub-manifold}~\cite{flux1kreadev2025}. This phase aims for rapid convergence to a fixed distribution characterized by consistent visual aesthetics and precise instruction following. To achieve this, we transition from the noisy supervision of pre-training to a curriculum dominated by highly curated images filtering by our data infrastructure and \emph{super detailed, grounded captions}. This rigorous supervision acts as an anchor, forcing the model to discard low-quality modes (\eg, unstable stylization or inconsistent rendering) and align strictly with detailed textual descriptions, shifting the model from a diversity-maximizing regime to a quality-maximizing operating point.

\paragraph{Concept Balancing with Tagged Resampling.}
A critical challenge in narrowing the distribution is the risk of catastrophic forgetting, particularly for long-tail concepts that are prone to being overshadowed by dominant modes during convergence. To address this, we enforce strict \textbf{class balancing} throughout the SFT phase. We employ a dynamic resampling strategy guided by world knowledge topological graph in Section~\ref{sec:data-infrastructure}. Specifically, we maintain a target prior over concepts and utilize BM25-based retrieval to compute rarity scores for training samples on the fly. Mini-batches are constructed by up-weighting under-represented concepts -- such as rare entities or specific artistic styles -- while down-weighting over-represented ones. This mechanism ensures that while the model converges to the target high-quality distribution, the marginal distribution over concepts remains uniform, effectively preserving the semantic diversity of the pre-trained model.

\paragraph{Robustness via Model Merging.}
Despite balanced training, SFT on specific high-quality datasets can introduce subtle biases or trade-offs between capabilities (\eg, photorealism vs. stylistic flexibility). To achieve a Pareto-optimal solution without complex inference routing, we employ \textbf{Model Merging}~\cite{wortsman2022model, zhang2025waver} as the final refinement step. We fine-tune multiple SFT variants initialized from the same backbone, each slightly biased towards different capability dimensions (\eg, strict instruction following or aesthetic rendering). We then perform a linear interpolation of their weights in the parameter space: $\theta_{\text{final}} = \sum_{i} \alpha_i \theta_{i}$. This lightweight merging strategy effectively smooths the loss landscape, neutralizing individual biases and resulting in a final model that exhibits superior stability and robustness across diverse prompts compared to any single SFT checkpoint.

\subsection{Few-Step Distillation}
\label{sec:distillation}
The goal of the Few-Step Distillation stage is to reduce the inference time of our foundational SFT model, achieving the efficiency demanded by real-world applications and large-scale deployment.
While our 6B foundational model represents a significant leap in efficiency compared to larger counterparts, the inference cost remains non-negligible. Due to the inherent iterative nature of diffusion models, our standard SFT model requires approximately 100 Number of Function Evaluations (NFEs) to generate high-quality samples using Classifier-Free Guidance (CFG)~\cite{ho2022classifier}. To bridge the gap between generation quality and interactive latency, we implemented a few-step distillation strategy.

Fundamentally, the distillation process involves teaching a student model to mimic the teacher’s denoising dynamics across fewer timesteps along its sampling trajectory.
The core challenge lies in reducing the inherent uncertainty of this trajectory, allowing the student to ``collapse'' its probabilistic path into a deterministic and highly efficient inference process.
Therefore, the key to enable a stable few-step integrator is to meticulously control the distillation process.
We initially selected the Distribution Matching Distillation (DMD)~\cite{yin2024improved,dmd} paradigm due to its promising performance in academic works. However, in practice, we encountered persistent artifacts such as the loss of high-frequency details and noticeable color shifts -- issues that have been increasingly documented by the community. These observations signaled a need for algorithmic refinement. Through a deeper exploration of the distillation mechanism, we gained new insights into the underlying dynamics of DMD, leading to two key technical advancements: \textit{Decoupled DMD}~\cite{liu2025decoupled} and \textit{DMDR}~\cite{jiang2025dmdr}. We refer interested readers to the respective academic papers for full technical details. Below, we introduce the practical application of these techniques in building \textbf{Z-Image-Turbo}.

\begin{figure}[h!]
  \centering
  \includegraphics[width=1\textwidth]{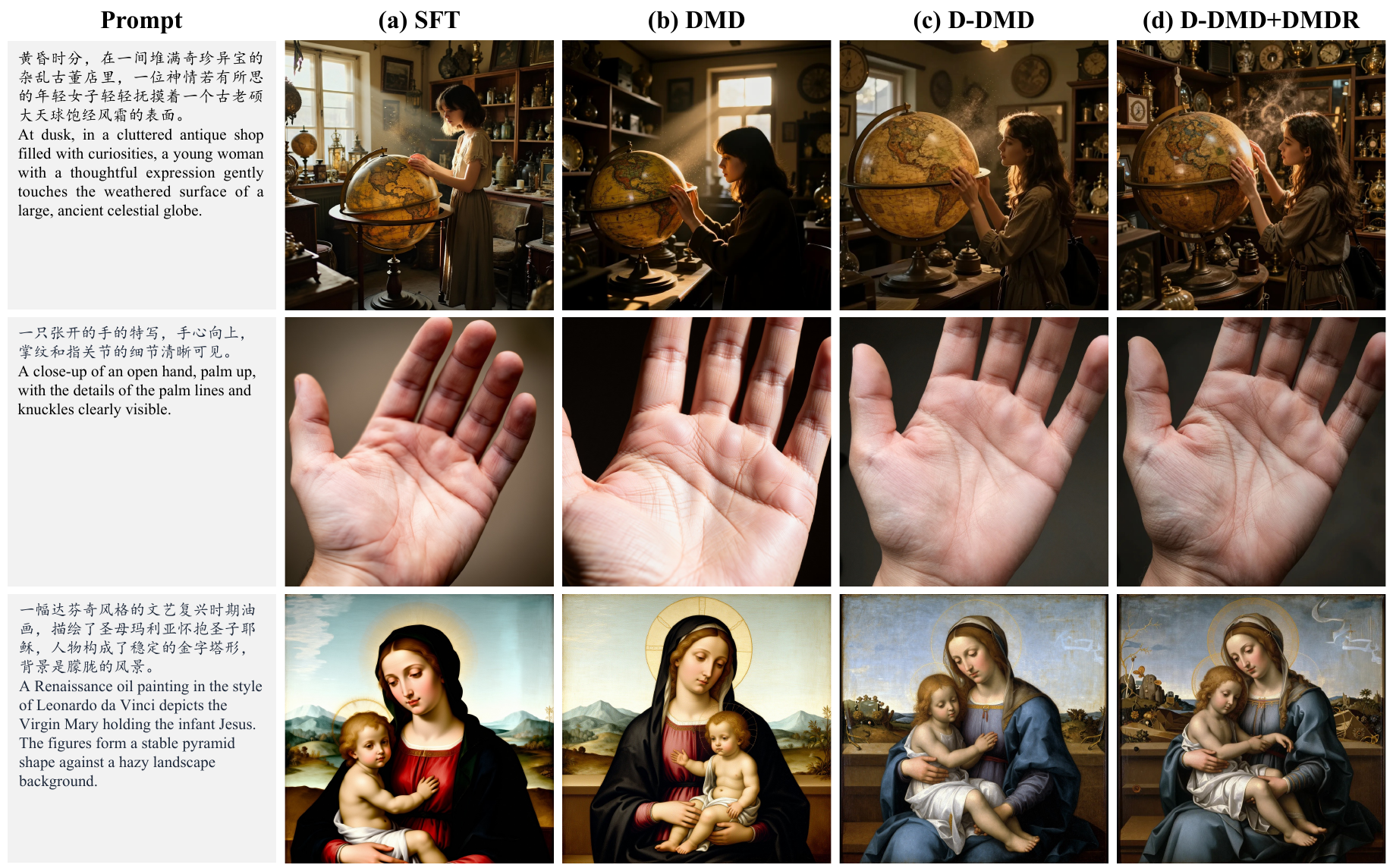}
  \caption{Few-Step Distillation visualization results across different distillation strategies: (a) the original SFT model; (b) Standard DMD; (c) \textit{Decoupled DMD (D-DMD)}; and (d) \textit{D-DMD+DMDR} (\textbf{Z-Image-Turbo}). The proposed approach achieves real-time 8-step inference while attaining superior perceived quality and aesthetic appeal.}
  \label{fig:fsd_res}
\end{figure}

\subsubsection{Decoupled DMD: Resolving Detail and Color Degradation}

Our investigation revealed a core insight: the effectiveness of existing DMD methods is not a monolithic phenomenon but the result of two independent, collaborating mechanisms:

\begin{itemize}
    \item \textbf{CFG-Augmentation (CA):} This acts as the primary engine driving the distillation process, efficiently building up the few-step generation capabilities of the student model. Despite its dominant role, this factor has been largely overlooked in previous literature.
    \item \textbf{Distribution Matching (DM):} This functions primarily as a powerful regularizer, ensuring the stability of the training process and removing the emerging artifacts.
\end{itemize}

By recognizing and decoupling these two mechanisms, we were able to study and optimize them in isolation. This motivation led to the development of \textit{Decoupled DMD}, an improved distillation framework that features a decoupled application of renoising schedules tailored specifically for the CA and DM terms.

In practice, \textit{Decoupled DMD} effectively addresses the pain points of traditional DMD, ensuring sharp detail preservation and color fidelity. Notably, the resulting distilled model not only matches the original multi-step teacher but even surpasses it in terms of photorealism and visual impact.

\subsubsection{DMDR: Enhancing Capacity with RL and Regularization}

To further push the performance boundaries of our few-step model, we incorporate Reinforcement Learning (RL) into the few-step distillation process. Applying RL to generative models typically faces the risk of "reward hacking", where the model exploits the reward function to generate high-scoring but visually nonsensical images. To mitigate this, external regularization is usually required.

Our insight from Decoupled DMD provides a natural solution: since we established that the Distribution Matching (DM) term functions as a high-quality regularizer, it can be organically combined with RL objectives. This synthesis gives rise to \textit{DMDR} (Distribution Matching Distillation meets Reinforcement Learning)~\cite{jiang2025dmdr}. In this framework, RL unlocks the student model's capacity to align with human preferences, while the DM term acts as a robust constraint, effectively preventing reward hacking. This synergy allows \textbf{Z-Image-Turbo} to achieve superior aesthetic alignment and semantic faithfulness while maintaining strict generative stability.

\subsubsection{Results and Analysis}

The efficacy of our \textit{Decoupled DMD} and \textit{DMDR} distillation strategy is visualized in Figure~\ref{fig:fsd_res}. 
The original SFT model (a) sets a high baseline but suffers from high latency. 
Standard DMD (b), while fast, exhibits characteristic degradation: blurred textures and shifted color tones. 
Our \textit{Decoupled DMD} (c) successfully resolves these artifacts, restoring sharp details and accurate colors. 
Finally, \textbf{Z-Image-Turbo} (d), refined via a combination of \textit{Decoupled DMD} and \textit{DMDR}, represents the optimal convergence of speed and quality. It achieves 8-step inference that is not only indistinguishable from the 100-step teacher but frequently surpasses it in perceived quality and aesthetic appeal.
In summary, our Few-Step Distillation framework resolves the long-standing tension between inference speed and visual fidelity.

\subsection{Reinforcement Learning with Human Feedback (RLHF)}
\label{sec:rl}
\begin{figure}[h!]
  \centering
   \vspace{-1em}
  \includegraphics[width=1\textwidth]{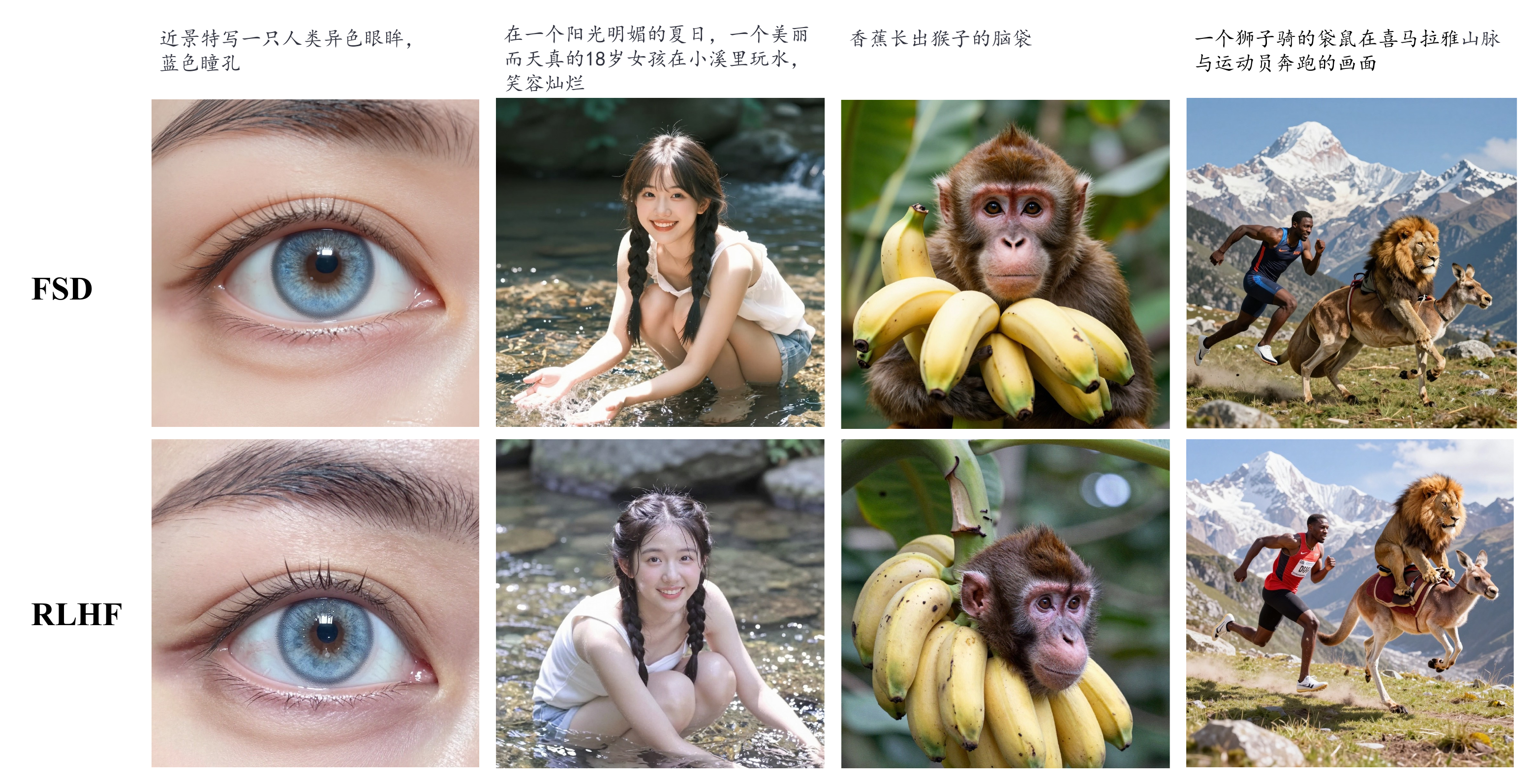}
  \caption{\textbf{Visual comparison between Few-Step Distillation (FSD, top row) and RLHF (bottom row).} Building upon the strong foundation of the FSD model, RLHF further enhances \textbf{photorealism}, \textbf{aesthetic quality}, and \textbf{instruction following}.}
  \label{fig:t2i_rl}
\end{figure}

Following the previous stages, the model has acquired strong foundational capabilities but may still exhibit inconsistencies in aligning with nuanced human preferences. To bridge this gap, we introduce a comprehensive post-training framework leveraging Reinforcement Learning with Human Feedback (RLHF). This framework hinges on a powerful, multi-dimensional reward model, which provides targeted feedback for online optimization. Guided by these feedback signals, our approach is structured into two sequential stages: an initial offline alignment phase using Direct Preference Optimization (DPO)~\cite{rafailov2023direct}, followed by an online refinement phase with Group Relative Policy Optimization (GRPO)~\cite{shao2024deepseekmathpushinglimitsmathematical, flowgrpo}. This two-stage strategy allows us to first efficiently instill robust adherence to objective standards and then leverage the fine-grained signals from our reward model for optimizing more subjective qualities. As illustrated in Figure~\ref{fig:t2i_rl}, this comprehensive process yields substantial improvements in photorealism, aesthetic quality, and instruction following.

\subsubsection{Reward Annotation and Training}

As an indispensable and critical component of the RLHF pipeline, our reward model is designed to evaluate the model’s performance along three key dimensions: instruction-following capability, AI-Content Detection perception, and aesthetic quality. The reward model is then trained specifically to provide targeted feedback along these axes. 
For instruction following, we perform syntactic and semantic decomposition of the prompt into a structured hierarchy that includes (i) core subject entities, (ii) attribute specifications, (iii) action or interaction requirements, (iv) spatial or compositional constraints, and (v) stylistic or rendering conditions. During annotation, human raters simply click on the elements that are not satisfied by the model’s output. We then compute the ratio of satisfied elements to obtain the final instruction-following score, which is used as the target reward.

\subsubsection{Stage 1: Offline Alignment with DPO on Objective Dimensions}
While manually curating preference pairs for DPO is feasible for capturing human aesthetic judgments, scaling this process to a large, high-quality dataset presents a significant bottleneck in real practice. Sourcing consistently informative preference pairs across subjective dimensions (\eg, aesthetics, style) is slow and requires extensive expert annotation. To address this scalability challenge and enhance annotation efficiency, our DPO strategy pivots to focus exclusively on objective, verifiable dimensions.
 
These dimensions, such as \textbf{text rendering} and \textbf{object counting}, offer clear and binary correctness criteria that are highly amenable to automated evaluation by modern Vision-Language Models (VLMs). For instance, given a prompt requiring specific text, an image with accurately rendered characters is designated as the positive sample (`chosen`), while an image with typographical errors becomes the negative sample (`rejected`). We leverage VLMs to programmatically generate a large corpus of such candidate preference pairs. This VLM-generated dataset is then subjected to a streamlined human verification and cleaning process, ensuring high fidelity. This hybrid VLM-human pipeline dramatically increases annotation throughput and consistency compared to purely human manual curation.
 
Furthermore, to smooth the learning curve, we implement a curriculum learning strategy for DPO training. The process begins with prompts of low complexity (\eg, rendering a single word, generating a small number of objects) and progressively advances to more challenging instructions involving multiple elements, complex layouts, or difficult styles. During this process, we also optimized our pair selection strategy. We observed that DPO's convergence is sensitive to the differentiation between positive and negative samples. To maximize training efficiency, our curriculum initially prioritizes pairs with moderate differentiation and gradually introduces more challenging pairs exhibiting larger or more subtle differences, which we found accelerate convergence and improve the final performance.
 
\subsubsection{Stage 2: Online Refinement with GRPO}

Building upon the robust foundation established by DPO, the second stage employs online reinforcement learning with GRPO. Guided by our reward model, this stage is designed to significantly enhance the model's capability for \textbf{photorealistic image generation}, alongside improving \textbf{aesthetic quality} and nuanced \textbf{instruction-following}.

During the GRPO training loop,  we compute a composite advantage function by aggregating the scores from our reward model (\eg, realism, aesthetics, instruction following, etc.). This multi-faceted feedback mechanism enables targeted, fine-grained optimization~\cite{xu2024visionreward}. By providing distinct signals for different aspects of the generation, GRPO can simultaneously enhance photorealistic image generation, aesthetic quality, improve semantic accuracy, and reduce undesirable artifacts. This integrated approach proved to be significantly more effective than optimizing against a single reward, allowing the model to achieve a better balance across multiple, often competing, quality dimensions.

\subsection{Continued Training for Image Editing}
\label{sec:editing}

Starting from the base model, the continued pre-training for image editing consists of two stages, as shown in Figure \ref{fig:model_arch}. In the continued pre-training stage, we train the model with our constructed editing pairs (see Section \ref{sec:editing-data}), together with our text-to-image SFT data to ensure high image quality. We first train the whole amount of editing data in resolution of $512^2$ for a few thousand steps for quick adaptation to editing tasks, and then increase the image resolution to $1024^2$ for high generation quality. Because image editing data pairs are expensive and difficult to acquire, their total volume is significantly smaller and far less diverse than that of text-to-image data. Therefore, we suggest a relatively higher ratio of text-to-image data (\eg, text-to-image:image-to-image $= 4:1$) to avoid performance degradation during training.

In the following SFT stage, a task-balanced, high-quality subset of the training data is manually constructed to further improve the model's  overall performance, especially its instruction following ability. However, synthetic data (\eg, the rendered text data for text editing), though easy-to-acquire and guaranteed to be 100\% accurate in terms of instruction following, are far from the distribution of real-world user input, and are thus heavily downsampled in this final training stage.

\subsection{Prompt Enhancer with Reasoning Chain}
\label{sec:pe}

\begin{figure}[h!]
  \centering
  \vspace{-1em}
  \includegraphics[width=1\textwidth]{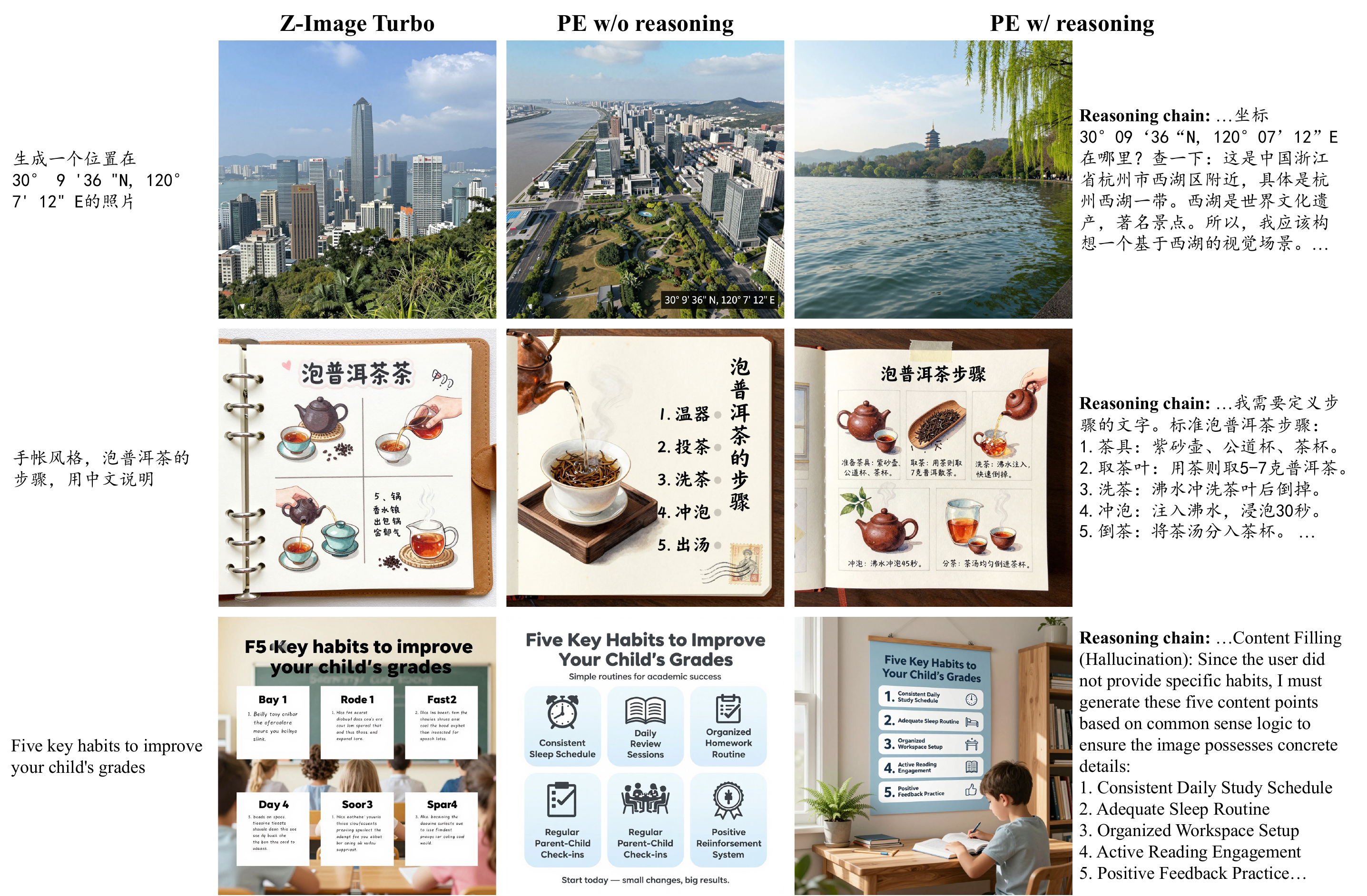}
  \caption{PE visualization. We compare generation results between PE without reasoning (middle column) and PE with reasoning (right column).
As shown in the top row, the reasoning chain enables the model to decipher raw coordinates into a specific scenic context (\eg, West Lake) rather than simply rendering the coordinate text.
In the second row, the reasoning module plans detailed steps for "brewing Pu-erh tea," allowing the model to generate specific illustrations for each step instead of a generic list.
This demonstrates that the reasoning chain effectively injects world knowledge and provides fine-grained content planning for complex user prompts.}
  \label{fig:t2i_pe}
\end{figure}

Due to limited model size (6B parameters), Z-Image exhibits limitations in world knowledge, intent understanding, and complex reasoning. However, it serves as a powerful text decoder capable of translating detailed prompts into realistic images. To address the cognitive gaps, we propose equipping Z-Image with a Prompt Enhancer (PE), powered by system prompt and a pretrained VLM, to improve its reasoning and knowledge capabilities.

Distinct from other methods, we keep the large VLM fixed during alignment. Instead, we process all input prompts (and input images for Z-Image-Edit) through our PE model during the Supervised Fine-Tuning (SFT) stage. This strategy ensures that Z-Image aligns effectively with the Prompt Enhancer during SFT. Furthermore, we identify the structured reasoning chain as a key factor for injecting reasoning and world knowledge. As shown in Figure~\ref{fig:t2i_pe}, without reasoning, the PE merely renders coordinate text onto the image when given geolocation data; with reasoning, it infers the location (\eg, West Lake) to generate the correct scene. Similarly, in generating journal-style instructions, the lack of reasoning leads to monotonous outputs, whereas the reasoning-enhanced model enriches the result by generating specific illustrations for each step.

\section{Performance Evaluation}

\subsection{Human Preference Evaluation}

\subsubsection{Elo-based Human Preference Evaluation}

\begin{figure}[h!]
  \centering
  \includegraphics[width=1\textwidth]{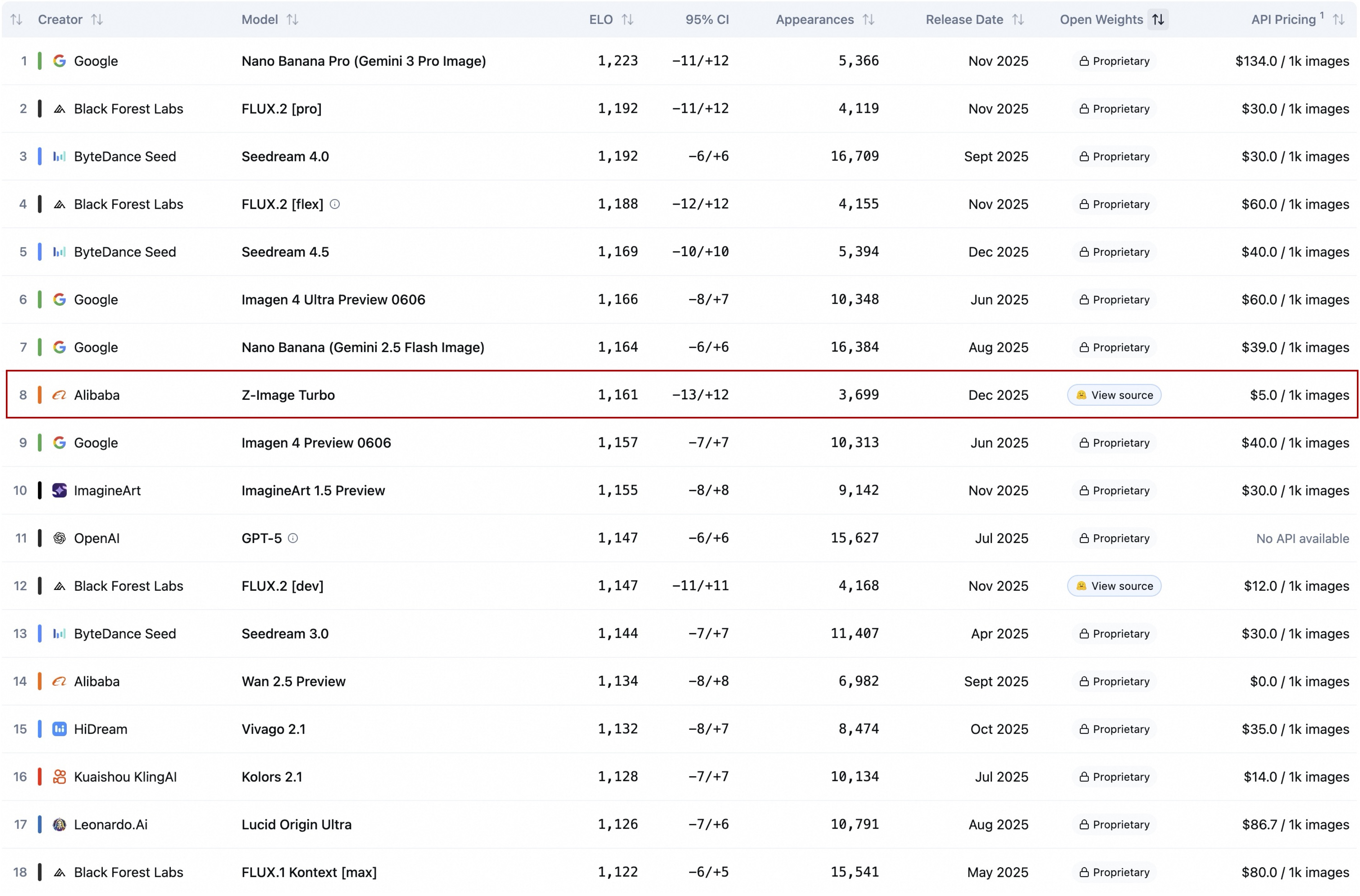}
  \caption{Z-Image-Turbo secured the 8th position among all evaluated models in the Text-to-Image Elo rankings provided by the Artificial Analysis AI Arena.}
  \label{fig:image_arena_all}
\end{figure}

\begin{figure}[h!]
  \centering
  \includegraphics[width=1\textwidth]{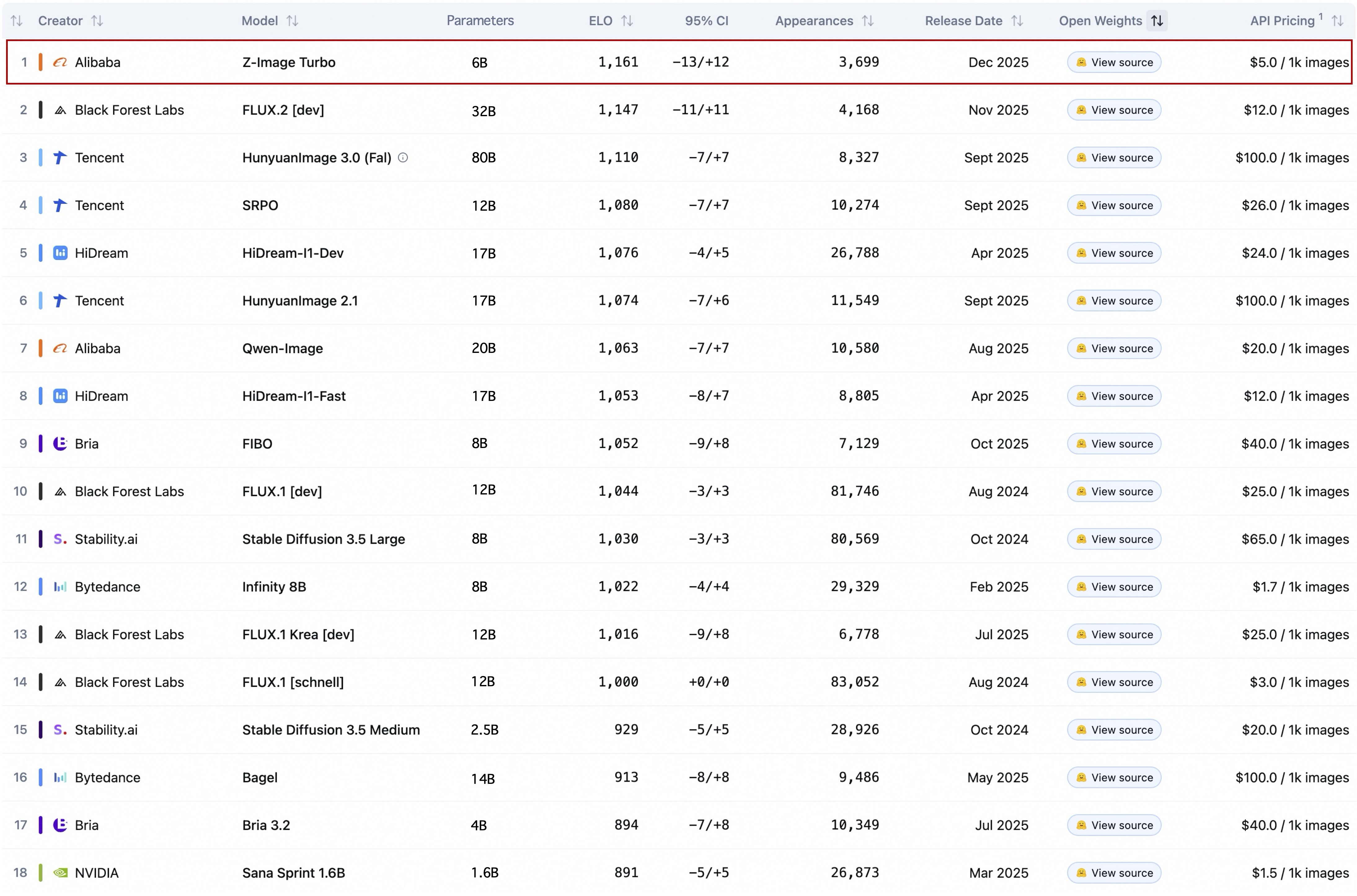}
  \caption{Z-Image-Turbo ranks first among open-source models in the Text-to-Image Elo rankings from the Artificial Analysis AI Arena.}
  \label{fig:image_arena_os}
\end{figure}

To rigorously benchmark Z-Image-Turbo's capabilities against the competitive landscape of generative models, we participated in \textit{Artificial Analysis Image Arena}\footnote{\url{https://artificialanalysis.ai/image/leaderboard/text-to-image}} and \textit{Alibaba AI Arena}\footnote{\url{https://aiarena.alibaba-inc.com/corpora/arena/leaderboard?arenaType=T2I}}, two public-facing, independent benchmarking platforms powered by Elo-based large-scale human judgment. Unlike automated metrics that frequently misalign with human perception, the Elo-based evaluation provides dynamic and unbiased model rankings based on thousands of pairwise comparisons, making it an ideal venue for objective performance assessment.

The evaluation protocol is built upon the Elo rating system  --  a well-established method for ranking competitors based on head-to-head outcomes. In each round, two images generated from the same text prompt by different models are displayed side-by-side with identities hidden. Evaluators are asked to select the image they perceive as superior in terms of visual coherence, detail rendering, prompt alignment, and artistic quality. Each vote updates the global Elo leaderboard dynamically, ensuring that rankings reflect collective human judgment over time.

In the evaluation, Z-Image-Turbo -- our high-efficiency diffusion architecture with 6B parameters and a low inference cost of 8 NFEs -- was benchmarked against other leading models. These included top-tier closed-source systems such as Nano Banana Pro~\cite{nanopro}, Imagen 4 Ultra Preview 0606~\cite{google2025imagen4} (Google), Flux.2 (Black Forest Labs)~\cite{labs2025flux}, Seedream (ByteDance)~\cite{seedream2025seedream}, GPT Image 1 [High] (OpenAI)~\cite{openai2025gptimage}, and FLUX.1 Kontext [Pro] (Black Forest Labs)~\cite{labs2025flux}, as well as open-source baselines like HunyuanImage 3.0~\cite{cao2025hunyuanimage} and Qwen-Image~\cite{qwenimage}.

According to the Artificial Analysis leaderboard in Figure~\ref{fig:image_arena_all} and Figure~\ref{fig:image_arena_os}, Z-Image-Turbo achieved an Elo score of 1,161, ranking 8th overall. This result is notable for two primary reasons. Firstly, Z-Image-Turbo is established as the top-ranked open-source model on this highly competitive platform, with performance comparable to that of leading proprietary systems like Google's Imagen 4 and ByteDance's Seedream. Secondly, the model exhibits exceptional efficiency: among the top ten models, it features not only the smallest parameter count (6B) but also the lowest inference cost (\$5.0 per 1,000 images). This unique combination of high-quality output and low computational overhead is a defining characteristic of our architecture. These findings are further corroborated by its performance on the Alibaba AI Arena (see Table~\ref{tab:elo}), where Z-Image-Turbo secures a higher rank of 4th overall while again leading the open-source category.


In summary, these results establish Z-Image-Turbo as one of the leading open text-to-image models in terms of both quality and efficiency. More than a high-performing generator, it represents a new baseline for efficiency-oriented architecture design, demonstrating that compact models can achieve elite-level performance without compromising usability. This combination of speed, fidelity, and openness enables deployment in resource-constrained environments, interactive applications, and community-driven innovation.

\begin{table}[h!]
\centering
\caption{Elo rankings of Text-to-Image models from \textit{Alibaba AI Arena}. Z-Image-Turbo achieves 4th globally and 1st among open-source models.}
\label{tab:elo}
\begin{adjustbox}{width=\textwidth}
\begin{tabular}{clccccr}
\toprule
\textbf{Rank} & \textbf{Model Name} & \textbf{Company} & \textbf{Type} & \textbf{95\% CI} & \textbf{Elo Score} & \textbf{Win Rate} \\
\midrule
1 & Imagen 4 Ultra Preview 0606 & Google & Closed-source & -16/+16 & 1048 & 48\% \\
2 & gemini-2.5-flash-image-preview & Google & Closed-source & -16/+14 & 1046 & 47\% \\
3 & Seedream 4.0 & ByteDance & Closed-source & -17/+16 & 1039 & 46\% \\
4 & \textbf{Z-Image-Turbo} & Alibaba & Open-source (6B) & -15/+17 & 1025 & 45\% \\
5 & Seedream 3.0 & ByteDance & Closed-source & -15/+19 & 1012 & 41\% \\
6 & Qwen-Image & Alibaba & Open-source (20B) & -16/+16 & 1008 & 41\% \\
7 & GPT Image 1 & OpenAI & Closed-source & -14/+17 & 986 & 38\% \\
8 & FLUX.1 Kontext Pro & Black Forest Labs & Closed-source & -15/+14 & 950 & 32\% \\
9 & Ideogram 3.0 & Ideogram & Closed-source & -15/+16 & 936 & 29\% \\
\bottomrule
\end{tabular}
\end{adjustbox}
\end{table}

\subsubsection{Human Preference Comparison with Flux 2 dev}

To benchmark our Z-Image model against the \textbf{latest} state-of-the-art open-source model, Flux 2 dev~\citep{flux-2-2025}, we conducted an additional user study. The study involved evaluating a set of 222 samples generated by Z-Image and Flux 2 dev with user-style prompts. Each sample was assessed by three independent annotators to ensure robustness and reduce bias in the evaluation process. The results, as summarized in Table \ref{tab:probability_stats_en}, demonstrate a significant advantage for Z-Image model. Specifically, the "G Rate" (\textit{i.e.}, Good Rate) achieve 46.4\%, indicating a high proportion of satisfactory generations. Concurrently, the ``S Rate'' (\textit{i.e.}, Same Rate) is 41.0\%. Consequently, the combined ``G+S Rate'' reached an impressive 87.4\%, suggesting that Z-Image consistently produces high-quality or acceptable outputs from user prompts. In contrast, the ``B Rate'' (\textit{i.e.}, Bad Rate) was only 12.6\%. Remarkably, these superior results were achieved with Z-Image having only 1/5 of the parameters compared to Flux 2 dev (6B vs. 32B parameters).

\begin{table}[H] 
\centering
\caption{User Preference Evaluation Results: G, S, B, and G+S Rates for Z-Image vs. Flux 2 dev Comparison}
\label{tab:probability_stats_en}
\begin{adjustbox}{width=0.4\textwidth,center}
\begin{tabular}{cccc}
\toprule
\textbf{G Rate} & \textbf{S Rate} & \textbf{B Rate} & \textbf{G+S Rate} \\
\midrule
46.4\% & 41.0\% & 12.6\% & 87.4\% \\
\bottomrule
\end{tabular}
\end{adjustbox}
\end{table}

\subsection{Quantitative Evaluation}
To comprehensively evaluate the generation and editing capabilities of \textbf{Z-Image} and its variants (\textbf{Z-Image-Turbo} and \textbf{Z-Image-Edit}), we conducted extensive experiments across multiple authoritative benchmarks. These evaluations cover general image generation, fine-grained instruction following, text rendering in both English and Chinese, and instruction-based image editing.

\subsubsection{Text-to-Image Generation}

\paragraph{CVTG-2K.} To evaluate our model's performance on text rendering tasks, we conduct quantitative experiments on the CVTG-2K benchmark~\cite{du2025textcrafter}. CVTG-2K is a specialized benchmark designed for Complex Visual Text Generation, encompassing diverse scenarios with varying numbers of text regions.
As presented in Table~\ref{tab:cvtg_2k_bench_results}, our model achieves superior performance on CVTG-2K across all evaluation metrics. Specifically, Z-Image attains the highest average Word Accuracy score of 0.8671, outperforming competitive baselines such as GPT-Image-1 ~\cite{openai2025gptimage} (0.8569) and Qwen-Image ~\cite{qwenimage} (0.8288). Notably, our model demonstrates robust performance across varying levels of complexity, maintaining consistent accuracy even as the number of text regions increases from 2 to 5. Furthermore, Z-Image-Turbo, our efficient variant, achieves the highest CLIP Score of 0.8048 among all models while maintaining competitive text accuracy (0.8585 average Word Accuracy), striking an effective balance between generation quality and inference efficiency. These results demonstrate the effectiveness of our approach in complex visual text generation scenarios.

\begin{table}[h!]
\centering
\caption{Quantitative evaluation results of English text rendering on CVTG-2K~\cite{du2025textcrafter}.}
\label{tab:cvtg_2k_bench_results}
\begin{adjustbox}{width=0.95\textwidth}
\begin{tabular}{c l | c c ccccc}
\toprule
\multirow{2}{*}{\textbf{Rank}} & \multirow{2}{*}{\textbf{Model}} & \multirow{2}{*}{\textbf{NED}} & \multirow{2}{*}{\textbf{CLIPScore}} & \multicolumn{5}{c}{\textbf{Word Accuracy}} \\
\cmidrule(lr){5-9}
& & & & 2 regions & 3 regions & 4 regions & 5 regions & average$\uparrow$ \\
\midrule
1 & \textbf{Z-Image} & 0.9367 & 0.7969 & \textbf{0.9006} & \textbf{0.8722} & 0.8652 & \textbf{0.8512} & \textbf{0.8671} \\
2 & \textbf{Z-Image-Turbo} & 0.9281 & \textbf{0.8048} & 0.8872 & 0.8662 & 0.8628 & 0.8347 & 0.8585 \\
3 & GPT Image 1 [High]~\cite{openai2025gptimage} & \textbf{0.9478} & 0.7982 & 0.8779 & 0.8659 & \textbf{0.8731} & 0.8218 & 0.8569 \\
4 & Qwen-Image~\cite{qwenimage} & 0.9116 & 0.8017 & 0.8370 & 0.8364 & 0.8313 & 0.8158 & 0.8288 \\
5 & TextCrafter~\cite{du2025textcrafter} & 0.8679 & 0.7868 & 0.7628 & 0.7628 & 0.7406 & 0.6977 & 0.7370 \\
6 & SD3.5 Large~\cite{esser2024scaling} & 0.8470 & 0.7797 & 0.7293 & 0.6825 & 0.6574 & 0.5940 & 0.6548 \\
7 & Seedream 3.0~\cite{gao2025seedream} & 0.8537 & 0.7821 & 0.6282 & 0.5962 & 0.6043 & 0.5610 & 0.5924 \\
8 & FLUX.1 [dev]~\cite{labs2025flux} & 0.6879 & 0.7401 & 0.6089 & 0.5531 & 0.4661 & 0.4316 & 0.4965 \\
9 & 3DIS~\cite{zhou20253disflux} & 0.6505 & 0.7767 & 0.4495 & 0.3959 & 0.3880 & 0.3303 & 0.3813 \\
10 & RAG-Diffusion~\cite{li2025ragdiffusion} & 0.4498 & 0.7797 & 0.4388 & 0.3316 & 0.2116 & 0.1910 & 0.2648 \\
11 & TextDiffuser-2~\cite{chen2024textdiffuser} & 0.4353 & 0.6765 & 0.5322 & 0.3255 & 0.1787 & 0.0809 & 0.2326 \\
12 & AnyText~\cite{tuo2023anytext} & 0.4675 & 0.7432 & 0.0513 & 0.1739 & 0.1948 & 0.2249 & 0.1804 \\
\bottomrule
\end{tabular}
\end{adjustbox}
\end{table}

\paragraph{LongText-Bench.} To further assess our model's capability in rendering longer texts, we evaluate its performance on LongText-Bench~\cite{geng2025xomni}, a specialized benchmark focusing on evaluating the performance of rendering longer texts in both English and Chinese.
As shown in Table~\ref{tab:longtext_bench_results}, our models demonstrate strong and consistent performance across both language settings. On LongText-Bench-EN, Z-Image achieves a competitive score of 0.935, ranking third among all evaluated models, while on LongText-Bench-ZH, it attains a score of 0.936, securing second place. Z-Image-Turbo also delivers impressive results, scoring 0.917 on the English benchmark and 0.926 on the Chinese benchmark, demonstrating strong efficiency-performance trade-offs. This consistent performance across both languages highlights our model's robust bilingual text rendering capability.

\begin{table}[h!]
\centering
\caption{Quantitative evaluation results on LongText-Bench~\cite{geng2025xomni}.}
\label{tab:longtext_bench_results}
\begin{adjustbox}{width=0.70\textwidth}
\begin{tabular}{c l | c c}
\toprule
\textbf{Rank} & \textbf{Model} & \textbf{LongText-Bench-EN}$\uparrow$ & \textbf{LongText-Bench-ZH$\uparrow$} \\
\midrule
1 & Qwen-Image~\cite{qwenimage} & 0.943 & \textbf{0.946} \\
2 & \textbf{Z-Image} & 0.935 & 0.936 \\
3 & \textbf{Z-Image-Turbo} & 0.917 & 0.926 \\
4 & Seedream 3.0~\cite{gao2025seedream} & 0.896 & 0.878 \\
5 & X-Omni~\cite{geng2025xomni} & 0.900 & 0.814 \\
6 & GPT Image 1 [High]~\cite{openai2025gptimage} & \textbf{0.956} & 0.619 \\
7 & Kolors 2.0~\cite{kuaishou2025kolors} & 0.258 & 0.329 \\
8 & BAGEL~\cite{deng2505emerging} & 0.373 & 0.310 \\
9 & OmniGen2~\cite{wu2025omnigen2} & 0.561 & 0.059 \\
10 & HiDream-I1-Full~\cite{cai2025hidream} & 0.543 & 0.024 \\
11 & BLIP3-o~\cite{chen2025blip3} & 0.021 & 0.018 \\
12 & Janus-Pro~\cite{chen2025januspro} & 0.019 & 0.006 \\
13 & FLUX.1 [Dev]~\cite{labs2025flux} & 0.607 & 0.005 \\
\bottomrule
\end{tabular}
\end{adjustbox}
\end{table}

\paragraph{OneIG.} We utilize the OneIG benchmark~\cite{chang2025oneig} to assess fine-grained alignment. 
As reported in Tables~\ref{tab:oneig_en_results} and \ref{tab:oneig_zh_results}, Z-Image achieves the highest overall score (0.546) on the English track, surpassing Qwen-Image (0.539) and GPT Image 1 [High] (0.533). 
Notably, Z-Image sets a new state-of-the-art in text rendering reliability with an English \textit{Text} score of 0.987 and a Chinese \textit{Text} score of 0.988, significantly outperforming competitors. 
On the Chinese track, Z-Image ranks second overall (0.535), confirming its multi-lingual robustness.
Additionally, our distilled version, Z-Image-Turbo, demonstrates impressive efficiency, maintaining strong performance with only a marginal decrease compared to the base model.

\begin{table}[h!]
\centering
\vspace{-1em}
\caption{Quantitative evaluation results on OneIG-EN~\cite{chang2025oneig}. The overall score is the average of the five dimensions.}
\label{tab:oneig_en_results}
\begin{adjustbox}{width=0.85\textwidth}
\begin{tabular}{cl|ccccc|c}
\toprule
\textbf{Rank} & \textbf{Model} & \textbf{Alignment} & \textbf{Text} & \textbf{Reasoning} & \textbf{Style} & \textbf{Diversity} & \textbf{Overall}$\uparrow$ \\
\midrule
1 & \textbf{Z-Image} & 0.881 & 0.987 & 0.280 & 0.387 & 0.194 & \textbf{0.546} \\
2 & Qwen-Image~\cite{qwenimage} & \textbf{0.882} & 0.891 & 0.306 & 0.418 & 0.197 & 0.539 \\
3 & GPT Image 1 [High]~\cite{openai2025gptimage} & 0.851 & 0.857 & \textbf{0.345} & \textbf{0.462} & 0.151 & 0.533 \\
4 & Seedream 3.0~\cite{gao2025seedream} & 0.818 & 0.865 & 0.275 & 0.413 & 0.277 & 0.530 \\
5 & \textbf{Z-Image-Turbo} & 0.840 & \textbf{0.994} & 0.298 & 0.368 & 0.139 & 0.528 \\
6 & Imagen 4~\cite{google2025imagen4} & 0.857 & 0.805 & 0.338 & 0.377 & 0.199 & 0.515 \\
7 & Recraft V3~\cite{recraftv3} & 0.810 & 0.795 & 0.323 & 0.378 & 0.205 & 0.502 \\
8 & HiDream-I1-Full~\cite{cai2025hidream} & 0.829 & 0.707 & 0.317 & 0.347 & 0.186 & 0.477 \\
9 & OmniGen2~\cite{wu2025omnigen2} & 0.804 & 0.680 & 0.271 & 0.377 & 0.242 & 0.475 \\
10 & SD3.5 Large~\cite{esser2024scaling} & 0.809 & 0.629 & 0.294 & 0.353 & 0.225 & 0.462 \\
11 & CogView4~\cite{zheng2024cogview3} & 0.786 & 0.641 & 0.246 & 0.353 & 0.205 & 0.446 \\
12 & FLUX.1 [Dev]~\cite{labs2025flux} & 0.786 & 0.523 & 0.253 & 0.368 & 0.238 & 0.434 \\
13 & Kolors 2.0~\cite{kuaishou2025kolors} & 0.820 & 0.427 & 0.262 & 0.360 & 0.300 & 0.434 \\
14 & Imagen 3~\cite{baldridge2024imagen} & 0.843 & 0.343 & 0.313 & 0.359 & 0.188 & 0.409 \\
15 & BAGEL~\cite{deng2505emerging} & 0.769 & 0.244 & 0.173 & 0.367 & 0.251 & 0.361 \\
16 & Lumina-Image 2.0~\cite{qin2025lumina} & 0.819 & 0.106 & 0.270 & 0.354 & 0.216 & 0.353 \\
17 & SANA-1.5-4.8B~\cite{xiesana1.5} & 0.765 & 0.069 & 0.217 & 0.401 & 0.216 & 0.334 \\
18 & SANA-1.5-1.6B~\cite{xiesana1.5} & 0.762 & 0.054 & 0.209 & 0.387 & 0.222 & 0.327 \\
19 & BAGEL+CoT~\cite{deng2505emerging} & 0.793 & 0.020 & 0.206 & 0.390 & 0.209 & 0.324 \\
20 & SD 1.5~\cite{rombach2022high} & 0.565 & 0.010 & 0.207 & 0.383 & \textbf{0.429} & 0.319 \\
21 & SDXL~\cite{podell2023sdxl} & 0.688 & 0.029 & 0.237 & 0.332 & 0.296 & 0.316 \\
22 & Show-o2-7B~\cite{xie2025show} & 0.817 & 0.002 & 0.226 & 0.317 & 0.177 & 0.308 \\
23 & BLIP3-o~\cite{chen2025blip3} & 0.711 & 0.013 & 0.223 & 0.361 & 0.229 & 0.307 \\
24 & Show-o2-1.5B~\cite{xie2025show} & 0.798 & 0.002 & 0.219 & 0.317 & 0.186 & 0.304 \\
25 & Janus-Pro~\cite{chen2025januspro} & 0.553 & 0.001 & 0.139 & 0.276 & 0.365 & 0.267 \\
\bottomrule
\end{tabular}
\end{adjustbox}
\end{table}

\begin{table}[h!]
\centering
\caption{Quantitative evaluation results on OneIG-ZH~\cite{chang2025oneig}. The overall score is the average of the five dimensions.}
\label{tab:oneig_zh_results}
\begin{adjustbox}{width=0.85\textwidth}
\begin{tabular}{cl|ccccc|c}
\toprule
\textbf{Rank} & \textbf{Model} & \textbf{Alignment} & \textbf{Text} & \textbf{Reasoning} & \textbf{Style} & \textbf{Diversity} & \textbf{Overall}$\uparrow$ \\
\midrule
1 & Qwen-Image~\cite{qwenimage} & \textbf{0.825} & 0.963 & 0.267 & 0.405 & 0.279 & \textbf{0.548} \\
2 & \textbf{Z-Image} & 0.793 & \textbf{0.988} & 0.266 & 0.386 & 0.243 & 0.535 \\
3 & Seedream 3.0~\cite{gao2025seedream} & 0.793 & 0.928 & 0.281 & 0.397 & 0.243 & 0.528 \\
4 & \textbf{Z-Image-Turbo} & 0.782 & 0.982 & 0.276 & 0.361 & 0.134 & 0.507 \\
5 & GPT Image 1 [High]~\cite{openai2025gptimage} & 0.812 & 0.650 & \textbf{0.300} & \textbf{0.449} & 0.159 & 0.474 \\
6 & Kolors 2.0~\cite{kuaishou2025kolors} & 0.738 & 0.502 & 0.226 & 0.331 & 0.333 & 0.426 \\
7 & BAGEL~\cite{deng2505emerging} & 0.672 & 0.365 & 0.186 & 0.357 & 0.268 & 0.370 \\
8 & Cogview4~\cite{zheng2024cogview3} & 0.700 & 0.193 & 0.236 & 0.348 & 0.214 & 0.338 \\
9 & HiDream-I1-Full~\cite{cai2025hidream} & 0.620 & 0.205 & 0.256 & 0.304 & 0.300 & 0.337 \\
10 & Lumina-Image 2.0~\cite{qin2025lumina} & 0.731 & 0.136 & 0.221 & 0.343 & 0.240 & 0.334 \\
11 & BAGEL+CoT~\cite{deng2505emerging} & 0.719 & 0.127 & 0.219 & 0.385 & 0.197 & 0.329 \\
12 & BLIP3-o~\cite{chen2025blip3} & 0.608 & 0.092 & 0.213 & 0.369 & 0.233 & 0.303 \\
13 & Janus-Pro~\cite{chen2025januspro} & 0.324 & 0.148 & 0.104 & 0.264 & \textbf{0.358} & 0.240 \\
\bottomrule
\end{tabular}
\end{adjustbox}
\end{table}

\paragraph{GenEval.} As shown in Table~\ref{tab:geneval_results}, we evaluate object-centric generation using GenEval~\cite{ghosh2023geneval}. Z-Image achieves an overall score of 0.84, securing a three-way tie for second place alongside Seedream 3.0~\cite{gao2025seedream} and GPT Image 1 [High]~\cite{openai2025gptimage}, trailing only Qwen-Image~\cite{qwenimage} (0.87).
Notably, Z-Image-Turbo delivers highly competitive performance with an overall score of 0.82, maintaining only a 2-point gap from the base model. These results indicate that our model possesses a robust capability for generating accurate and distinct entities.

\begin{table}[h!]
\centering
\caption{Quantitative Evaluation results on GenEval~\cite{ghosh2023geneval}.}
\label{tab:geneval_results}
\begin{adjustbox}{width=\textwidth}
\begin{tabular}{cl|cccccc|c}
\toprule
\textbf{Rank} & \textbf{Model} & \textbf{Single Object} & \textbf{Two Object} & \textbf{Counting} & \textbf{Colors} & \textbf{Position} & \textbf{Attribute Binding} & \textbf{Overall}$\uparrow$\\
\midrule
1 & Qwen-Image~\cite{qwenimage} & 0.99 & 0.92 & 0.89 & 0.88 & 0.76 & 0.77 & \textbf{0.87} \\
2 & \textbf{Z-Image} & \textbf{1.00} & 0.94 & 0.78 & \textbf{0.93} & 0.62 & 0.77 & 0.84 \\
2 & Seedream 3.0~\cite{gao2025seedream} & 0.99 & 0.96 & \textbf{0.91} & \textbf{0.93} & 0.47 & \textbf{0.80} & 0.84 \\
2 & GPT Image 1 [High]~\cite{openai2025gptimage} & 0.99 & 0.92 & 0.85 & 0.92 & 0.75 & 0.61 & 0.84 \\
5 & HiDream-I1-Full~\cite{cai2025hidream} & \textbf{1.00} & \textbf{0.98} & 0.79 & 0.91 & 0.60 & 0.72 & 0.83 \\
6 & \textbf{Z-Image-Turbo} & \textbf{1.00} & 0.95 & 0.77 & 0.89 & 0.65 & 0.68 & 0.82 \\
7 & Janus-Pro-7B~\cite{chen2025januspro} & 0.99 & 0.89 & 0.59 & 0.90 & \textbf{0.79} & 0.66 & 0.80 \\
8 & Lumina-Image 2.0~\cite{qin2025lumina} & - & 0.87 & 0.67 & - & - & 0.62 & 0.73 \\
9 & SD3.5-Large~\cite{esser2024scaling} & 0.98 & 0.89 & 0.73 & 0.83 & 0.34 & 0.47 & 0.71 \\
10 & FLUX.1 [Dev]~\cite{labs2025flux} & 0.98 & 0.81 & 0.74 & 0.79 & 0.22 & 0.45 & 0.66 \\
11 & JanusFlow~\cite{ma2024janusflow} & 0.97 & 0.59 & 0.45 & 0.83 & 0.53 & 0.42 & 0.63 \\
12 & SD3 Medium~\cite{esser2024scaling} & 0.98 & 0.74 & 0.63 & 0.67 & 0.34 & 0.36 & 0.62 \\
13 & Emu3-Gen~\cite{wang2024emu3} & 0.98 & 0.71 & 0.34 & 0.81 & 0.17 & 0.21 & 0.54 \\
14 & Show-o~\cite{xie2024show} & 0.95 & 0.52 & 0.49 & 0.82 & 0.11 & 0.28 & 0.53 \\
15 & PixArt-$\alpha$~\cite{chen2023pixart} & 0.98 & 0.50 & 0.44 & 0.80 & 0.08 & 0.07 & 0.48 \\
\bottomrule
\end{tabular}
\end{adjustbox}
\end{table}

\paragraph{DPG-Bench.} Table~\ref{tab:dpg_results} presents the  comparison on the DPG-Bench benchmark~\cite{hu2024ella}, which evaluates the ability of prompt following in dense prompts. 
Z-Image achieves a strong global performance, ranking third overall with a score of 88.14, closely trailing Seedream 3.0~\cite{gao2025seedream} and Qwen-Image~\cite{qwenimage}. Notably, our model demonstrates robust performance in the \textit{Attribute} dimension (93.16), surpassing the leading Qwen-Image (92.02) and Seedream 3.0 (91.36). 
Furthermore, our 8-step distillation model (Z-Image-Turbo), maintains competitive performance while achieving high efficiency.

\begin{table}[h!]
\centering
\caption{Quantitative evaluation results on DPG~\cite{hu2024ella}.}
\label{tab:dpg_results}
\begin{adjustbox}{width=0.85\textwidth}
\begin{tabular}{cl|ccccc|c}
\toprule
\textbf{Rank} & \textbf{Model} & \textbf{Global} & \textbf{Entity} & \textbf{Attribute} & \textbf{Relation} & \textbf{Other} & \textbf{Overall}$\uparrow$ \\
\midrule
1  & Qwen-Image~\cite{qwenimage} & 91.32 & 91.56 & 92.02 & 94.31 & \textbf{92.73} & \textbf{88.32} \\
2  & Seedream 3.0~\cite{gao2025seedream} & \textbf{94.31} & \textbf{92.65} & 91.36 & 92.78 & 88.24 & 88.27 \\
3  & \textbf{Z-Image} & 93.39 & 91.22 & \textbf{93.16} & 92.22 & 91.52 & 88.14 \\
4  & Lumina-Image 2.0~\cite{qin2025lumina} & - & 91.97 & 90.20 & \textbf{94.85} & - & 87.20 \\
5  & HiDream-I1-Full~\cite{cai2025hidream} & 76.44 & 90.22 & 89.48 & 93.74 & 91.83 & 85.89 \\
6  & GPT Image 1 [High]~\cite{openai2025gptimage} & 88.89 & 88.94 & 89.84 & 92.63 & 90.96 & 85.15 \\
7  & \textbf{Z-Image-Turbo} & 91.29 & 89.59 & 90.14 & 92.16 & 88.68 & 84.86 \\
8  & Janus-Pro-7B~\cite{chen2025januspro} & 86.90 & 88.90 & 89.40 & 89.32 & 89.48 & 84.19 \\
9  & SD3 Medium~\cite{esser2024scaling} & 87.90 & 91.01 & 88.83 & 80.70 & 88.68 & 84.08 \\
10 & FLUX.1 [Dev]~\cite{labs2025flux} & 74.35 & 90.00 & 88.96 & 90.87 & 88.33 & 83.84 \\
11 & DALL-E 3~\cite{betker2023improving} & 90.97 & 89.61 & 88.39 & 90.58 & 89.83 & 83.50 \\
12 & Janus-Pro-1B~\cite{chen2025januspro} & 87.58 & 88.63 & 88.17 & 88.98 & 88.30 & 82.63 \\
13 & Emu3-Gen~\cite{wang2024emu3} & 85.21 & 86.68 & 86.84 & 90.22 & 83.15 & 80.60 \\
14 & PixArt-$\Sigma$~\cite{chen2024pixart} & 86.89 & 82.89 & 88.94 & 86.59 & 87.68 & 80.54 \\
15 & Janus~\cite{wu2025janus} & 82.33 & 87.38 & 87.70 & 85.46 & 86.41 & 79.68 \\
16 & Hunyuan-DiT~\cite{li2024hunyuandit} & 84.59 & 80.59 & 88.01 & 74.36 & 86.41 & 78.87 \\
17 & Playground v2.5~\cite{li2024playground} & 83.06 & 82.59 & 81.20 & 84.08 & 83.50 & 75.47 \\
18 & SDXL~\cite{podell2023sdxl} & 83.27 & 82.43 & 80.91 & 86.76 & 80.41 & 74.65 \\
19 & Lumina-Next~\cite{gao2024lumina-next} & 82.82 & 88.65 & 86.44 & 80.53 & 81.82 & 74.63 \\
20 & PixArt-$\alpha$~\cite{chen2023pixart} & 74.97 & 79.32 & 78.60 & 82.57 & 76.96 & 71.11 \\
21 & SD1.5~\cite{rombach2022high} & 74.63 & 74.23 & 75.39 & 73.49 & 67.81 & 63.18 \\
\bottomrule
\end{tabular}
\end{adjustbox}
\end{table}

\paragraph{TIIF.} Table~\ref{tab:tiif_bench_results} details the performance on the TIIF benchmark testmini~\cite{wei2025tiif}, which systematically evaluates instruction-following capabilities. Z-Image and Z-Image-Turbo achieve the 4th and 5th ranks, respectively. 
These results demonstrate that both the base and distilled versions possess exceptional capabilities in interpreting and executing complex user instructions across diverse categories.

\begin{table}[h!]
\centering
\caption{Quantitative evaluation results on TIIF Bench testmini~\cite{wei2025tiif}.}
\label{tab:tiif_bench_results}
\begin{adjustbox}{width=\textwidth}
\begin{tabular}{c l | cc | cc cc cc cc cc cc cc cc cc cc cc}
\toprule
\multirow{3}{*}{\textbf{Rank}} & \multirow{3}{*}{\textbf{Model}} & \multicolumn{2}{c|}{\multirow{2}{*}{\textbf{Overall}$\uparrow$}} & \multicolumn{8}{c}{\textbf{Basic Following}} & \multicolumn{12}{c}{\textbf{Advanced Following}} & \multicolumn{2}{c}{\textbf{Designer}} \\
\cmidrule(lr){5-12} \cmidrule(lr){13-24} \cmidrule(lr){25-26}
& & & & \multicolumn{2}{c}{Avg} & \multicolumn{2}{c}{Attribute} & \multicolumn{2}{c}{Relation} & \multicolumn{2}{c}{Reasoning} & \multicolumn{2}{c}{Avg} & \multicolumn{2}{c}{Attr.+Rela.} & \multicolumn{2}{c}{Attr.+Reas.} & \multicolumn{2}{c}{Rela.+Reas.} & \multicolumn{2}{c}{Style} & \multicolumn{2}{c}{Text} & \multicolumn{2}{c}{Real World} \\
& & short & long & short & long & short & long & short & long & short & long & short & long & short & long & short & long & short & long & short & long & short & long & short & long \\
\midrule
1 & GPT Image 1 [High]~\cite{openai2025gptimage} & \textbf{89.15} & \textbf{88.29} & \textbf{90.75} & \textbf{89.66} & \textbf{91.33} & 87.08 & 84.57 & 84.57 & \textbf{96.32} & \textbf{97.32} & \textbf{88.55} & \textbf{88.35} & \textbf{87.07} & \textbf{89.44} & \textbf{87.22} & \textbf{83.96} & \textbf{85.59} & \textbf{83.21} & 90.00 & 93.33 & 89.83 & 86.83 & 89.73 & \textbf{93.46} \\
2 & Qwen-Image~\cite{qwenimage} & 86.14 & 86.83 & 90.18 & 87.22 & 90.50 & \textbf{91.50} & 88.22 & \textbf{90.78} & 79.81 & 79.38 & 79.30 & 80.88 & 79.21 & 78.94 & 78.85 & 81.69 & 75.57 & 78.59 & \textbf{100.00} & \textbf{100.00} & 92.76 & 89.14 & \textbf{90.30} & 91.42 \\
3 & Seedream 3.0~\cite{gao2025seedream} & 86.02 & 84.31 & 87.07 & 84.93 & 90.50 & 90.00 & \textbf{89.85} & 85.94 & 80.86 & 78.86 & 79.16 & 80.60 & 79.76 & 81.82 & 77.23 & 78.85 & 75.64 & 78.64 & \textbf{100.00} & 93.33 & \textbf{97.17} & 87.78 & 83.21 & 83.58 \\
4 & \textbf{Z-Image} & 80.20 & 83.04 & 78.36 & 82.79 & 79.50 & 86.50 & 80.45 & 79.94 & 75.13 & 81.94 & 72.89 & 77.02 & 72.91 & 77.56 & 66.99 & 73.82 & 73.89 & 75.62 & 90.00 & 93.33 & 94.84 & \textbf{93.21} & 88.06 & 85.45 \\
5 & \textbf{Z-Image-Turbo} & 77.73 & 80.05 & 81.85 & 81.59 & 86.50 & 87.00 & 82.88 & 79.99 & 76.17 & 77.77 & 68.32 & 74.69 & 72.04 & 75.24 & 60.22 & 73.33 & 68.90 & 71.92 & 83.33 & 93.33 & 83.71 & 84.62 & 85.82 & 77.24 \\
6 & DALL-E 3~\cite{betker2023improving} & 74.96 & 70.81 & 78.72 & 78.50 & 79.50 & 79.83 & 80.82 & 78.82 & 75.82 & 76.82 & 73.39 & 67.27 & 73.45 & 67.20 & 72.01 & 71.34 & 63.59 & 60.72 & 89.66 & 86.67 & 66.83 & 54.83 & 72.93 & 60.99 \\
7 & FLUX.1 [dev]~\cite{labs2025flux} & 71.09 & 71.78 & 83.12 & 78.65 & 87.05 & 83.17 & 87.25 & 80.39 & 75.01 & 72.39 & 65.79 & 68.54 & 67.07 & 73.69 & 73.84 & 73.34 & 69.09 & 71.59 & 66.67 & 66.67 & 43.83 & 52.83 & 70.72 & 71.47 \\
8 & FLUX.1 [Pro]~\cite{labs2025flux} & 67.32 & 69.89 & 79.08 & 78.91 & 78.83 & 81.33 & 82.82 & 83.82 & 75.57 & 71.57 & 61.10 & 65.37 & 62.32 & 65.57 & 69.84 & 71.47 & 65.96 & 67.72 & 63.00 & 63.00 & 35.83 & 55.83 & 71.80 & 68.80 \\
9 & Midjourney V7~\cite{midjourneyV7} & 68.74 & 65.69 & 77.41 & 76.00 & 77.58 & 81.83 & 82.07 & 76.82 & 72.57 & 69.32 & 64.66 & 60.53 & 67.20 & 62.70 & 81.22 & 71.59 & 60.72 & 64.59 & 83.33 & 80.00 & 24.83 & 20.83 & 68.83 & 63.61 \\
10 & SD 3~\cite{esser2024scaling} & 67.46 & 66.09 & 78.32 & 77.75 & 83.33 & 79.83 & 82.07 & 78.82 & 71.07 & 74.07 & 61.46 & 59.56 & 61.07 & 64.07 & 68.84 & 70.34 & 50.96 & 57.84 & 66.67 & 76.67 & 59.83 & 20.83 & 63.23 & 67.34 \\
11 & SANA 1.5~\cite{xiesana1.5} & 67.15 & 65.73 & 79.66 & 77.08 & 79.83 & 77.83 & 85.57 & 83.57 & 73.57 & 69.82 & 61.50 & 60.67 & 65.32 & 56.57 & 69.96 & 73.09 & 62.96 & 65.84 & 80.00 & 80.00 & 17.83 & 15.83 & 71.07 & 68.83 \\
12 & Janus-Pro-7B~\cite{chen2025januspro} & 66.50 & 65.02 & 79.33 & 78.25 & 79.33 & 82.33 & 78.32 & 73.32 & 80.32 & 79.07 & 59.71 & 58.82 & 66.07 & 56.20 & 70.46 & 70.84 & 67.22 & 59.97 & 60.00 & 70.00 & 28.83 & 33.83 & 65.84 & 60.25 \\
13 & Infinity~\cite{han2025infinity} & 62.07 & 62.32 & 73.08 & 75.41 & 74.33 & 76.83 & 72.82 & 77.57 & 72.07 & 71.82 & 56.64 & 54.98 & 60.44 & 55.57 & 74.22 & 64.71 & 60.22 & 59.71 & 80.00 & 73.33 & 10.83 & 23.83 & 54.28 & 56.89 \\
14 & PixArt-$\Sigma$~\cite{chen2024pixart} & 62.00 & 58.12 & 70.66 & 75.25 & 69.33 & 78.83 & 75.07 & 77.32 & 67.57 & 69.57 & 57.65 & 49.50 & 65.20 & 56.57 & 66.96 & 61.72 & 66.59 & 54.59 & 83.33 & 70.00 & 1.83 & 1.83 & 62.11 & 52.41 \\
15 & Show-o~\cite{xie2024show} & 59.72 & 58.86 & 73.08 & 75.83 & 74.83 & 79.83 & 78.82 & 78.32 & 65.57 & 69.32 & 53.67 & 50.38 & 60.95 & 56.82 & 68.59 & 68.96 & 66.46 & 56.22 & 63.33 & 66.67 & 3.83 & 2.83 & 55.02 & 50.92 \\
16 & LightGen~\cite{wu2025lightgen} & 53.22 & 43.41 & 66.58 & 47.91 & 55.83 & 47.33 & 74.82 & 45.82 & 69.07 & 50.57 & 46.74 & 41.53 & 62.44 & 40.82 & 61.71 & 50.47 & 50.34 & 45.34 & 53.33 & 53.33 & 0.00 & 6.83 & 50.92 & 50.55 \\
17 & Hunyuan-DiT~\cite{li2024hunyuandit} & 51.38 & 53.28 & 69.33 & 69.00 & 65.83 & 69.83 & 78.07 & 73.82 & 64.07 & 63.32 & 42.62 & 45.45 & 50.20 & 41.57 & 59.22 & 61.84 & 47.84 & 51.09 & 56.67 & 73.33 & 0.00 & 0.83 & 40.10 & 44.20 \\
18 & Lumina-Next~\cite{gao2024lumina-next} & 50.93 & 52.46 & 64.58 & 66.08 & 56.83 & 59.33 & 67.57 & 71.82 & 69.32 & 67.07 & 44.75 & 45.63 & 51.44 & 43.20 & 51.09 & 59.72 & 44.72 & 54.46 & 70.00 & 66.67 & 0.00 & 0.83 & 47.56 & 49.05 \\
\bottomrule
\end{tabular}
\end{adjustbox}
\end{table}

\begin{table}[h!]
\centering

\caption{Quantitative results on PRISM-Bench~\cite{fang2025flux} evaluated by Qwen2.5-VL-72B~\cite{bai2025qwen2.5}.}
\label{tab:prism_bench_results}
\begin{adjustbox}{width=\textwidth}
\begin{tabular}{c l | ccc ccc ccc ccc ccc ccc ccc | ccc}
\toprule
\multirow{2}{*}{\textbf{Rank}} & \multirow{2}{*}{\textbf{Model}} & \multicolumn{3}{c}{\textbf{Imagination}} & \multicolumn{3}{c}{\textbf{Entity}} & \multicolumn{3}{c}{\textbf{Text rendering}} & \multicolumn{3}{c}{\textbf{Style}} & \multicolumn{3}{c}{\textbf{Affection}} & \multicolumn{3}{c}{\textbf{Composition}} & \multicolumn{3}{c}{\textbf{Long text}} & \multicolumn{3}{|c}{\textbf{Overall}$\uparrow$} \\
\cmidrule(lr){3-5} \cmidrule(lr){6-8} \cmidrule(lr){9-11} \cmidrule(lr){12-14} \cmidrule(lr){15-17} \cmidrule(lr){18-20} \cmidrule(lr){21-23} \cmidrule(lr){24-26}
& & Ali. & Aes. & Avg. & Ali. & Aes. & Avg. & Ali. & Aes. & Avg. & Ali. & Aes. & Avg. & Ali. & Aes. & Avg. & Ali. & Aes. & Avg. & Ali. & Aes. & Avg. & Ali. & Aes. & Avg. \\
\midrule
1 & GPT-Image-1 [High]~\cite{openai2025gptimage} & 79.8 & \textbf{53.3} & \textbf{66.6} & \textbf{87.3} & 81.0 & \textbf{84.1} & 66.7 & \textbf{86.8} & 76.8 & 87.3 & 87.8 & 87.5 & 88.1 & 79.8 & 84.0 & 92.2 & 84.9 & 88.5 & \textbf{77.2} & 77.5 & 77.4 & 82.7 & 78.7 & \textbf{80.7} \\
2 & Gemini 2.5-Flash-Image~\cite{google2025gemini2.5flashimage} & \textbf{84.7} & 38.1 & 61.4 & 86.0 & 76.7 & 81.3 & \textbf{72.8} & 84.3 & \textbf{78.5} & \textbf{89.5} & 87.8 & \textbf{88.6} & \textbf{94.3} & 74.8 & 84.5 & 91.2 & 88.2 & \textbf{89.7} & 76.3 & \textbf{80.6} & \textbf{78.4} & \textbf{85.0} & 75.8 & 80.4 \\
3 & \textbf{Z-Image-Turbo} & 65.7 & 50.1 & 57.9 & 75.7 & \textbf{82.3} & 79.0 & 59.6 & 84.9 & 72.2 & 76.7 & 88.2 & 82.4 & 85.1 & \textbf{87.4} & \textbf{86.2} & 89.0 & \textbf{90.2} & 89.6 & 69.8 & 79.0 & 74.4 & 74.5 & \textbf{80.3} & 77.4 \\
4 & Seedream 3.0~\cite{gao2025seedream} & 75.8 & 38.0 & 56.9 & 81.3 & 74.2 & 77.7 & 58.8 & 74.0 & 66.4 & 84.4 & 84.1 & 84.2 & 90.5 & 74.6 & 82.5 & 93.6 & 85.1 & 89.3 & 76.2 & 76.4 & 76.3 & 80.1 & 72.3 & 76.2 \\
5 & \textbf{Z-Image} & 68.0 & 47.3 & 57.6 & 75.0 & 74.4 & 74.7 & 59.3 & 81.6 & 70.4 & 78.0 & \textbf{89.0} & 83.5 & 84.3 & 80.1 & 82.2 & 89.1 & 85.1 & 87.1 & 70.6 & 76.6 & 73.6 & 74.9 & 76.2 & 75.6 \\
6 & Qwen-Image~\cite{qwenimage} & 75.5 & 37.4 & 56.5 & 79.5 & 64.5 & 72.0 & 57.9 & 71.2 & 64.5 & 86.6 & 84.4 & 85.5 & 89.9 & 70.4 & 80.1 & \textbf{93.9} & 79.5 & 86.7 & 76.8 & 70.9 & 73.8 & 80.0 & 68.3 & 74.1 \\
7 & FLUX.1-Krea-dev~\cite{labs2025flux} & 69.6 & 43.1 & 56.3 & 72.2 & 70.7 & 71.4 & 51.7 & 76.1 & 63.9 & 80.0 & 86.6 & 83.3 & 82.6 & 78.7 & 80.6 & 90.8 & 87.1 & 88.9 & 73.6 & 73.4 & 73.5 & 74.4 & 73.7 & 74.0 \\
8 & HiDream-I1-Full~\cite{cai2025hidream} & 73.0 & 44.0 & 58.5 & 76.3 & 72.8 & 74.5 & 60.5 & 76.4 & 68.4 & 81.4 & 81.5 & 81.4 & 90.0 & 76.6 & 83.3 & 88.5 & 80.3 & 84.4 & 66.3 & 48.6 & 57.4 & 76.6 & 68.6 & 72.6 \\
9 & SD3.5-Large~\cite{esser2024scaling} & 66.7 & 43.4 & 55.0 & 76.8 & 72.7 & 74.8 & 53.6 & 73.1 & 63.3 & 77.3 & 78.2 & 77.7 & 85.6 & 73.9 & 79.7 & 87.8 & 80.9 & 84.3 & 65.8 & 52.2 & 59.0 & 73.4 & 67.8 & 70.6 \\
10 & HiDream-I1-Dev~\cite{cai2025hidream} & 68.8 & 45.8 & 57.3 & 73.5 & 68.1 & 70.8 & 56.7 & 75.7 & 66.2 & 70.2 & 77.4 & 73.8 & 88.2 & 74.3 & 81.2 & 84.7 & 78.5 & 81.6 & 64.0 & 49.3 & 56.6 & 72.3 & 67.0 & 69.6 \\
11 & FLUX.1-dev~\cite{labs2025flux} & 65.5 & 42.9 & 54.2 & 70.6 & 61.9 & 66.2 & 52.3 & 73.0 & 62.6 & 72.6 & 74.2 & 73.4 & 86.0 & 72.9 & 79.4 & 87.4 & 75.8 & 81.6 & 70.5 & 53.8 & 62.1 & 72.1 & 64.9 & 68.5 \\
12 & SD3.5-Medium~\cite{esser2024scaling} & 65.1 & 34.7 & 49.9 & 72.5 & 70.9 & 71.7 & 36.6 & 64.5 & 50.5 & 75.5 & 80.0 & 77.7 & 81.8 & 73.9 & 77.9 & 85.4 & 81.0 & 83.2 & 63.5 & 50.6 & 57.0 & 68.6 & 65.1 & 66.8 \\
13 & SD3-Medium~\cite{esser2024scaling} & 64.3 & 37.7 & 51.0 & 69.4 & 63.3 & 66.3 & 38.5 & 63.3 & 50.9 & 74.6 & 79.5 & 77.0 & 80.5 & 75.5 & 78.0 & 85.6 & 79.5 & 82.5 & 63.4 & 50.3 & 56.8 & 68.0 & 64.2 & 66.1 \\
14 & FLUX.1-schnell~\cite{labs2025flux} & 62.8 & 35.6 & 49.2 & 64.8 & 56.8 & 60.8 & 54.3 & 68.1 & 61.2 & 70.3 & 71.5 & 70.9 & 75.4 & 65.9 & 70.6 & 81.7 & 75.6 & 78.6 & 68.7 & 54.4 & 61.5 & 68.3 & 61.1 & 64.7 \\
15 & Janus-Pro-7B~\cite{chen2025januspro} & 65.0 & 38.8 & 51.9 & 68.6 & 63.5 & 66.0 & 23.1 & 50.3 & 36.7 & 70.7 & 75.2 & 72.9 & 80.7 & 68.0 & 74.3 & 82.4 & 71.1 & 76.7 & 63.9 & 49.0 & 56.4 & 64.9 & 59.4 & 62.1 \\
16 & Bagel~\cite{deng2505emerging} & 68.0 & 45.0 & 56.5 & 67.6 & 53.4 & 60.5 & 29.4 & 42.3 & 35.8 & 69.0 & 69.7 & 69.3 & 87.1 & 66.7 & 76.9 & 86.6 & 69.2 & 77.9 & 64.5 & 50.2 & 57.3 & 67.5 & 56.6 & 62.0 \\
17 & Bagel-CoT~\cite{deng2505emerging} & 68.0 & 44.1 & 56.0 & 67.6 & 53.4 & 60.5 & 29.4 & 42.3 & 35.8 & 69.0 & 69.7 & 69.3 & 87.1 & 66.7 & 76.9 & 86.6 & 69.2 & 77.9 & 64.5 & 50.2 & 57.3 & 67.5 & 56.5 & 62.0 \\
18 & Playground~\cite{li2024playground} & 59.0 & 39.0 & 49.0 & 69.4 & 56.7 & 63.0 & 15.3 & 31.9 & 23.6 & 74.6 & 74.6 & 74.6 & 88.8 & 66.0 & 77.4 & 72.2 & 61.3 & 66.7 & 56.0 & 35.3 & 45.6 & 62.2 & 52.1 & 57.1 \\
19 & SDXL~\cite{podell2023sdxl} & 54.5 & 34.1 & 44.3 & 71.1 & 65.0 & 68.0 & 18.6 & 37.3 & 27.9 & 71.7 & 72.6 & 72.1 & 78.7 & 66.5 & 72.6 & 72.2 & 67.8 & 70.0 & 54.1 & 34.5 & 44.3 & 60.1 & 54.0 & 57.0 \\
20 & SD2.1~\cite{rombach2022high} & 48.9 & 28.4 & 38.6 & 66.0 & 57.6 & 61.8 & 16.7 & 31.4 & 24.0 & 62.7 & 66.5 & 64.6 & 68.5 & 62.1 & 65.3 & 64.8 & 58.3 & 61.5 & 50.7 & 29.8 & 40.2 & 54.0 & 47.7 & 50.8 \\
21 & SD1.5~\cite{rombach2022high} & 40.7 & 23.7 & 32.2 & 61.2 & 52.7 & 56.9 & 11.4 & 24.1 & 17.8 & 56.7 & 61.5 & 59.1 & 66.9 & 60.7 & 63.8 & 57.5 & 53.4 & 55.4 & 47.3 & 26.8 & 37.0 & 48.8 & 43.3 & 46.0 \\
\bottomrule
\end{tabular}
\end{adjustbox}
\end{table}

\begin{table}[h!]
\centering
\caption{Quantitative results on PRISM-Bench-ZH~\cite{fang2025flux} evaluated by Qwen2.5-VL-72B~\cite{bai2025qwen2.5}.}
\label{tab:prism_bench_zh_results}
\begin{adjustbox}{width=\textwidth}
\begin{tabular}{c l | ccc ccc ccc ccc ccc ccc ccc | ccc}
\toprule
\multirow{2}{*}{\textbf{Rank}} & \multirow{2}{*}{\textbf{Model}} & \multicolumn{3}{c}{\textbf{Imagination}} & \multicolumn{3}{c}{\textbf{Entity}} & \multicolumn{3}{c}{\textbf{Text rendering}} & \multicolumn{3}{c}{\textbf{Style}} & \multicolumn{3}{c}{\textbf{Affection}} & \multicolumn{3}{c}{\textbf{Composition}} & \multicolumn{3}{c}{\textbf{Long text}} & \multicolumn{3}{|c}{\textbf{Overall}$\uparrow$} \\
\cmidrule(lr){3-5} \cmidrule(lr){6-8} \cmidrule(lr){9-11} \cmidrule(lr){12-14} \cmidrule(lr){15-17} \cmidrule(lr){18-20} \cmidrule(lr){21-23} \cmidrule(lr){24-26}
& & Ali. & Aes. & Avg. & Ali. & Aes. & Avg. & Ali. & Aes. & Avg. & Ali. & Aes. & Avg. & Ali. & Aes. & Avg. & Ali. & Aes. & Avg. & Ali. & Aes. & Avg. & Ali. & Aes. & Avg. \\
\midrule
1 & GPT-Image-1 [High]~\cite{openai2025gptimage} & \textbf{73.0} & \textbf{37.6} & \textbf{55.3} & \textbf{80.4} & 82.1 & \textbf{81.3} & 73.1 & 89.9 & 81.5 & \textbf{77.1} & \textbf{92.4} & \textbf{84.8} & 78.0 & 77.8 & \textbf{77.9} & \textbf{91.9} & 85.7 & \textbf{88.8} & 72.4 & \textbf{76.3} & \textbf{74.4} & \textbf{78.0} & \textbf{77.4} & \textbf{77.7} \\
2 & \textbf{Z-Image} & 69.5 & 34.1 & 51.6 & 70.6 & 73.7 & 72.2 & \textbf{76.8} & \textbf{90.0} & \textbf{83.4} & 74.1 & 88.2 & 81.2 & 77.6 & 73.5 & 75.5 & 89.3 & \textbf{88.0} & 88.6 & 71.6 & 75.6 & 73.6 & 75.7 & 74.9 & 75.3 \\
3 & \textbf{Z-Image-Turbo} & 64.1 & 37.2 & 50.7 & 72.9 & \textbf{82.4} & 77.6 & 69.4 & 89.7 & 79.6 & 72.9 & 89.2 & 81.0 & 74.0 & \textbf{80.9} & 77.5 & 87.2 & 85.8 & 86.5 & 71.7 & 74.8 & 73.3 & 73.1 & 77.1 & 75.1 \\
4 & Seedream 3.0~\cite{gao2025seedream} & 71.4 & 36.6 & 54.0 & 74.8 & 73.8 & 74.3 & 70.7 & 88.0 & 79.4 & 74.1 & 88.0 & 81.1 & \textbf{79.0} & 71.4 & 75.2 & 90.3 & 83.2 & 86.8 & \textbf{73.0} & 71.2 & 72.1 & 76.2 & 73.2 & 74.7 \\
5 & Qwen-Image~\cite{qwenimage} & 71.4 & 29.9 & 50.7 & 74.7 & 67.8 & 71.3 & 64.3 & 73.1 & 68.7 & 75.2 & 83.2 & 79.2 & 77.3 & 64.5 & 70.9 & 89.8 & 74.1 & 82.0 & 72.6 & 65.8 & 69.2 & 75.0 & 65.5 & 70.3 \\
6 & Bagel-CoT~\cite{deng2505emerging} & 64.4 & 36.6 & 50.5 & 62.6 & 53.8 & 58.2 & 25.2 & 51.9 & 38.6 & 65.4 & 76.7 & 71.1 & 74.0 & 65.0 & 69.5 & 81.3 & 71.3 & 76.3 & 61.4 & 46.6 & 54.0 & 62.0 & 57.4 & 59.7 \\
7 & Bagel~\cite{deng2505emerging} & 64.6 & 36.3 & 50.5 & 62.7 & 55.5 & 59.1 & 18.6 & 26.3 & 22.5 & 66.0 & 76.6 & 71.3 & 74.9 & 66.2 & 70.6 & 81.3 & 72.2 & 76.8 & 62.4 & 47.3 & 54.9 & 61.5 & 54.3 & 57.9 \\
8 & HiDream-I1-Full~\cite{cai2025hidream} & 51.2 & 30.8 & 41.0 & 60.1 & 61.3 & 60.7 & 20.7 & 40.6 & 30.7 & 64.5 & 73.8 & 69.2 & 65.2 & 69.1 & 67.2 & 72.4 & 69.0 & 70.7 & 57.1 & 42.8 & 50.0 & 55.9 & 55.3 & 55.6 \\
9 & HiDream-I1-Dev~\cite{cai2025hidream} & 48.3 & 24.6 & 36.5 & 52.6 & 54.1 & 53.4 & 18.6 & 35.3 & 27.0 & 59.0 & 68.3 & 63.7 & 65.9 & 62.3 & 64.1 & 66.5 & 64.6 & 65.6 & 54.2 & 38.6 & 46.4 & 52.2 & 49.7 & 50.9 \\
\bottomrule
\end{tabular}
\end{adjustbox}
\end{table}

\paragraph{PRISM-Bench.} We evaluate our models on PRISM-Bench~\cite{fang2025flux}, a VLM-powered benchmark assessing reasoning and aesthetics across seven tracks. On the English track (Table~\ref{tab:prism_bench_results}), Z-Image-Turbo achieves the 3rd rank (77.4), outperforming the base model and Qwen-Image, which highlights its superior efficiency and generation quality. On the Chinese track (Table~\ref{tab:prism_bench_zh_results}), Z-Image ranks 2nd (75.3), demonstrating robust multi-lingual performance with exceptional scores in \textit{Text Rendering} (83.4) and \textit{Composition} (88.6).

\begin{table}[h!]
\centering
\caption{Quantitative Evaluation results on ImgEdit~\cite{ye2025imgedit}.}
\label{tab:imgedit_results}
\setlength{\tabcolsep}{1.1mm}
\begin{adjustbox}{width=\textwidth}
\begin{tabular}{cl|ccccccccc|c}
\toprule
\textbf{Rank} & \textbf{Model} & \textbf{Add} & \textbf{Adjust} & \textbf{Extract} & \textbf{Replace} & \textbf{Remove} & \textbf{Background} & \textbf{Style} & \textbf{Hybrid} & \textbf{Action} & \textbf{Overall}$\uparrow$ \\
\midrule
1 & UniWorld-V2~\cite{li2025uniworld} & 4.29 & \textbf{4.44} & \textbf{4.32} & \textbf{4.69} & \textbf{4.72} & {4.41} & {4.91} & {3.83} & {4.83} & \textbf{4.49} \\
2 & Qwen-Image-Edit [2509]~\cite{qwenimage} & 4.32 & 4.36 & 4.04 & 4.64 & 4.52 & 4.37 & 4.84 & 3.39 & 4.71 & 4.35 \\
3 & \textbf{Z-Image-Edit} & {4.40} & {4.14} & {4.30} & {4.57} & {4.13} & {4.14} & {4.85} & {3.63} & {4.50} & {4.30} \\
4 & Qwen-Image-Edit~\cite{qwenimage} & 4.38 & 4.16 & 3.43 & 4.66 & 4.14 & 4.38 & 4.81 & 3.82 & 4.69 & 4.27 \\
5 & GPT-Image-1 [High]~\cite{openai2025gptimage} & \textbf{4.61} & 4.33 & 2.90 & 4.35 & 3.66 & \textbf{4.57} & \textbf{4.93} & \textbf{3.96} & \textbf{4.89} & 4.20 \\
6 & FLUX.1 Kontext [Pro]~\cite{labs2025flux} & 4.25 & 4.15 & 2.35 & 4.56 & 3.57 & 4.26 & 4.57 & 3.68 & 4.63 & 4.00 \\
7 & OmniGen2~\cite{wu2025omnigen2} & 3.57 & 3.06 & 1.77 & 3.74 & 3.20 & 3.57 & 4.81 & 2.52 & 4.68 & 3.44 \\
8 & UniWorld-V1~\cite{lin2025uniworld} & 3.82 & 3.64 & 2.27 & 3.47 & 3.24 & 2.99 & 4.21 & 2.96 & 2.74 & 3.26 \\
9 & BAGEL~\cite{deng2505emerging} & 3.56 & 3.31 & 1.70 & 3.30 & 2.62 & 3.24 & 4.49 & 2.38 & 4.17 & 3.20 \\
10 & Step1X-Edit~\cite{liu2025step1x} & 3.88 & 3.14 & 1.76 & 3.40 & 2.41 & 3.16 & 4.63 & 2.64 & 2.52 & 3.06 \\
11 & ICEdit~\cite{zhang2025context} & 3.58 & 3.39 & 1.73 & 3.15 & 2.93 & 3.08 & 3.84 & 2.04 & 3.68 & 3.05 \\
12 & OmniGen~\cite{xiao2025omnigen} & 3.47 & 3.04 & 1.71 & 2.94 & 2.43 & 3.21 & 4.19 & 2.24 & 3.38 & 2.96 \\
13 & UltraEdit~\cite{zhao2024ultraedit} & 3.44 & 2.81 & 2.13 & 2.96 & 1.45 & 2.83 & 3.76 & 1.91 & 2.98 & 2.70 \\
14 & AnyEdit~\cite{yu2025anyedit} & 3.18 & 2.95 & 1.88 & 2.47 & 2.23 & 2.24 & 2.85 & 1.56 & 2.65 & 2.45 \\
15 & MagicBrush~\cite{zhang2023magicbrush} & 2.84 & 1.58 & 1.51 & 1.97 & 1.58 & 1.75 & 2.38 & 1.62 & 1.22 & 1.90 \\
16 & Instruct-Pix2Pix~\cite{brooks2023instructpix2pix} & 2.45 & 1.83 & 1.44 & 2.01 & 1.50 & 1.44 & 3.55 & 1.20 & 1.46 & 1.88 \\
\bottomrule
\end{tabular}
\end{adjustbox}
\end{table}

\begin{table}[h!]
\centering

\caption{Quantitative Evaluation results on GEdit-Bench~\cite{liu2025step1x}.}
\label{tab:gedit_results}
\setlength{\tabcolsep}{2.0mm}
\begin{adjustbox}{width=0.7\textwidth}
\begin{tabular}{cl|ccc|ccc}
\toprule
\multirow{2}{*}{\textbf{Rank}} & \multirow{2}{*}{\textbf{Model}} & \multicolumn{3}{c|}{\textbf{GEdit-Bench-EN}} & \multicolumn{3}{c}{\textbf{GEdit-Bench-CN}} \\
& & G\_SC & G\_PQ & G\_O$\uparrow$ & G\_SC & G\_PQ & G\_O$\uparrow$ \\
\midrule
1 & UniWorld-V2~\cite{li2025uniworld} & \textbf{8.39} & \textbf{8.02} & \textbf{7.83} & - & - & - \\
2 & Qwen-Image-Edit [2509]~\cite{qwenimage} & 8.15 & 7.86 & 7.54 & \textbf{8.08} & \textbf{7.89} & \textbf{7.54} \\
3 & \textbf{Z-Image-Edit} & {8.11} & {7.72} & {7.57} & {8.03} & {7.80} & \textbf{7.54} \\
4 & Qwen-Image-Edit~\cite{qwenimage} & 8.00 & 7.86 & 7.56 & 7.82 & 7.79 & 7.52 \\
5 & GPT-Image-1 [High]~\cite{openai2025gptimage} & 7.85 & 7.62 & 7.53 & 7.67 & 7.56 & 7.30 \\
6 & Step1X-Edit~\cite{liu2025step1x} & 7.66 & 7.35 & 6.97 & 7.20 & 6.87 & 6.86 \\
7 & BAGEL~\cite{deng2505emerging} & 7.36 & 6.83 & 6.52 & 7.34 & 6.85 & 6.50 \\
8 & OmniGen2~\cite{wu2025omnigen2} & 7.16 & 6.77 & 6.41 & - & - & - \\
9 & FLUX.1 Kontext [Pro]~\cite{labs2025flux} & 7.02 & 7.60 & 6.56 & 1.11 & 7.36 & 1.23 \\
10 & FLUX.1 Kontext [Dev]~\cite{labs2025flux} & 6.52 & 7.38 & 6.00 & - & - & - \\
11 & OmniGen~\cite{xiao2025omnigen} & 5.96 & 5.89 & 5.06 & - & - & - \\
12 & UniWorld-V1~\cite{lin2025uniworld} & 4.93 & 7.43 & 4.85 & - & - & - \\
13 & MagicBrush~\cite{zhang2023magicbrush} & 4.68 & 5.66 & 4.52 & - & - & - \\
14 & Instruct-Pix2Pix~\cite{brooks2023instructpix2pix} & 3.58 & 5.49 & 3.68 & - & - & - \\
15 & AnyEdit~\cite{yu2025anyedit} & 3.18 & 5.82 & 3.21 & - & - & - \\
\bottomrule
\end{tabular}
\end{adjustbox}
\end{table}

\newpage
\subsubsection{Instruction-based Image Editing}

\textbf{ImgEdit.}
Table~\ref{tab:imgedit_results} shows the evaluation of Z-Image-Edit on the ImgEdit Benchmark~\cite{ye2025imgedit}, 
where the metric combines instruction completion and visual quality. 
Across 9 common editing tasks, 
Z-Image-Edit shows competitive editing performance with leading models , 
especially object addition and extraction.

\textbf{GEdit.}
We also evaluate Z-Image-Edit on the GEdit-Bench~\cite{liu2025step1x}, 
which evaluates visual naturalness~(G\_PQ) and bilingual instruction following~(G\_SC). 
GEdit-Bench-EN abd GEdit-Bench-CN adopt English and Chinese instructions in the evaluation, respectively.
As shown in Table~\ref{tab:gedit_results}, 
Z-Image-Edit achieves 3rd rank, 
demonstrating robust bilingual editing capabilities. 

\subsection{Qualitative Evaluation}
\label{sec:qualitative_evaluation}
To further demonstrate the visual generation capacity of Z-Image~\footnote{In the section, all results of Z-Image are generated by our Turbo version.}, we first give the qualitative comparison against state-of-the-art open-source models (Lumina-Image 2.0~\cite{qin2025lumina}, Qwen-Image~\cite{qwenimage}, HunyuanImage 3.0~\cite{cao2025hunyuanimage}, and FLUX 2 dev~\cite{flux-2-2025}) and close-source models (Imagen 4 Ultra~\cite{google2025imagen4}, Seedream 4.0~\cite{seedream2025seedream} and Nano Banana Pro~\cite{nanopro}). We then evaluate Z-Image-Edit across four core dimensions: complex instruction-following, localized interactive editing, camera perspective control, and multi-image ID consistency. Finally, we showcase the model's enhanced reasoning capabilities enabled by our prompt enhancer, as well as its emerging multilingual and multicultural understanding.

\subsubsection{Superior Photorealistic Generation}
\begin{figure}[H]
  \centering
  \vspace{2em}
  \includegraphics[width=1\textwidth]{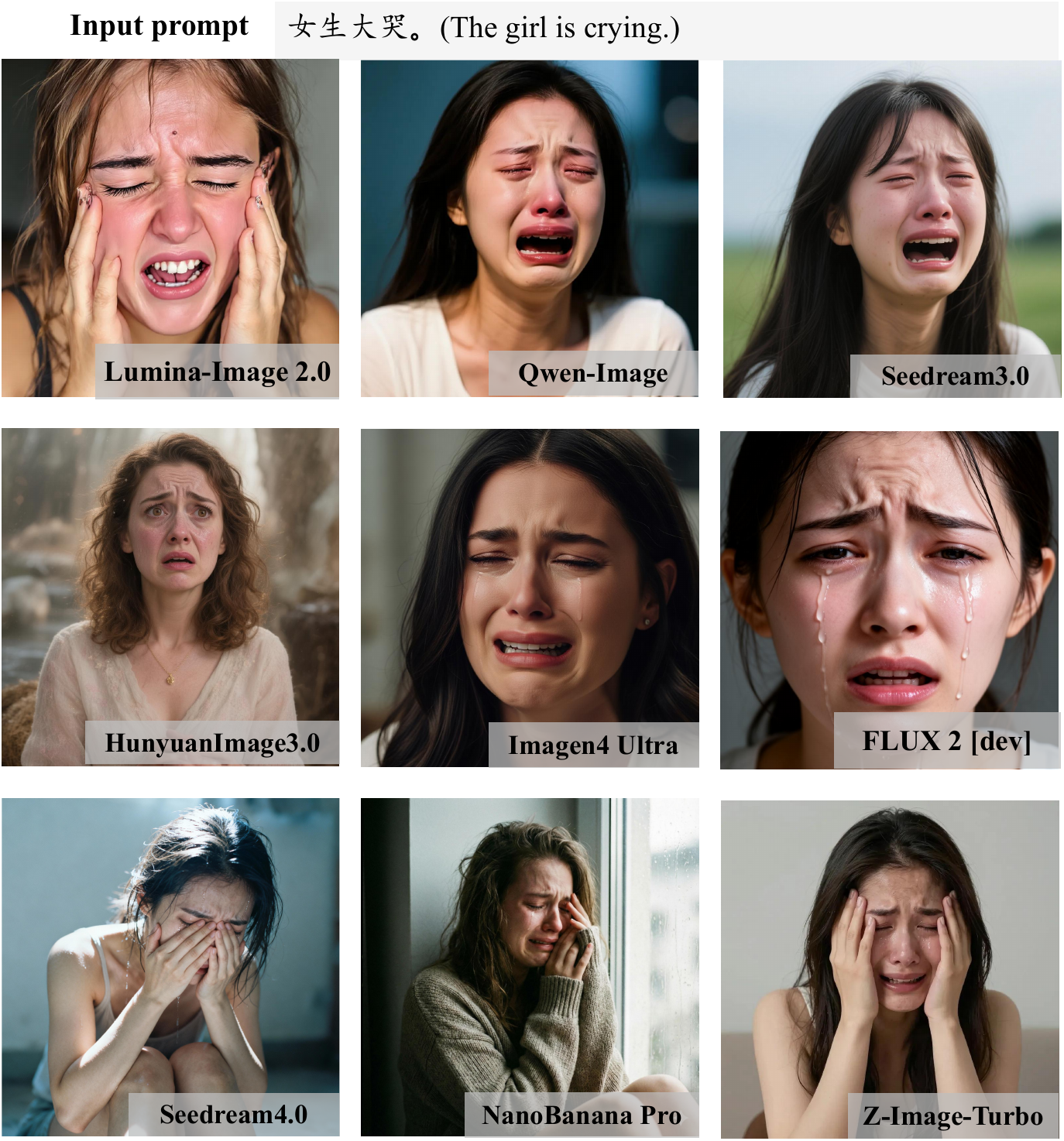}
  \caption{Comparison of close-up portrait generation, which indicates that Z-Image exhibits strong capabilities in character emotion and skin texture rendering. Better to zoom in to check the subtle expressions and the texture of the skin. }
  \label{fig:human_c1_woman}
\end{figure}

\begin{figure}[H]
  \centering
  \includegraphics[width=1\textwidth]{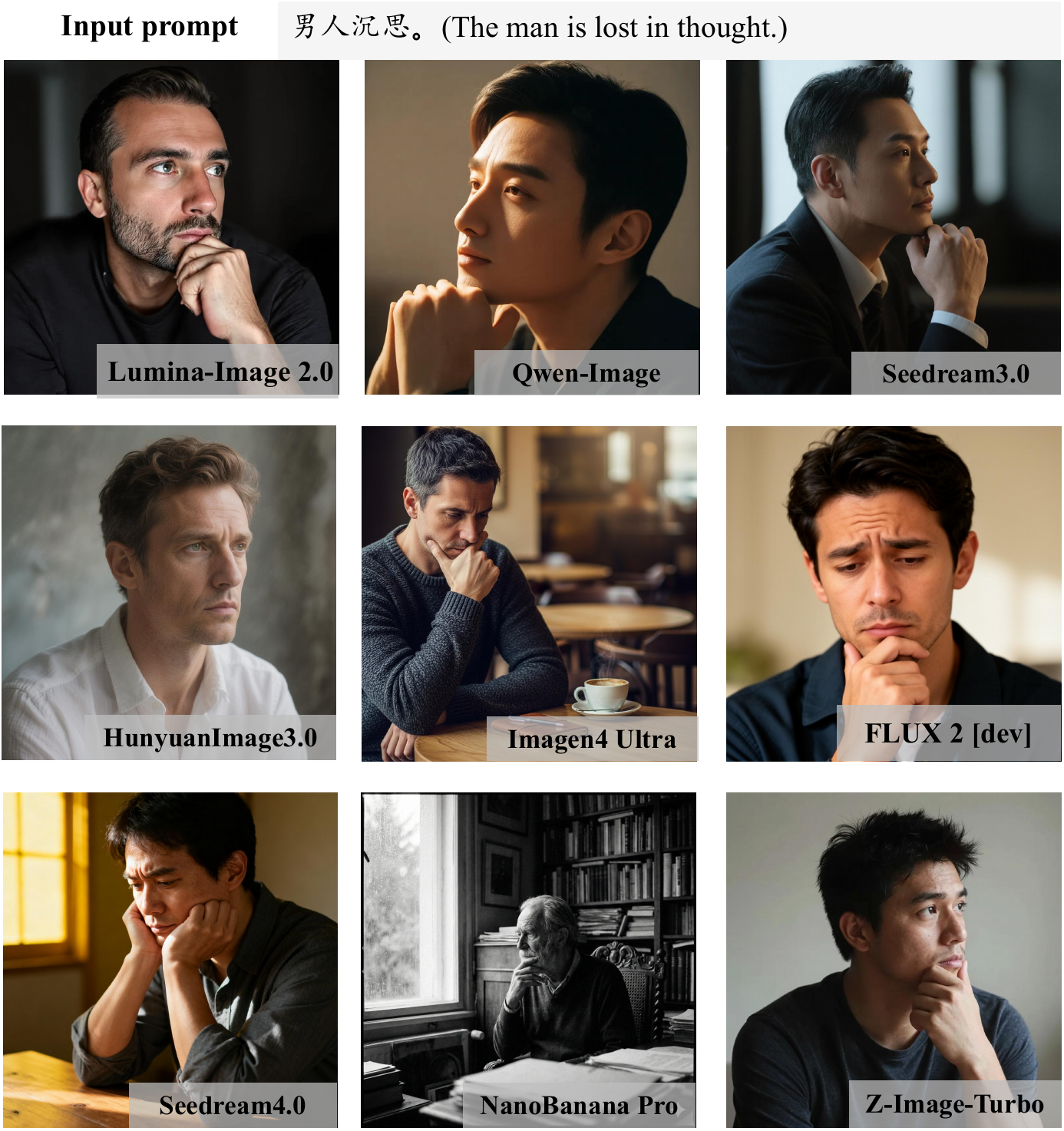}
  \caption{Comparison of close-up portrait generation, which indicates that Z-Image exhibits strong capabilities in character emotion and skin texture rendering. Better to zoom in to check the subtle expressions and the texture of the skin. }
  \label{fig:human_c1_man}
\end{figure}

\begin{figure}[H]
  \vspace{-1em}
  \centering
  \includegraphics[width=1\textwidth]{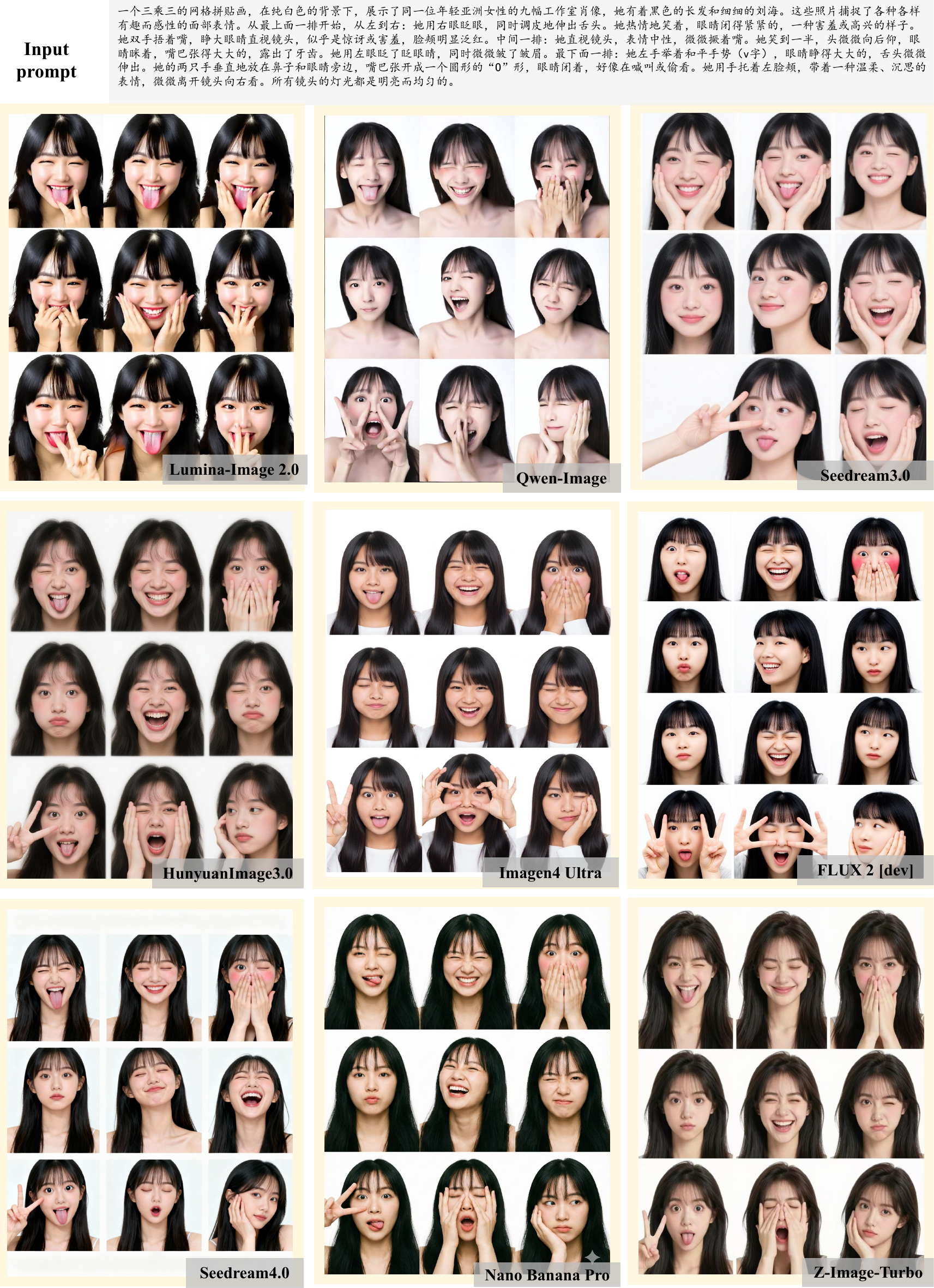}
  \caption{Comparison of complex close-up portrait generation, which indicates that Z-Image-Turbo has a strong ability in rendering character expressions and skin textures, as well as generating aesthetic images. Better to zoom in to check the subtle expressions.}
  \label{fig:human_c2}
\end{figure}

\begin{figure}[H]
\vspace{-0em}
  \centering
  \includegraphics[width=1\textwidth]{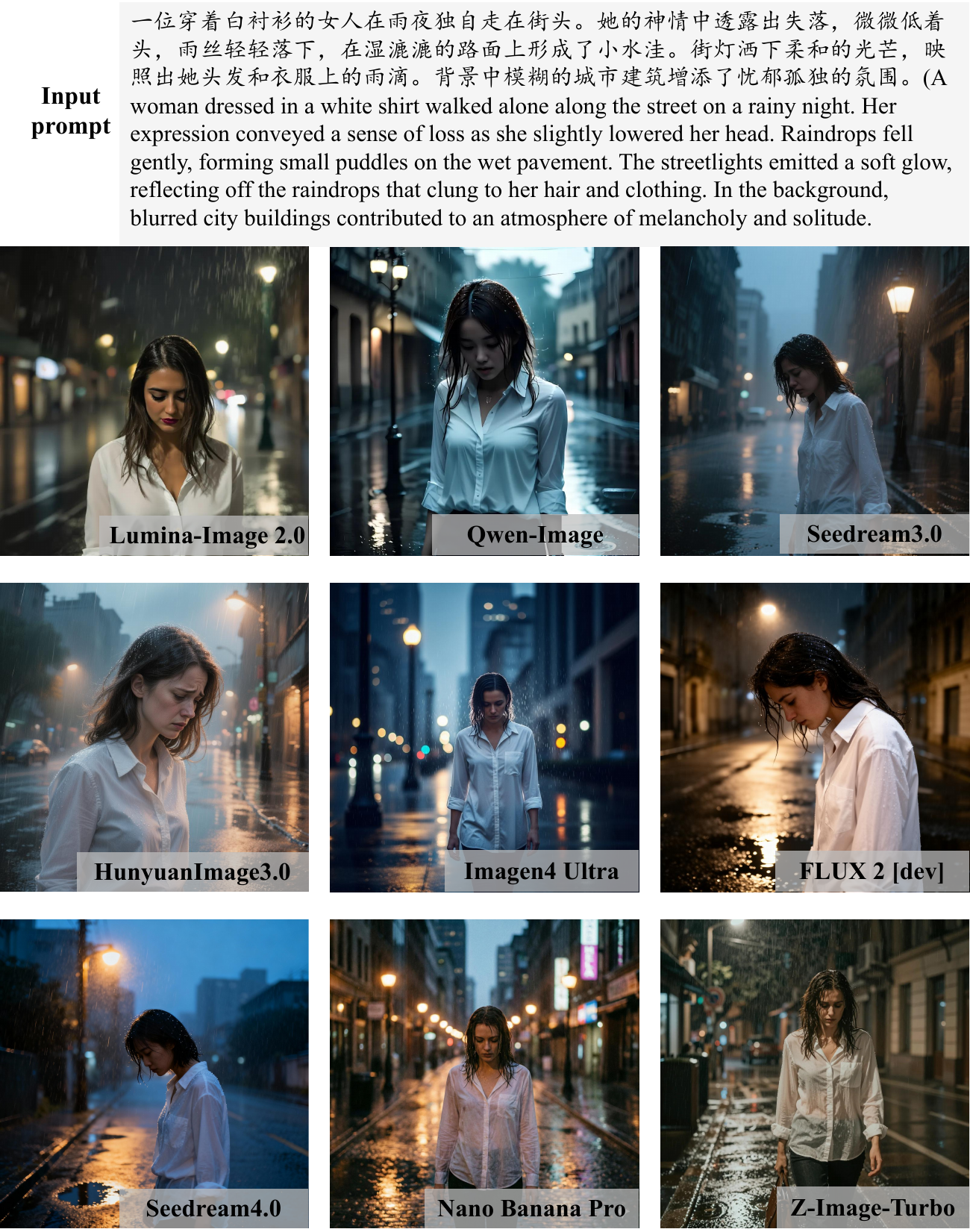}
  \caption{Comparison of scene shooting, which indicates that Z-Image-Turbo shows strong performance in the authenticity of both the person and the background, as well as the aesthetic appeal of layout and posture. Better to zoom in to check the texture of the clothes and hair.}
  \label{fig:human_s1}
\end{figure}

\begin{figure}[H]
\vspace{-0em}
  \centering
  \includegraphics[width=1\textwidth]{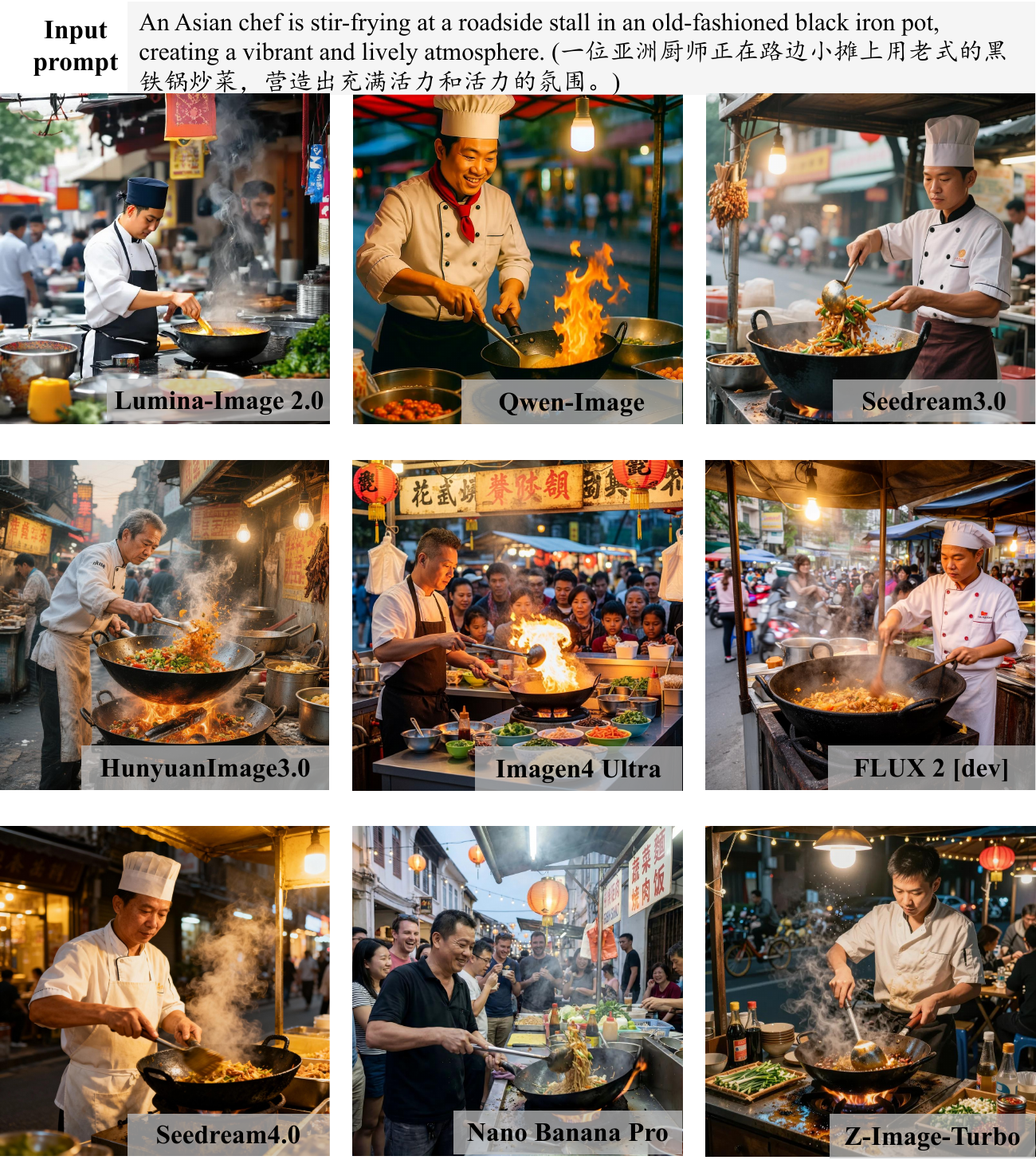}
  \caption{Comparison of scene shooting, which indicates that Z-Image-Turbo shows strong performance in the authenticity of both the person and the background, as well as the aesthetic appeal of layout and posture. Better to zoom in to check the details.}
  \label{fig:human_s2}
\end{figure}

As shown in Figure~\ref{fig:human_c1_woman} and ~\ref{fig:human_c1_man}, Z-Image-Turbo shows excellent character close-up generation (\eg, the skin details on a man's face and a girl's tears). When  asked to generate multi-expression portraits of one person (Figure~\ref{fig:human_c2}), Z-Image-Turbo can produce images that are more aesthetically pleasing and have more realistic expressions, while Qwen-Image, HunyuanImage3.0, FLUX 2 dev, and Seedream 4.0 would sometimes generate exaggerated and unrealistic expressions, thus lacking authenticity and beauty.  Moreover, when generating a scene captured by a mobile phone (Figure~\ref{fig:human_s1} and ~\ref{fig:human_s2}),  Z-Image-Turbo shows strong performance
in the authenticity of both the person and the background, as well as the aesthetic appeal of layout and posture. while  Qwen-Image, HunyuanImage3.0, and FLUX 2 dev would
generate unrealistic things (\eg, clothes that remain completely unsoaked in the heavy rain).

\subsubsection{Outstanding Bilingual Text Rendering}

\begin{figure}[H]
  \centering  
  \includegraphics[width=1\textwidth]{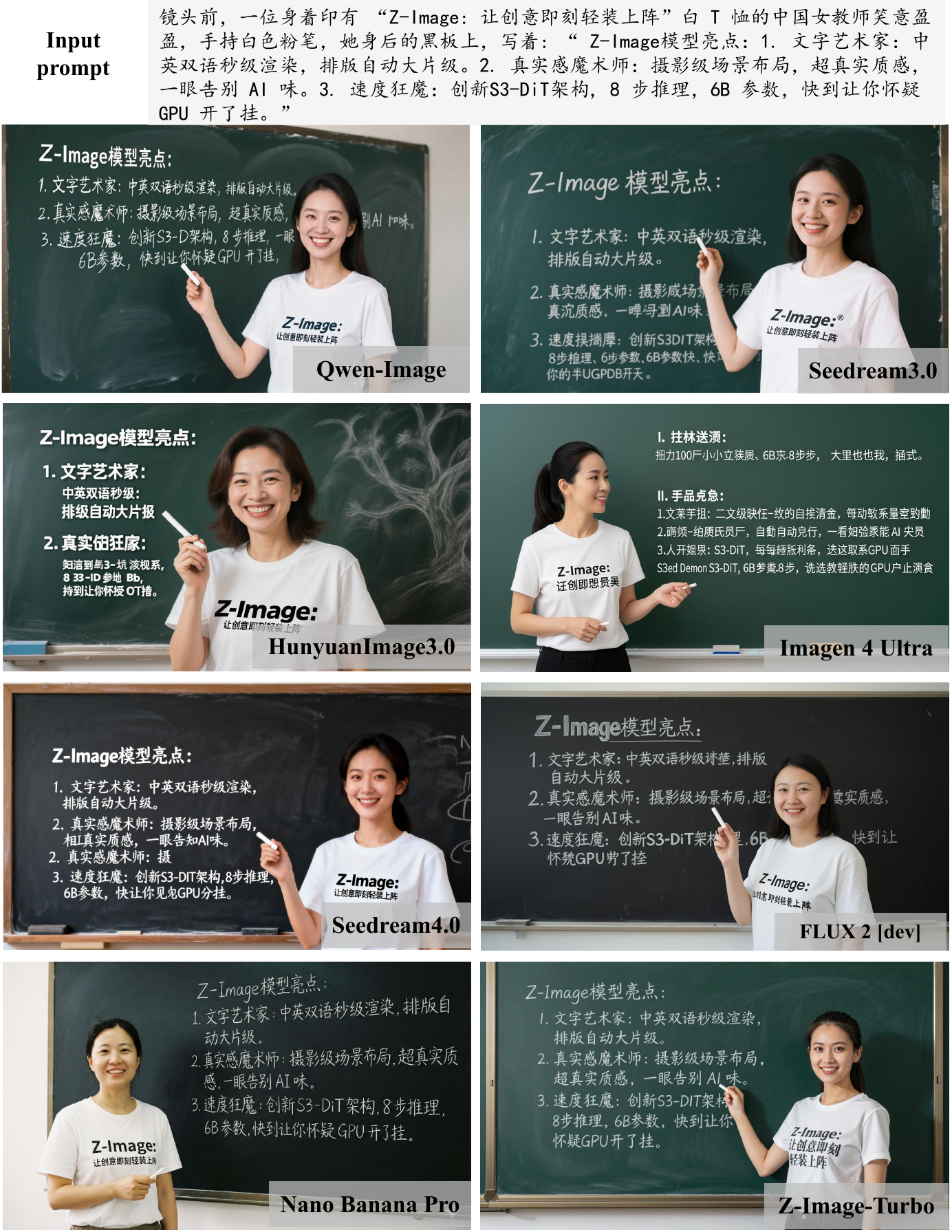}
  \caption{Comparison of complex Chinese text rendering. It shows that only Z-Image-Turbo and Nano Banana Pro can accurately
generates the expected Chinese couplet. Better to zoom in to check the correctness of the rendered text and the authenticity of the person.}
  \label{fig:text_c1}
\end{figure}

\begin{figure}[H]
  \centering
  \includegraphics[width=1\textwidth]{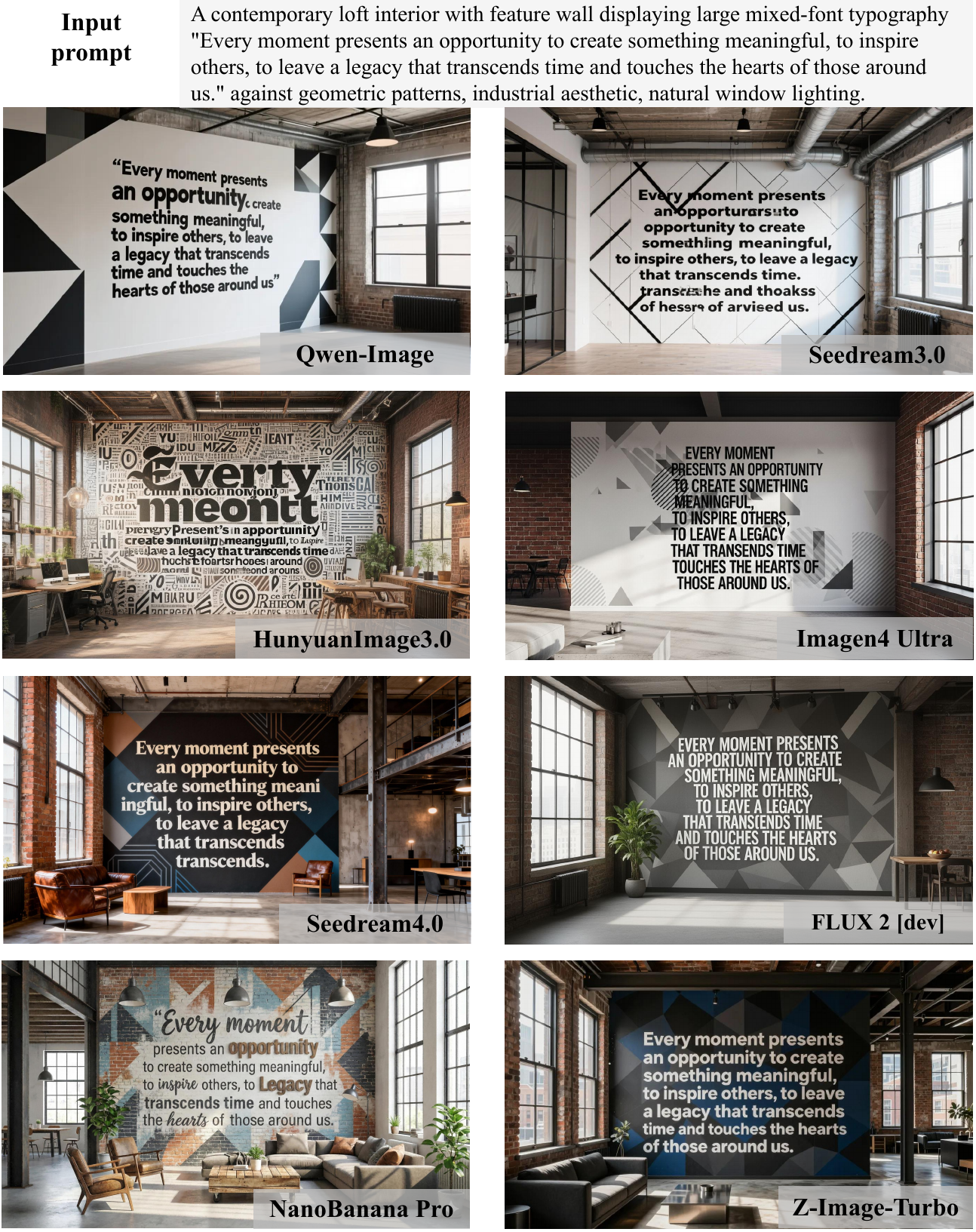}
  \caption{Comparison of complex English text rendering. It shows that only Z-Image-Turbo and Nano Banana Pro can accurately
generates the expected English couplet. Better to zoom in to check the correctness of the rendered text and the layout of the scene.}
  \label{fig:text_e1}
\end{figure}

\begin{figure}[H]
  \centering
  \vspace{-5em}
  \includegraphics[width=0.9\textwidth]{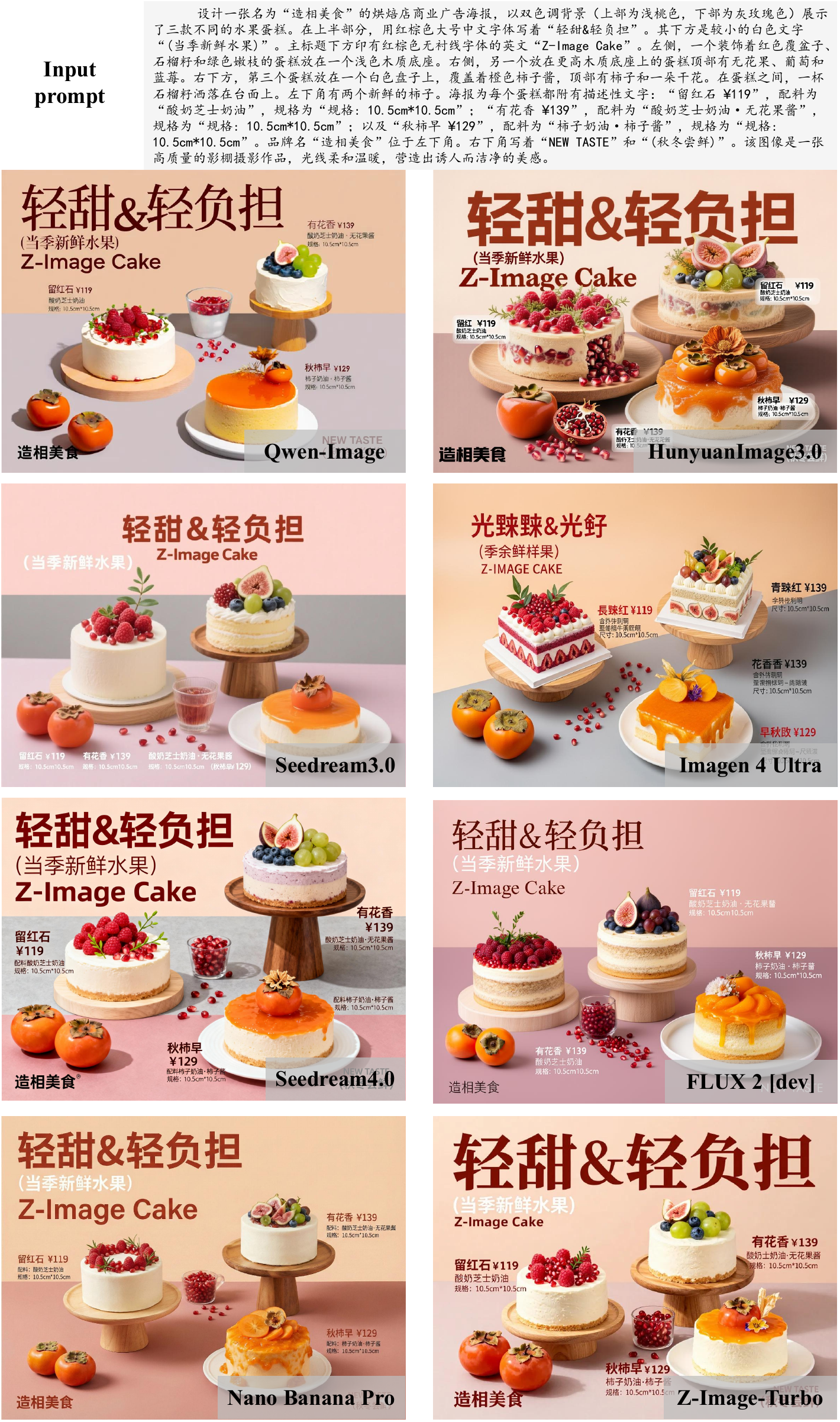}
  \caption{Comparison of Chinese text rendering in poster design. Z-Image-Turbo not only presents correct text rendering, but also designs a more aesthetically pleasing and realistic poster. Better to zoom in to check the correctness of the rendered text and the fidelity of the food. }
  \label{fig:text_c2}
\end{figure}

\begin{figure}[H]
  \centering
  \includegraphics[width=1\textwidth]{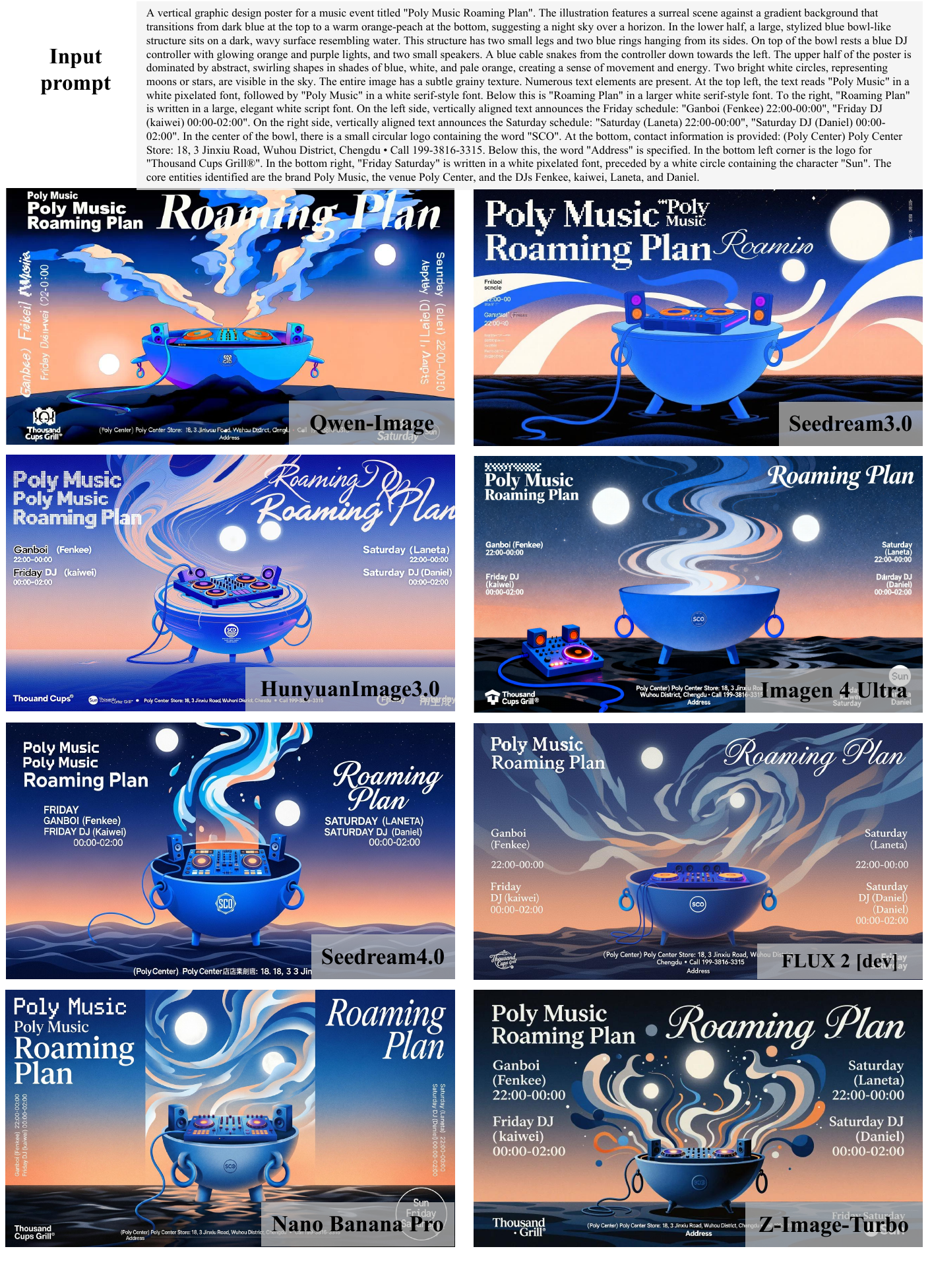}
  \caption{Comparison of English text rendering in poster design. Only Z-Image-Turbo presents correct text rendering with a pleasing and realistic poster. Better to zoom in to check the correctness of the rendered text and the details of the poster.}
  \label{fig:text_e2}
\end{figure}

Figure~\ref{fig:text_c1} and Figure~\ref{fig:text_e1} show the qualitative comparison of Chinese and English text rendering. As shown in Figure~\ref{fig:text_c1} and Figure~\ref{fig:text_e1}, Z-Image-Turbo accurately rendered the required text while maintaining the aesthetic appeal and authenticity of other parts (\eg, the authenticity of the human face in Figure~\ref{fig:text_c1} and the layout of the scene in Figure~\ref{fig:text_e1}). Note that this is comparable to the leading closed-source model Nano Banana Pro, and surpasses other candidates.  When rendering text in poster design (Figure~\ref{fig:text_c2} and Figure~\ref{fig:text_e2}), Z-Image-Turbo not only presents correct text rendering, but also designs a more aesthetically pleasing and realistic poster. For example, as shown in  Figure~\ref{fig:text_e2}), Qwen-Image, HunyuanImage3.0, FLUX 2 dev, and Imagen 4 Ultra make errors when rendering very small characters, Seedream4.0 and Nano Banana Pro make errors of repeatedly rendering the text, while Z-Image-Turbo gets the poster with the right rendered text  and satisfactory design.



\subsubsection{Precise and Interactive Localized Editing}
\begin{figure}[H]
  \centering
  \includegraphics[width=1\textwidth]{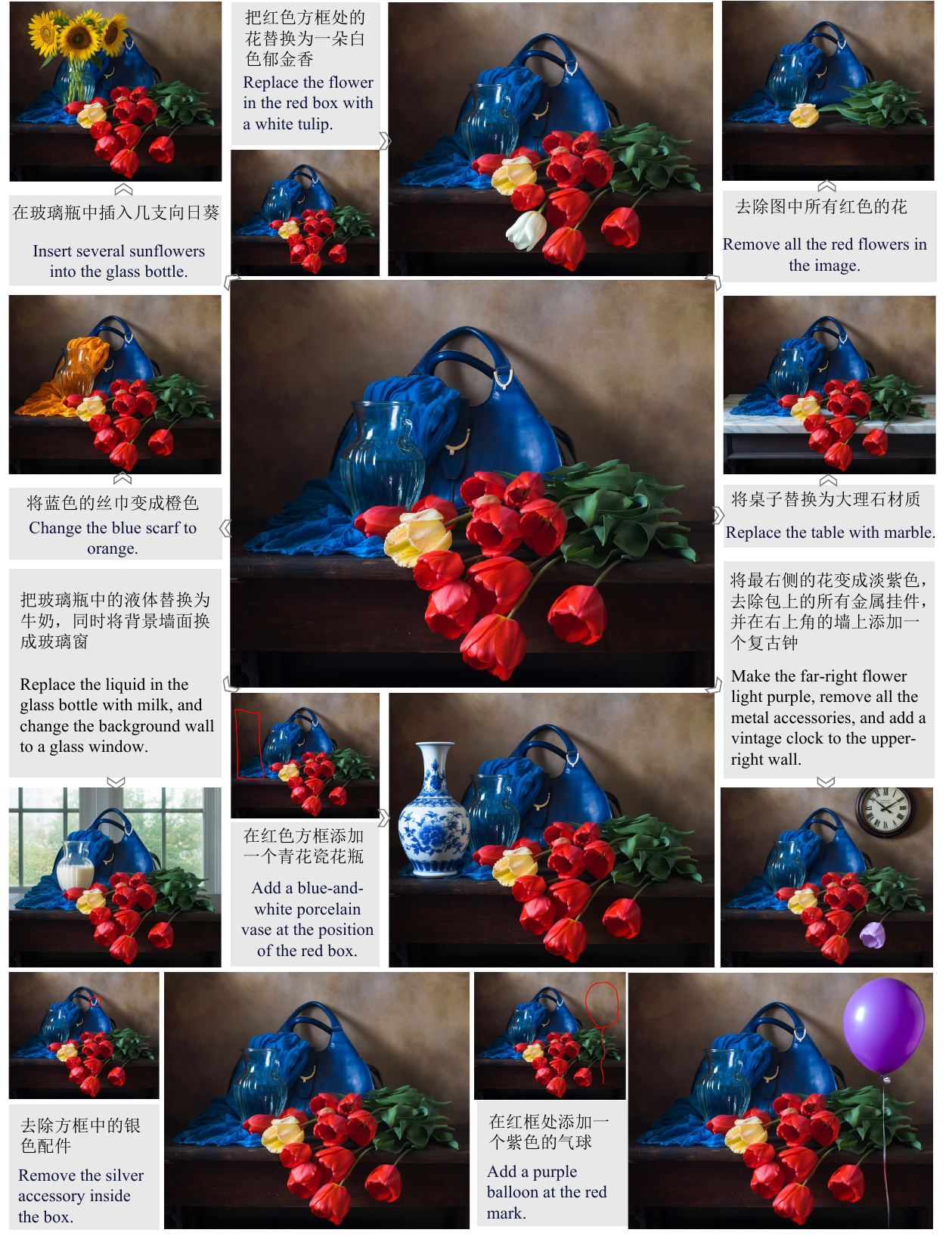}
  \caption{Localized image manipulation results of Z-Image-Edit. Radiating from the central reference still-life image, ten diverse editing results demonstrate our model's capacity for seamless addition, removal, and replacement. The model naturally supports interactive inputs such as bounding boxes (red boxes) and user scribbles (red curves), while preserving rigorous physical consistency, including realistic lighting, shadows, and glass refractions.}
  \label{fig:local_editing}
\end{figure}

Z-Image-Edit exhibits extremely impressive performance in localized image manipulation, supporting seamless addition, removal, and replacement of specific elements, as shown in Figure~\ref{fig:local_editing}. Taking a still-life still setup as the central reference, our model can effortlessly execute localized object replacements, such as changing the blue scarf to orange, replacing the wooden table with marble, or replacing the liquid inside the glass bottle with milk. It also performs flawless object removal (e.g., removing all red flowers or removing the bag's silver accessories) and realistic additions (e.g., inserting sunflowers into the vase or placing a vintage clock on the background wall).

Furthermore, Z-Image-Edit enables multi-modal interactive editing guided by bounding boxes and user scribbles. As shown in the red-marked miniatures, by mapping modifications to explicit spatial coordinates and sketch strokes, users can intuitively direct localized changes, such as replacing a specific red flower with a white tulip via a red bounding box, or adding a purple balloon via a red scribble. Throughout these edits, our model maintains exceptional "zero-seam" blending and shadow harmonization, automatically recalculating physically realistic shadows, reflections, and light refractions (e.g., the milk inside the glass container) to perfectly match the original ambient light.

\subsubsection{Advanced Camera Controllability and Spatial Consistency}
\begin{figure}[H]
  \centering
  \includegraphics[width=1\textwidth]{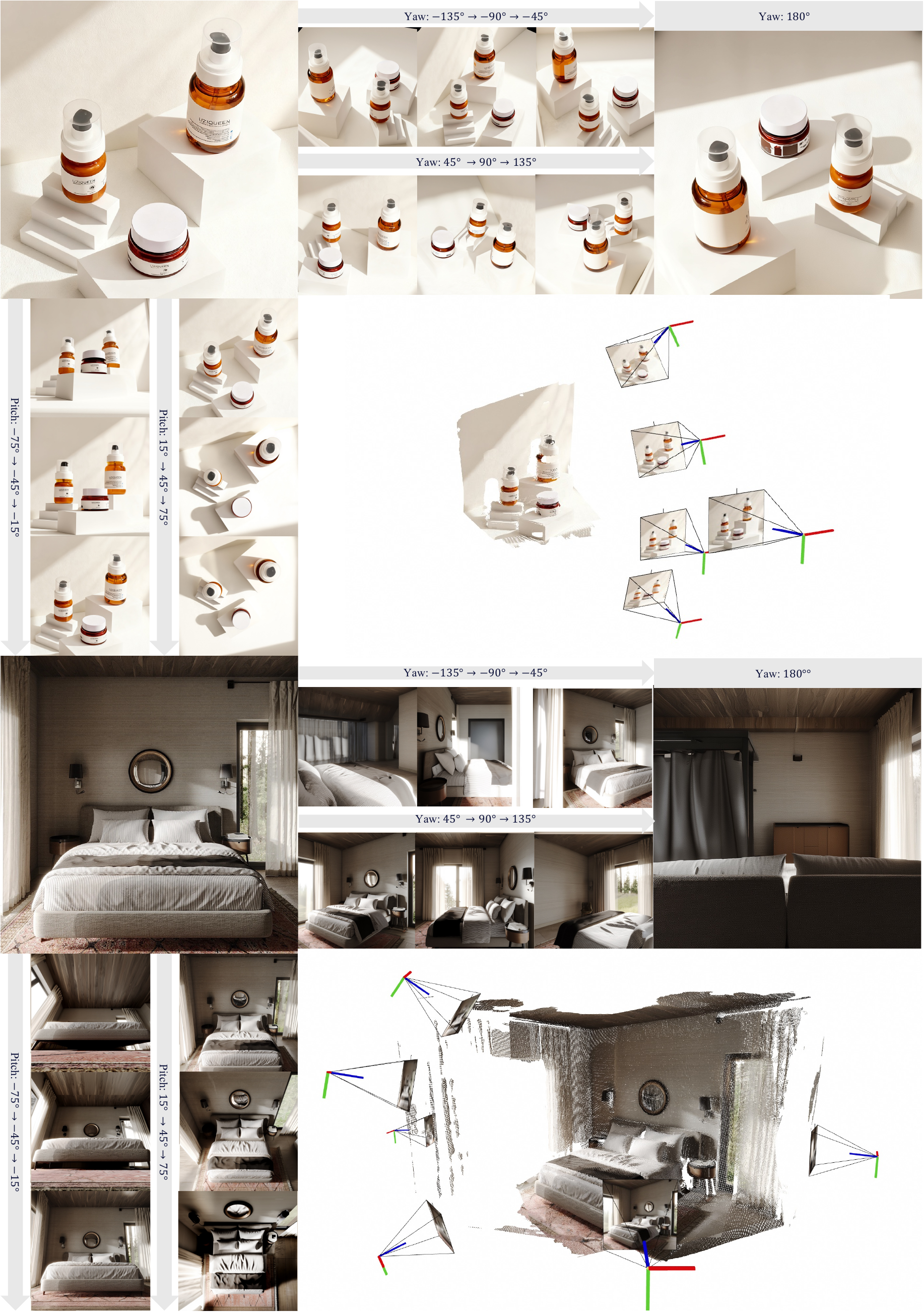}
  \caption{Multi-view camera control and 3D reconstruction validation of Z-Image-Edit. We demonstrate perspective-controlled generation across various yaw and pitch angles. 3D reconstructions via VGGT validate the geometric consistency and precise spatial alignment of our outputs.}
  \label{fig:view_transfer}
\end{figure}

Z-Image-Edit displays exceptional camera and viewpoint control, maintaining rigorous spatial and geometric consistency under dramatic perspective shifts. As shown in Figure~\ref{fig:view_transfer}, our model can precisely generate novel views of both object-centric setups (Row 1) and complex indoor environments (Row 2) under continuous camera trajectories. This includes wide-angle horizontal yaw rotations (ranging from $-135^\circ$ to $135^\circ$, and even a complete $180^\circ$ reverse-angle view) as well as vertical pitch adjustments (from $-75^\circ$ to $75^\circ$). 

To verify the strict geometric fidelity of the generated perspective sequences, we reconstruct the 3D sparse point clouds and estimate camera frustums from the generated multi-view images using the VGGT pipeline. The successful 3D reconstructions and accurate camera alignments demonstrate that Z-Image-Edit dynamically adapts object scale, depth cues, and background parallax with high spatial integrity, avoiding structural distortion or texture warping.

\subsubsection{Robust Multi-Image ID Consistency}
\begin{figure}[H]
  \centering
  \includegraphics[width=1\textwidth]{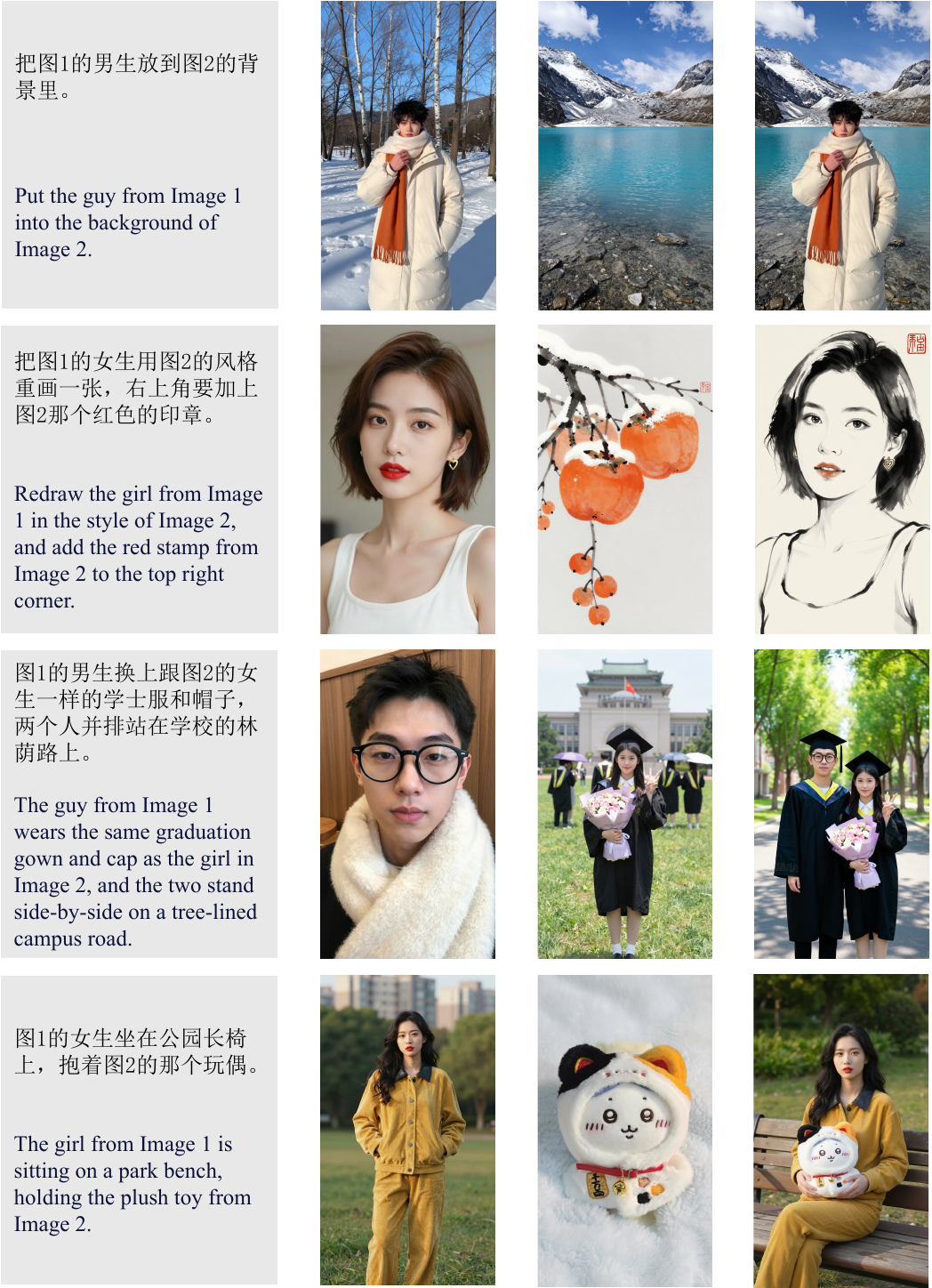}
  \caption{Multi-image editing and identity-preserving composition by Z-Image-Edit. From left to right: prompts, subject source (Img 1), reference source (Img 2), and our edited results. Z-Image-Edit achieves robust ID preservation across background swapping (Row 1), style transfer (Row 2), multi-subject composition (Row 3), and object interaction (Row 4).}
  \label{fig:multi_images}
\end{figure}

Z-Image-Edit overcomes the long-standing challenge of preserving character identity (ID) across stylistic and background transformations by extracting robust identity embeddings from reference images and injecting them into target layouts. As illustrated in Figure~\ref{fig:multi_images}, our model handles diverse multi-image composition paradigms with exceptional fidelity. 

Specifically, in background transplanting (Row 1), a winter-clad character from Image 1 is seamlessly blended into the scenic lake environment of Image 2. In style and element transfer (Row 2), the model redrafts a realistic portrait into a traditional Chinese ink-painting style while precisely extracting and placing a localized red signature stamp. Furthermore, Z-Image-Edit supports multi-subject composition (Row 3), uniting two separate identities into a harmonious campus scene with matched graduation attire. Finally, it facilitates complex subject-object interactions (Row 4), enabling a character to naturally sit on a park bench while holding a specific reference plush toy from the second image.

\subsubsection{Enhanced Reasoning Capacity and World Knowledge through Prompt Enhancer}

\begin{figure}[H]
  \centering
  \includegraphics[width=0.94\textwidth]{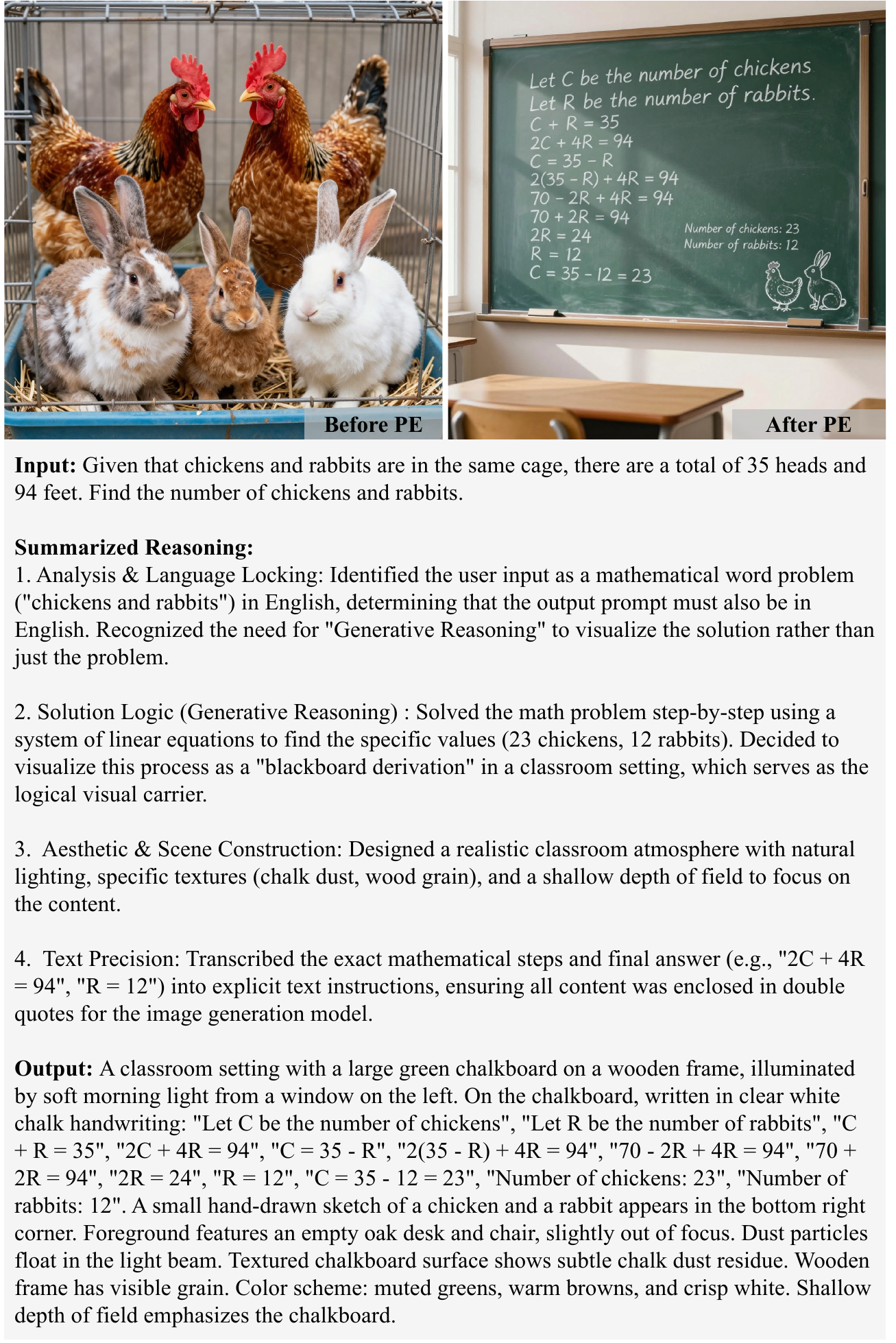}
  \caption{Showcases of prompt enhancer for logical reasoning.}
  \label{fig:pe_case_1}
\end{figure}

\begin{figure}[H]
  \centering
  \includegraphics[width=0.89\textwidth]{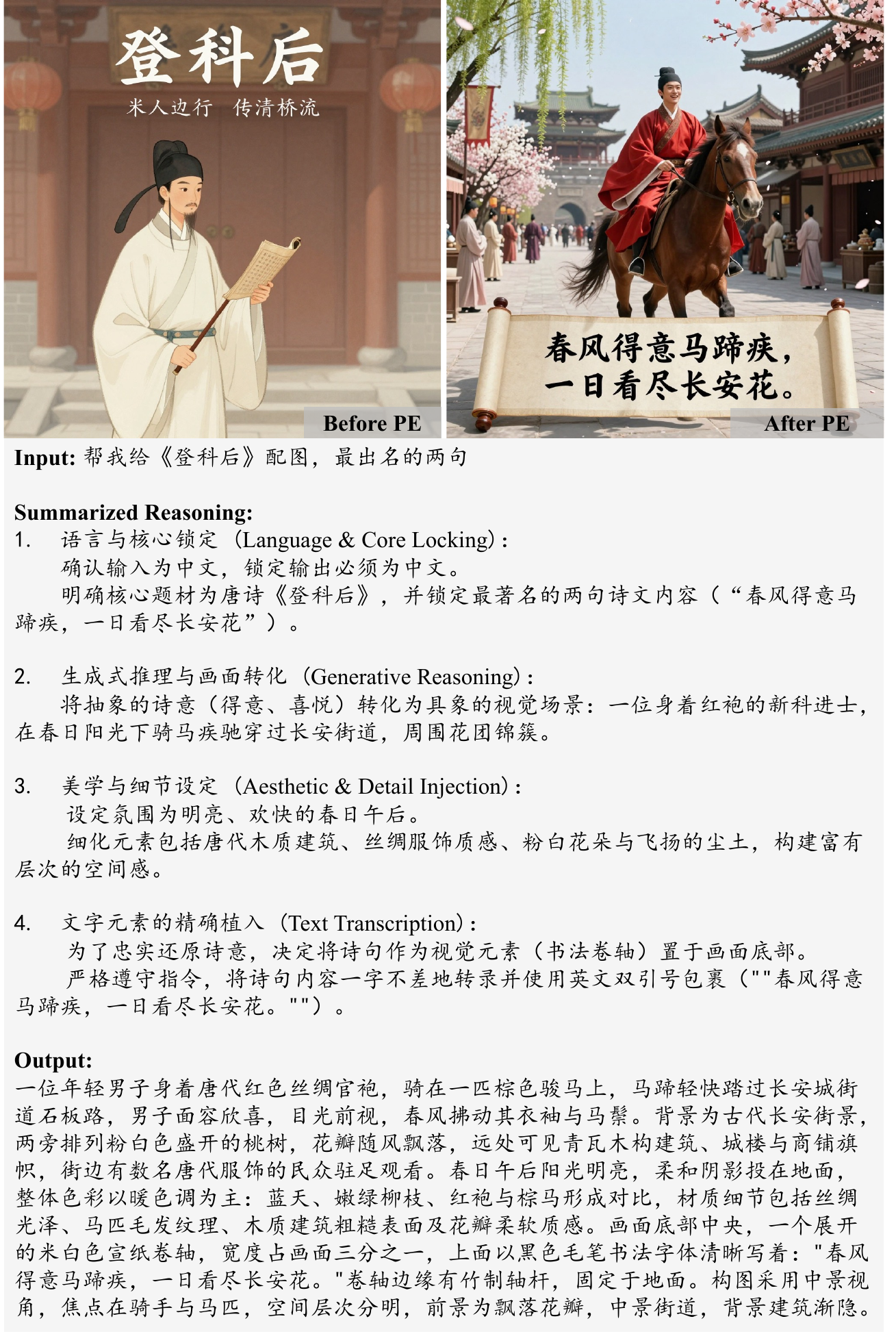}
  \caption{Showcases of prompt enhancer for world knowledge injection. Given the poem title "After Passing the Imperial Examination" (\zh{《登科后》}), the baseline (Left) lacks cultural context. Our method (Right) leverages LLM priors to retrieve specific historical details (e.g., the galloping horse, red official robe) and the famous couplet: "\zh{春风得意马蹄疾，一日看尽长安花。}", the reasoning module (center) translates these literary semantics into visual cues, ensuring a culturally faithful rendering with precise text transcription.}
  \label{fig:pe_case_2}
\end{figure}

As demonstrated in Figure~\ref{fig:t2i_pe} and Figures~\ref{fig:pe_case_1}-\ref{fig:pe_case_2}, our prompt enhancer leverages a structured reasoning chain -- comprising core subject analysis, problem solving/world knowledge injection, aesthetic enhancement, and comprehensive description -- to equip the model with logical reasoning and world knowledge capabilities. This allows the model to handle diverse tasks, ranging from solving complex logical puzzles (\eg, the chicken-and-rabbit problem) and interpreting user intent (\eg, visualizing classical poetry or inferring scenes from coordinates) to performing text rendering and question answering.

\begin{figure}[H]
  \centering
  \includegraphics[width=0.94\textwidth]{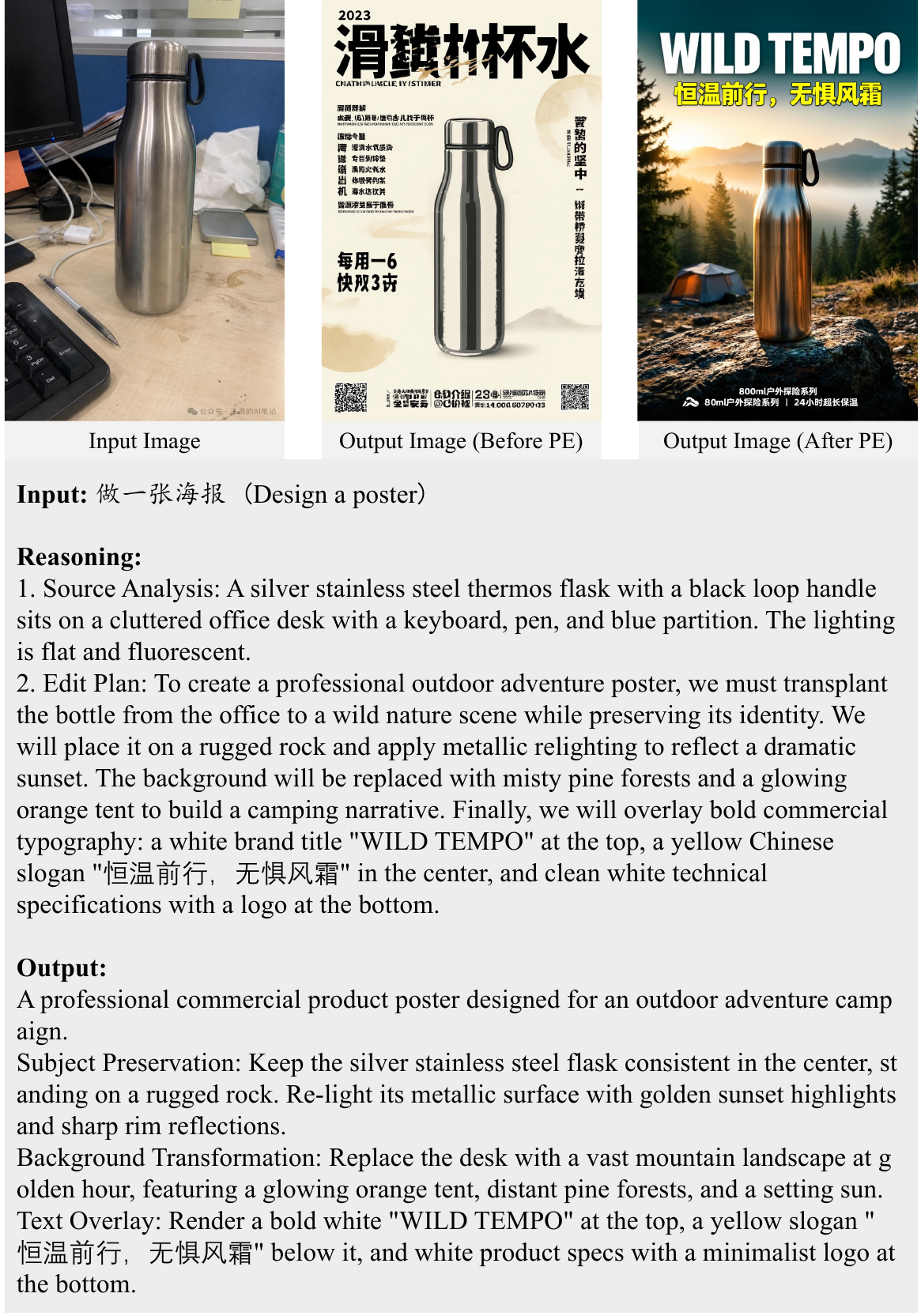}
  \caption{Showcases of prompt enhancer in image editing for handling ambiguous and unclear instructions.}
  \label{fig:edit_pe_case_1}
\end{figure}

\begin{figure}[H]
  \centering
  \includegraphics[width=0.94\textwidth]{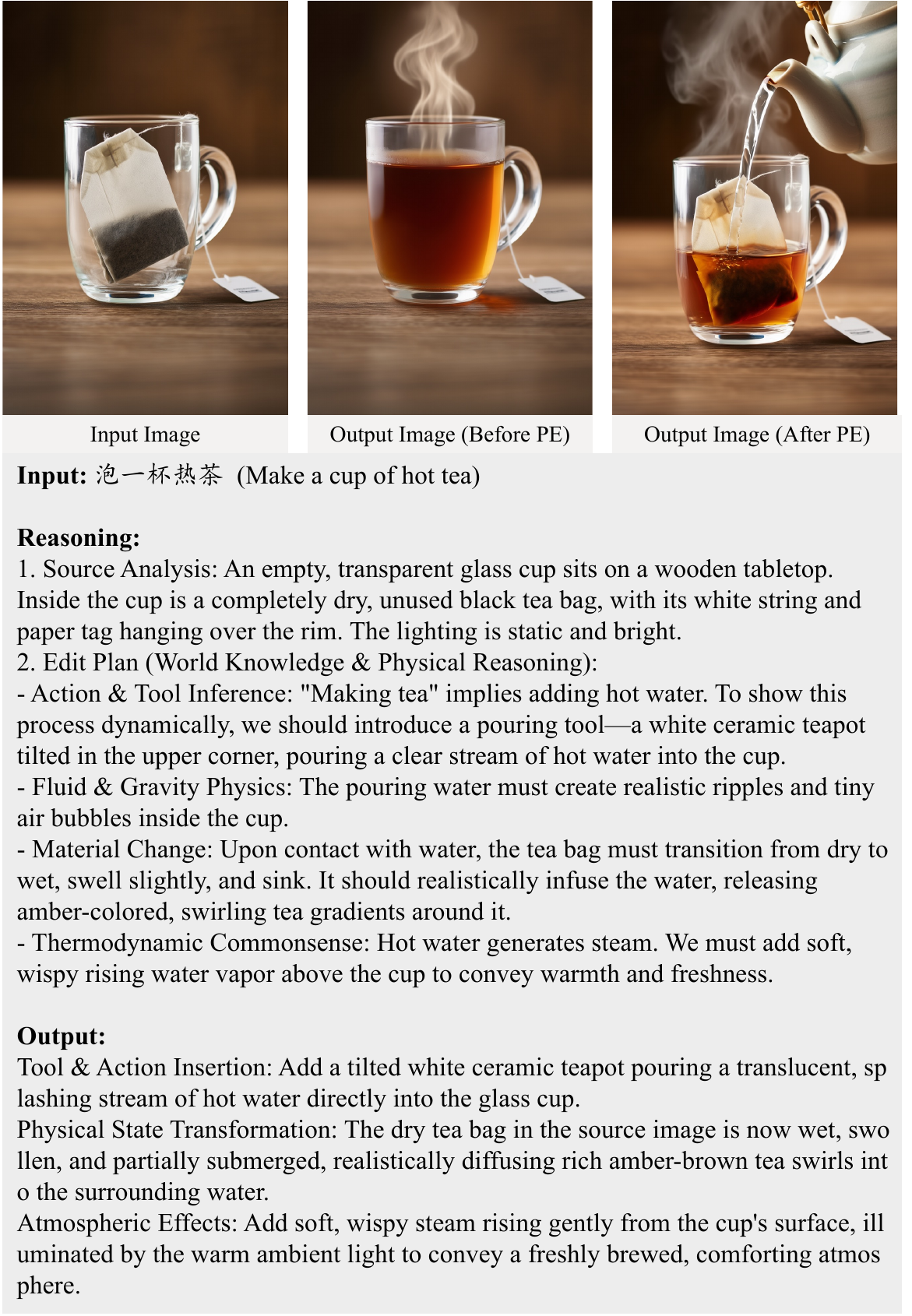}
  \caption{Showcases of prompt enhancer in image editing for world knowledge injection and reasoning.}
  \label{fig:edit_pe_case_2}
\end{figure}

In the context of image editing, a prompt enhancer is also crucial for addressing ambiguous intentions, injecting world knowledge, and enabling physical reasoning, as shown in Figures~\ref{fig:edit_pe_case_1}-\ref{fig:edit_pe_case_2}. For example, in Figure~\ref{fig:edit_pe_case_1}, the ambiguous instruction "Design a poster" yields a chaotic, poorly formatted layout without PE. The prompt enhancer resolves this by generating a structured plan that transplants the product into a professional outdoor scene with thematic lighting and clean commercial typography. Similarly, in Figure~\ref{fig:edit_pe_case_2}, editing without PE lacks the physical reasoning of tea brewing, failing to show the interaction between water and the tea bag. The prompt enhancer corrects this by introducing a teapot pouring water, realistic fluid dynamics, and rising steam.

\subsubsection{Emerging Multi-lingual and Multi-cultural Understanding Capacity}
\begin{figure}[H]
  \centering
  \includegraphics[width=1\textwidth]{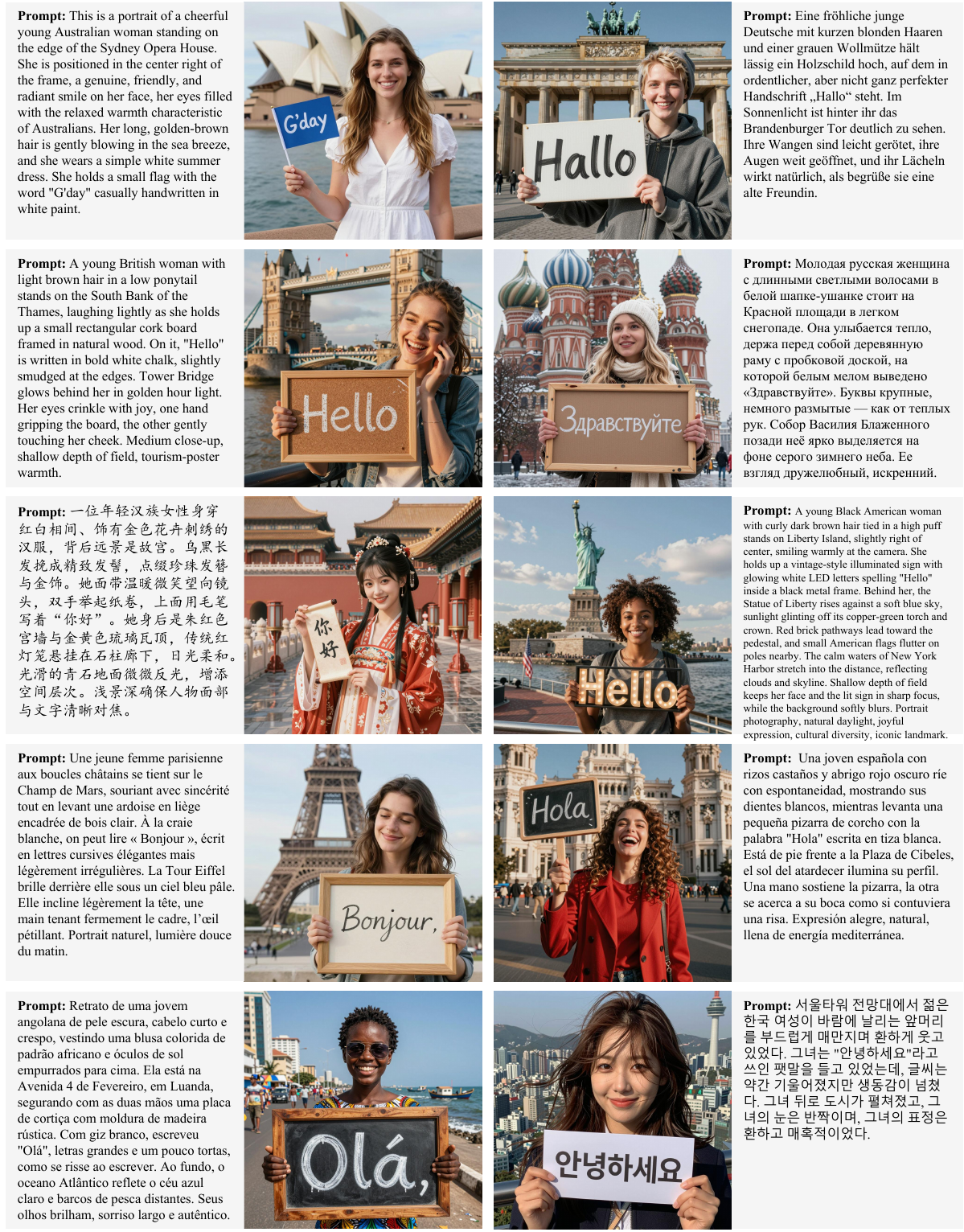}
  \caption{Emerging Multi-lingual and Multi-cultural Understanding Capacity of Z-Image-Turbo. It shows that Z-Image-Turbo can not only understand prompts in multiple languages but also leverage its world knowledge to generate images that align with local cultures and landmarks. }
  \label{fig:mlti_l_c}
\end{figure}

After trained with bilingual data, we are surprised to find that Z-Image has initially emerged with the ability to handle multilingual input. As shown in Figure~\ref{fig:mlti_l_c}, Z-Image can not only understand prompts in multiple languages but also  generate images that align with local cultures and landmarks.

\subsubsection{Emerging Composite Instructions Understanding Capacity}
\begin{figure}[H]
  \centering
  \includegraphics[width=1\textwidth]{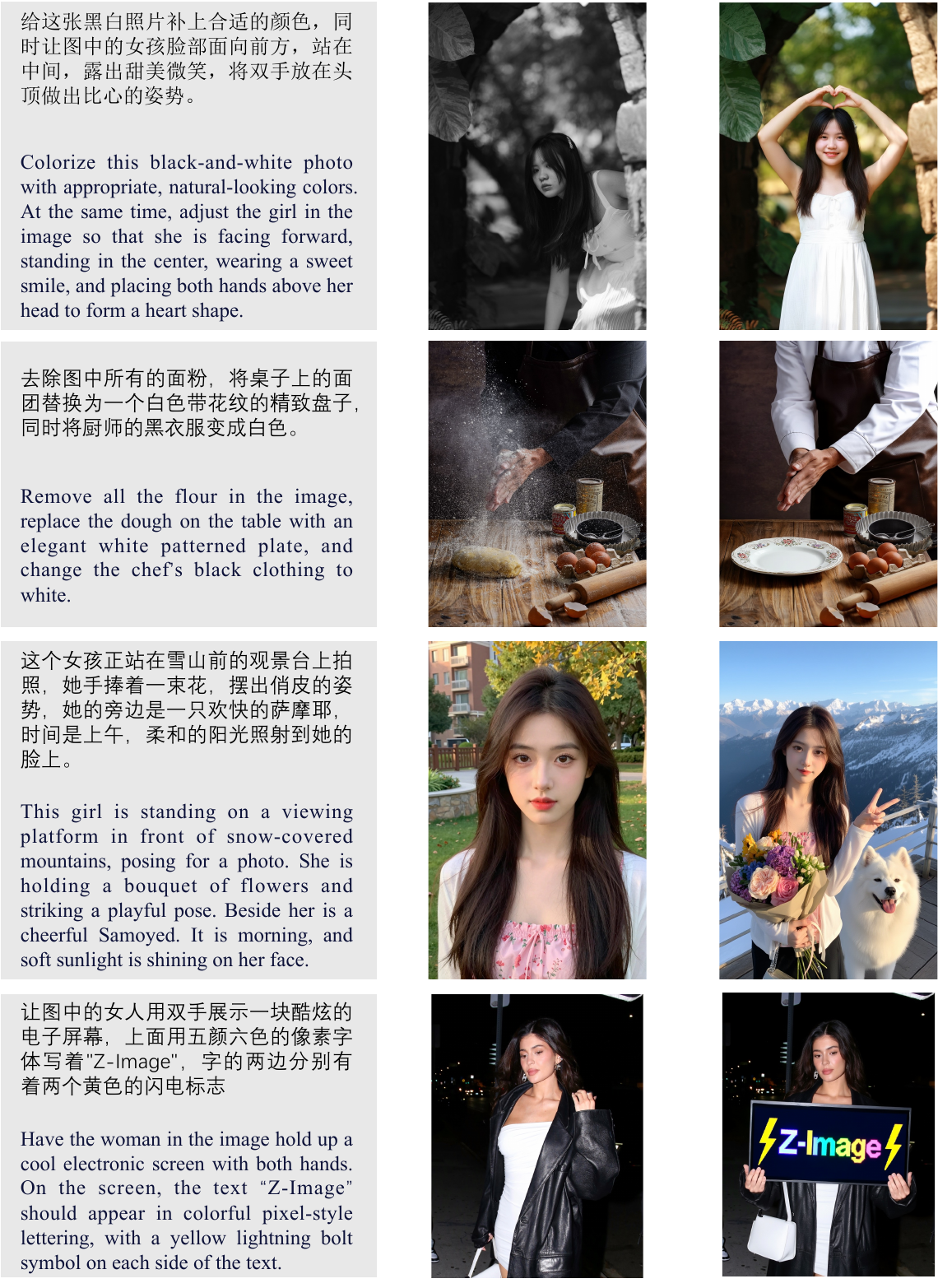}
  \caption{Emerging Understanding of Composite Instructions of Z-Image-Edit. From left to right: input prompts (bilingual), source images, and our edited results. Z-Image-Edit excels at handling multi-attribute modifications, complex identity-preserving scene transformations, and localized text rendering.}
  \label{fig:mixed_editing}
\end{figure}

After edit training, Z-Image-Edit emerges a stronger ability to understand and follow composite instructions expressed in natural language. As demonstrated in Figure~\ref{fig:mixed_editing}, the edited results reflect a more faithful interpretation of the full instruction rather than isolated attribute changes. Specifically, the first two rows show that the model can jointly satisfy multiple dependent edit intents—e.g., colorizing a black-and-white photo while transforming the character’s pose into a heart-shape gesture (Row 1), or removing dynamic elements such as flying flour while simultaneously replacing tabletop objects and changing the clothing color (Row 2).

In addition, Z-Image-Edit exhibits emergent instruction-consistent identity preservation and localized text rendering. As shown in the bottom two rows, it can transplant a close-up portrait into a scenic snowy background while introducing new elements (a Samoyed and flowers) and still strictly preserve facial identity (Row 3). It also renders stylized text (``Z-Image'') onto an inserted screen in a way that matches the subject’s natural interaction and viewpoint (Row 4).


\newpage

\newpage
\newpage

\newpage
\newpage

\newpage
\newpage

\newpage
\newpage

\newpage
\newpage

\newpage
\newpage

\newpage
\newpage

\newpage
\newpage



\newpage
\newpage

\newpage
\newpage

\newpage
\newpage



\newpage
\newpage

\newpage
\newpage

\newpage
\newpage

\newpage
\newpage


\newpage
\newpage

\newpage
\newpage

\section{Conclusion}

In this report, we introduce the \textbf{Z-Image series}, a suite of high-performance 6B-parameter models built upon a Scalable Single-Stream Diffusion Transformer (S3-DiT). Challenging the prevailing ``scale-at-all-costs'' paradigm, we propose a holistic end-to-end solution anchored by four strategic pillars: (1) a curated, efficient data infrastructure; (2) a scalable single-stream architecture; (3) a streamlined training strategy; and (4) advanced optimization techniques for high-quality and efficient inference, encompassing PE-
aware supervised fine-tuning, few-step distillation, and reward post-training.

This synergy allows us to complete the entire workflow \textbf{within 314K H800 GPU hours} at a total cost of \textbf{under \$630K}, delivering top-tier photorealistic synthesis and bilingual text rendering. Beyond the robust base model, our pipeline yields \textbf{Z-Image-Turbo}, which enables \textbf{sub-second inference (<1s)} on an enterprise-grade H800 GPU and fits  \textbf{comfortably within 16G VRAM consumer-grade hardware}. Additionally, we develop \textbf{Z-Image-Edit}, an editing model \textbf{efficiently derived} via our omni-pretraining paradigm. Through this pipeline, we provide the community with a blueprint for developing accessible, budget-friendly, yet state-of-the-art generative models.




\section{Authors}

\subsection[Core Contributors]{Core Contributors\footnote{Core Contributors are listed in alphabetical order of the last name.}}
Huanqia Cai, Sihan Cao, Ruoyi Du, Peng Gao$^{\dagger}$, Aiming Hao$^{\ddagger}$, Steven Hoi, Zhaohui Hou, Dengyang Jiang, Yuming Jiang, Xin Jin, Liangchen Li, Zhen Li, Zhong-Yu Li, David Liu, Dongyang Liu, Qilong Wu, Feng Yu, Zechao Zhan, Chi Zhang, Shifeng Zhang, Ruikai Zhou, Shilin Zhou%
\footnote{Project Leadership: $^{\dagger}$ Project Leader, $^{\ddagger}$ Image Editing Co-Leader.}

\subsection[Contributors]{Contributors\footnote{Contributors are listed in alphabetical order of the last name.}}
Chenglin Cai, Yujing Dou, Yan Gao, Minghao Guo, Songzhi Han, Wei Hu, Xiuhan Hu, Shijie Huang, Yuyan Huang, Haonan Li, Xu Li, Zefu Li, Heng Lin, Jiaming Liu, Linhong Luo, Qingqing Mao, MERJIC, Jingyuan Ni, Chuan Qin, Lin Qu, Junhan Shi, Jinghua Sun, Peng Wang, Ping Wang, Shanshan Wang, Xuecong Wang, Yi Wang, Yue Wang, Tingkun Wen, Junde Wu, Minggang Wu, Xiongwei Wu, Yi Xin, Haibo Xing, Jingjie Xu, Xiaoxiao Xu, Xiaoyu Xu, Ze Xu, Xunliang Yang, Wentao Ye, Shuting Yu, Yucheng Zhao, Jianan Zhang, Jianfeng Zhang, Jiawei Zhang, Qiang Zhang,  Xudong Zhao, Yu Zheng, Haijian Zhou, Hanzhang Zhou





\clearpage

\addcontentsline{toc}{section}{References}
\bibliography{s3_report}
\bibliographystyle{plain}

\appendix
\section{Prompts Used in the Report}
Here we summarize the prompts/instructions used in Figure~\ref{fig:showcase_realistic}-\ref{fig:showcase_editing}, which can be directly input into Z-Image-Turbo (with PE disabled) to reproduce our generation results.

\subsection{Figure 1}
\label{sec:fig_1}
\textbf{Column \#1}
\begin{itemize}
    \item Case \#1: 

    \zh{一张中景手机自拍照片拍摄了一位留着长黑发的年轻东亚女子在灯光明亮的电梯内对着镜子自拍。她穿着一件带有白色花朵图案的黑色露肩短上衣和深色牛仔裤。她的头微微倾斜，嘴唇嘟起做亲吻状，非常可爱俏皮。她右手拿着一部深灰色智能手机，遮住了部分脸，后置摄像头镜头对着镜子。电梯墙壁由抛光不锈钢制成，反射着头顶的荧光灯和主体。左侧墙上有一个带有许多圆形按钮和小型数字显示屏的垂直面板。在按钮下方，可以看到一个金属扶手。后墙上贴着带有文字的长方形标志。地面铺着带有白色纹理的深色大理石瓷砖。整体照明为人造光，明亮，具有电梯内部的特征。}

    (Translation: A mid-range phone selfie captured a young East Asian woman with long black hair taking a selfie in front of a mirror in a brightly lit elevator. She was wearing a black off shoulder short top with a white floral pattern and dark jeans. Her head tilted slightly, and her lips curled up in a kiss, very cute and playful. She held a dark gray smartphone in her right hand, covering part of her face, with the rear camera facing the mirror. The elevator walls are made of polished stainless steel, reflecting the fluorescent lights and main body above the head. There is a vertical panel on the left wall with many circular buttons and a small digital display screen. Below the button, you can see a metal armrest. There is a rectangular sign with text on the back wall. The ground is covered with dark marble tiles with white texture. The overall lighting is artificial light, bright, and has the characteristics of an elevator interior.)
    
    \item Case \#2: 

    \zh{一张充满动感的运动摄影照片，捕捉到一名橄榄球运动员在比赛中奔跑的瞬间，他正向右冲刺，左手抱着橄榄球。中心人物是一名30多岁的黑人男性橄榄球运动员，身穿绿白条纹球衣，球衣正面有“Turbo”字样，胸口白色小字清晰的写着“MAI”，没有其他明显logo，白色袜子带有绿色条纹，脚穿白色钉鞋。他的左侧，两名身穿黑色和黄色条纹球衣的对手正朝相反方向奔跑，两个人球衣正面分别写着有“Base”和“Edit”字样，搭配黄色袜子和白色钉鞋，均略微失焦。背景是模糊的体育场观众席，观众穿着各色服装，还有蓝色和白色的体育场座椅，以及一块橙色广告牌，上面有部分可见的白色文字“Tongyi Lab”和一个黄色圆形标志。前景是维护良好的绿色橄榄球场。动作摄影，体育摄影，浅景深，中心球员对焦清晰，背景虚化，自然光照明，色彩鲜艳，高对比度，构图动感，对手球员有运动模糊，充满活力，竞技氛围，户外体育场场景。}

    (Translation: A dynamic sports photography shot capturing a rugby player running during a game, sprinting to the right with a rugby ball tucked under his left arm. The central figure is a Black male rugby player in his 30s, wearing a green and white striped jersey with text "Turbo" on it, the white text "MAI" clearly visible on the chest with no other prominent logos, white socks with green stripes, and white cleats. To his left, two opponents in black and yellow striped jerseys are running in the opposite direction, their jerseys displaying "Base" and "Edit" on the front, paired with yellow socks and white cleats, both slightly out of focus. The background shows a blurred stadium crowd wearing various colored clothing, blue and white stadium seats, and an orange advertising board with partially visible white text "Tongyi Lab" and a yellow circular logo. The foreground features a well-maintained green rugby field. Action photography, sports photography, shallow depth of field, central player in sharp focus, blurred background, natural lighting, vibrant colors, high contrast, dynamic composition, motion blur on opponent players, energetic, competitive atmosphere, outdoor stadium scene.)
    
    \item Case \#3: 

    \zh{一张广角、平视角度的照片捕捉到了一个充满活力的街景，地点是一条铺着不平整鹅卵石的狭窄古老小巷。小巷两旁是两到三层的红砖建筑，具有传统建筑特色的深色木门、窗框和悬挑的楼层。在左边，两名女子站在一片阳光下交谈。一名女子在格子衬衫和深色裤子外套着红色围裙，另一名女子则披着深色披肩。一只黑色小狗躺在她们脚下温暖的石头上。在前景中，一只体型较大、有着蓬松卷曲尾巴的金色狗正在嗅探鹅卵石路面。沿着小巷的中心再往前，一个人正骑着一辆小型摩托车远去，另一只黑色小狗则坐在街道的右侧。明亮的阳光和深邃的阴影在整个场景中形成鲜明对比，突显了砖块和石头的纹理。小巷尽头可见的天空是苍白的阴白色。}

    (Translation: A wide-angle, head up photo captures a vibrant street scene in a narrow ancient alley paved with uneven pebbles. On both sides of the alley are two to three story red brick buildings with traditional architectural features such as dark wooden doors, window frames, and cantilevered floors. On the left, two women are standing in a sunny area talking. A woman is wearing a red apron over a checkered shirt and dark pants, while another woman is draped in a dark shawl. A small black dog lay on the warm stone beneath their feet. In the foreground, a large golden dog with a fluffy and curly tail is sniffing the cobblestone road surface. Continuing along the center of the alley, a person is riding a small motorcycle away, while another black puppy is sitting on the right side of the street. The bright sunlight and deep shadows create a sharp contrast throughout the scene, highlighting the texture of the bricks and stones. The sky visible at the end of the alley is pale and gloomy white.)
    
    \item Case \#4: 

    \zh{一张宁静的、全景横向照片，捕捉了一个小孩侧身站在郁郁葱葱的绿色草岸上，旁边是平静的水体。场景设置在黄金时刻，很可能是日落时分，色调柔和而温暖。孩子位于画面左侧，戴着一顶浅色的编织草帽，在浅蓝白格纹长袖衬衫外穿着一件橄榄绿色的短袖背心，下身是宽松的深蓝色牛仔裤，裤脚卷起，露出棕色的鞋子。孩子的右手拿着一朵黄色小花的茎，左手提着一个银色的小号镀锌金属喷壶。他/她正朝右边望去，看向水面。前景是点缀着黄色小野花的草坡。中景是河流或池塘的静水，倒映着天空温暖的粉橙色调。对岸有绿色植被和一堆灰色岩石。背景是柔和模糊的，展现了广阔的绿色田野、远处的树林线以及一些模糊的建筑轮廓，这一切都在一片广阔的天空下，天空布满了柔和的、带有粉色和橙色渐变的云彩。摄影风格的特点是浅景深，在背景中创造了显著的散景效果，使主体突出。光线自然而漫射，营造出平和、田园诗般和怀旧的氛围。孩子的脚被画面的底部边缘轻微截断。}

    (Translation: A peaceful, panoramic horizontal photo captures a child standing sideways on a lush green grassy bank, with a calm body of water beside it. The scene is set in prime time, most likely at sunset, with soft and warm tones. The child is located on the left side of the screen, wearing a light colored woven straw hat, an olive green short sleeved vest over a light blue and white checkered long sleeved shirt, and loose dark blue jeans with rolled up hemlines revealing brown shoes. The child holds a stem of a small yellow flower in their right hand and a silver small galvanized metal spray can in their left hand. He/she is looking to the right, towards the water surface. The prospect is a grassy slope adorned with small yellow wildflowers. The central view is the still water of a river or pond, reflecting the warm pink orange tones of the sky. There is green vegetation and a pile of gray rocks on the opposite bank. The background is soft and blurry, showing vast green fields, distant forest lines, and some blurry building contours, all under a vast sky filled with soft, pink and orange gradient clouds. The characteristic of photography style is shallow depth of field, which creates a significant bokeh effect in the background, making the subject stand out. The natural and diffuse light creates a peaceful, pastoral, and nostalgic atmosphere. The child's feet were slightly cut off by the bottom edge of the screen.)
    
    \item Case \#5: 

    \zh{一张广角风景照片，拍摄于阴天的安徽宏村古村落。画面被平静水体的岸线水平分割，形成了村庄与天空近乎完美的镜面倒影。在中景部分，一簇密集的传统徽派建筑沿水边排列，具有独特的白墙和深灰色瓦顶。几栋建筑的屋檐下悬挂着红色的纸灯笼，在柔和的背景中增添了鲜艳的色彩点缀。水边的石板路上和房屋之间散布着许多光秃秃的落叶树。一些身影微小的人沿着这条小路行走或坐着。在背景中，一片朦胧的蓝绿色山脉在淡灰色的天空下连绵起伏。右侧山坡上可以看到一个小型输电塔。在画面中心偏右的一栋建筑上，门楣上方挂着一块横向的木匾，上面有黑色的汉字“世德堂”。该摄影作品的风格特点是构图对称，光线柔和漫射，景深较大，整个场景都清晰锐利，色调清冷而宁静，以白色、灰色和蓝色为主，红色作为强烈的点缀色。整体氛围平和、安详且具有永恒感。}

    (Translation: A wide-angle landscape photo taken on a cloudy day in the ancient village of Hongcun, Anhui. The screen is horizontally divided by the calm shoreline of the water, forming a nearly perfect mirror reflection of the village and the sky. In the central area, a dense cluster of traditional Huizhou style buildings are arranged along the water's edge, featuring unique white walls and dark gray tiled roofs. Red paper lanterns hang under the eaves of several buildings, adding vibrant color accents to the soft background. There are many bare deciduous trees scattered between the stone roads and houses by the water's edge. Some small figures walked or sat along this path. In the background, a hazy blue-green mountain range undulates continuously under a light gray sky. A small transmission tower can be seen on the right slope. On a building to the right of the center of the screen, there is a horizontal wooden plaque hanging above the lintel, with black Chinese characters "\zh{世德堂}" on it. The stylistic features of this photography work are symmetrical composition, soft and diffuse lighting, large depth of field, clear and sharp entire scene, cool and peaceful tones, mainly white, gray and blue, with red as a strong accent color. The overall atmosphere is peaceful, serene, and has a sense of eternity.)
    
    \item Case \#6: 

    \zh{一张充满活力的广角夜景照片，捕捉了中国广州猎德大桥上空壮观的烟花表演。场景设置在漆黑的夜空下，被多朵巨大的烟花爆炸瞬间照亮。烟花主要为白色和红色/粉色，在画面的上半部分形成了灿烂的牡丹花状图案，周围环绕着浓浓的硝烟。猎德大桥，一座现代化的斜拉桥，在中景处横跨珠江。其独特的拱形中央桥塔被温暖的黄色灯光照亮。在这个中央桥塔的正面，可以看到一个被部分遮挡的红色小标志。桥面也被路灯照亮。在前景中，黑暗的江水映照出烟花和桥灯的缤纷倒影。左下角可以看到一艘小船的黑色剪影，远处还散布着其他更小的船只。背景是闪闪发光的现代化城市天际线，摩天大楼和其他建筑上的无数灯光点缀其间。该摄影风格以长曝光为特点，这从烟花的轨迹中可以明显看出，营造出一种动感和喜庆的氛围。图像对比度高，对焦清晰，在黑暗的环境中呈现出鲜艳的色彩。}

    (Translation: A vibrant wide-angle night view photo captures the spectacular fireworks display over the Liede Bridge in Guangzhou, China. The scene is set in the pitch black night sky, instantly illuminated by multiple huge fireworks explosions. The fireworks are mainly white and red/pink, forming a brilliant peony shaped pattern in the upper part of the picture, surrounded by thick gunpowder smoke. Liede Bridge, a modern cable-stayed bridge, crosses the the Pearl River in the middle view. Its unique arched central bridge tower is illuminated by warm yellow lights. On the front of this central bridge tower, a partially obscured red small sign can be seen. The bridge deck is also illuminated by streetlights. In the foreground, the dark river reflects the colorful reflections of fireworks and bridge lights. In the lower left corner, a black silhouette of a small boat can be seen, with other smaller boats scattered in the distance. The background is a sparkling modern city skyline, adorned with countless lights from skyscrapers and other buildings. This photography style is characterized by long exposures, which can be clearly seen from the trajectory of fireworks, creating a dynamic and festive atmosphere. The image has high contrast, clear focus, and presents bright colors in dark environments.)
    
\end{itemize}

\textbf{Column \#2}
\begin{itemize}
    \item Case \#1: 

    {A stylish young woman sits casually on an unmade bed bathed in soft daylight, wearing a pastel yellow oversized T-shirt with subtle white text and cozy light gray sweatpants. Her skin glows fresh beneath glossy deep lavender hydrogel under-eye patches, while her hair is tied back loosely with a scrunchie, complemented by delicate gold hoop earrings. Nearby, a tube of hand cream and an open laptop rest casually atop soft, slightly rumpled sheets. The natural window light gently illuminates her radiant skin and the subtle sheen of the hydrogel patches, enhancing the cozy textures of her loungewear and bedding. Shot from a top-down selfie angle, the framing captures her face, shoulders, and upper torso with realistic iPhone grain, conveying an authentic, relaxed self-care morning moment in a softly lit bedroom scene -- skincare selfie, shot on iPhone.}

    \item Case \#2: 
    
    \zh{一张逼真的年轻东亚女性肖像，位于画面中心偏左的位置，带着浅浅的微笑直视观者。她身着以浓郁的红色和金色为主的传统中式服装。她的头发被精心盘起，饰有精致的红色和金色花卉和叶形发饰。她的眉心之间额头上绘有一个小巧、华丽的红色花卉图案。她左手持一把仿古扇子，扇面上绘有一位身着传统服饰的女性、一棵树和一只鸟的场景。她的右手向前伸出，手掌向上，托着一个悬浮的发光的霓虹黄色闪电亚克力灯牌，这是画面中最亮的元素。背景是模糊的夜景，带有暖色调的人工灯光，一场户外文化活动或庆典。在远处的背景中，她头部的左侧略偏，是一座高大、多层、被暖光照亮的西安大雁塔。中景可见其他模糊的建筑和灯光，暗示着一个繁华的城市或文化背景。光线是低调的，闪电符号为她的脸部和手部提供了显著的照明。整体氛围神秘而迷人。人物的头部、手部和上半身完全可见，下半身被画面底部边缘截断。图像具有中等景深，主体清晰聚焦，背景柔和模糊。色彩方案温暖，以红色、金色和闪电的亮黄色为主。}

    (Translation: A realistic portrait of a young East Asian woman, located to the left of the center of the image, looking directly at the viewer with a faint smile. She was dressed in traditional Chinese clothing dominated by rich red and gold colors. Her hair was carefully styled, adorned with delicate red and gold flowers and leaf shaped hair accessories. There is a small and gorgeous red floral pattern painted on her forehead between her eyebrows. She held an antique style fan in her left hand, with a scene of a woman dressed in traditional clothing, a tree, and a bird painted on the fan surface. Her right hand extended forward, palm up, holding a suspended glowing neon yellow lightning acrylic light tag, which was the brightest element in the picture. The background is a blurry night scene with warm toned artificial lighting, representing an outdoor cultural event or celebration. In the distant background, to the left of her head is a tall, multi-layered, warm lit Xi'an Big Wild Goose Pagoda. Other blurry buildings and lights can be seen in the middle of the scene, implying a bustling city or cultural background. The light is low-key, and the lightning symbol provides significant illumination for her face and hands. The overall atmosphere is mysterious and charming. The head, hands, and upper body of the character are fully visible, while the lower body is cut off by the bottom edge of the screen. The image has a moderate depth of field, the subject is clearly focused, and the background is soft and blurry. The color scheme is warm, with red, gold, and bright yellow of lightning as the main colors.)
    
    \item Case \#3: 

    {A full-body, eye-level photograph of a young, beautiful East Asian woman posing cheerfully inside a brightly lit LEGO store or brand exhibition space. The woman, positioned slightly right of center, has long dark hair and is smiling at the camera. She wears a vibrant yellow ribbed beanie, a white diamond-quilted puffer jacket over a white t-shirt, and medium-wash blue jeans with cuffs rolled up at the ankles. She is wearing white lace-up sneakers and white socks, with a small red heart visible on her left sock. In her left hand, she holds a black structured handbag. Her pose is playful, with her left leg kicked up behind her. To her left is a large, multi-tiered display stand in bright yellow, which features the official LEGO logo -- white text in a red square with a black and yellow outline -- in the upper left corner. On this stand are two large-scale LEGO Minifigure statues: a policeman in a blue uniform and hat stands in the foreground, and behind him is a Santa Claus figure in red. The background shows more yellow retail shelving stocked with various LEGO sets and products. The floor is made of large, light grey tiles, and a white dome security camera is visible on the ceiling. The image is a sharp, well-lit snapshot with a vibrant color palette, dominated by yellow, red, and blue, creating a joyful and commercial atmosphere.}
    
    \item Case \#4: 

    {A candid mid-2010s-style snapshot featuring a pale young woman with icy platinum hair styled casually loose, seated on a metal bench inside a monochrome concept store. She wears a huge black hoodie, sheer tights, and maroon platform creepers, complemented by a beanie embroidered with “Z-Image Real \& Fast” The subject’s relaxed expression gazes off to the side, conveying subtle, ambiguous emotion. The lighting is cold and matte with soft shadows stretching along a wooden floor, intentionally exhibiting muted color saturation, softened contrast, and distinctly cool-toned bluish-gray shadows. Visible textures include realistic skin details, detailed fabric grain of the hoodie and tights, individual icy hair strands, and clear accessory textures. The framing is slightly off-center and casually tilted, capturing spontaneous intimacy and informal snapshot aesthetics characteristic of mid-2010s casual youth photography.}
    


    
    
    
    
    
\end{itemize}

\textbf{Column \#3}
\begin{itemize}

    
    \item Case \#1: 

    \zh{一位男士和他的贵宾犬穿着配套的服装参加狗狗秀，室内灯光，背景中有观众。}

    (Translation: A man and his poodle participated in a dog show wearing matching costumes, with indoor lighting and an audience in the background.)

    \item Case \#2: 

    \zh{一张特写、逼真的东亚婴儿肖像，婴儿穿着一件印有心形图案的奶油色蓬松冬季连体衣，直视观者。婴儿拥有深色头发和红扑扑的脸颊。婴儿手边部分可见一个色彩鲜艳的玩具，背景模糊处有一位穿着格子衬衫的人。室内光线具有柔和的阴影和高光，营造出温暖的色调，婴儿脸部清晰聚焦，背景柔和模糊。低饱和度、颗粒感、老胶片风格。}

    (Translation: A close-up, realistic portrait of an East Asian baby wearing a creamy fluffy winter jumpsuit with a heart-shaped pattern, looking straight at the viewer. Babies have dark hair and rosy cheeks. A brightly colored toy can be seen near the baby's side, with a person wearing a checkered shirt in a blurry background. The indoor lighting features soft shadows and highlights, creating a warm tone. The baby's face is clearly focused, and the background is soft and blurry. Low saturation, graininess, and vintage film style.)
    
    \item Case \#3: 

    \zh{北京国家体育场（鸟巢）的照片，蓝天背景下，体育场的外观由复杂的交织钢结构形成网状图案主导。前景中一个人穿着休闲装，略微偏中心位置行走。背景通过钢结构可以看到体育场内部的红色座位区。“A30”用红色标记在钢结构的左下角。图像从低角度拍摄，突显建筑的宏伟和规模。照片，高对比度，戏剧性光线，蓝天，低角度视角，建筑摄影，聚焦清晰，现代设计，精细钢结构，鲜艳红色点缀，视觉冲击力强，构图平衡。}

    (Translation: A photo of the Beijing National Stadium (Bird's Nest), with a blue sky background, the appearance of the stadium is dominated by a complex interwoven steel structure forming a mesh pattern. In the foreground, a person is wearing casual clothing and walking slightly off center. The background shows the red seating area inside the stadium through the steel structure. "A30" is marked in red on the bottom left corner of the steel structure. The image is taken from a low angle to highlight the grandeur and scale of the building. Photos, high contrast, dramatic lighting, blue sky, low angle perspective, architectural photography, clear focus, modern design, fine steel structure, bright red accents, strong visual impact, balanced composition.)
    
\end{itemize}

\subsection{Figure 2}
\label{sec:fig_2}
\textbf{Row \#1}
\begin{itemize}
    \item Case \#1: 

    \zh{杂志封面设计。 文案：大标题“「造相」Z-Image”。 小标题：“Winter Release. Spring for Generative Art.”。 版本号：“ VOL 1.0”。 中间底部极小字“通义多模态交互出版社”。 拉开一片白雪茫茫下的拉链，拉链下漏出绿草鲜花的春天，移轴微距，拉链是一个冒着白烟远去的火车头，精美构图，夸张的俯视视角，视觉冲击力，高对比度，高饱和度。}

    (Translation: Magazine cover design. Copy: Headline "\zh{「造相」Z-Image}". Subtitle: "Winter Release. Spring for Generative Art. Version number: "VOL 1.0". The extremely small font at the bottom of the middle reads '\zh{通义多模态交互出版社}'. Pulling open a zipper under a vast expanse of white snow, the spring of green grass and flowers peeks out from under the zipper. Moving the axis macro, the zipper is a locomotive emitting white smoke far away, with exquisite composition, exaggerated top-down perspective, visual impact, high contrast, and high saturation.)
    
    \item Case \#2: 

    \zh{一幅垂直构图、风格化的数字插画，设计为一张励志海报。场景描绘了夜间的沙漠景观，头顶是广阔无垠、繁星密布的天空，其中银河清晰可见。前景和中景以深蓝色近乎黑色的剪影为特色。左侧，一棵巨大而细节丰富的约书亚树剪影占据了画面主导。更远处可以看到两棵较小的约书亚树。右侧，两个人的剪影站在一个小山丘上，仰望着天空。天空从底部的深海军蓝过渡到顶部的浅蓝色，布满星辰，明亮的银河带以柔和的白色、紫色和蓝色调，从右上角划过。图像上覆盖有五处独立的渲染中文字样。顶部是白色大号字体，内容为“于无垠黑暗中，寻见你的微光”。在中间靠近人物的位置，有较小的黑色字体写着“心之所向，宇宙回响”。在最底部，是白色大号艺术字的主标题“仰望·逐梦”，其下方是稍小的白色字体“心的旅程由此开始”。在山丘上靠近人物的地方，有一个非常小、几乎隐藏的黑色签名“观星者”。整体风格图形化且简约，将扁平的剪影与细节更丰富、富有绘画感的天空相结合，营造出一种深沉、引人深思且充满希望的氛围。}

    \zh{(Translation: A vertically composed and stylized digital illustration designed as an inspirational poster. The scene depicts a desert landscape at night, with a vast and starry sky overhead, among which the Milky Way is clearly visible. The foreground and middle ground are characterized by deep blue and almost black silhouettes. On the left, a large and detailed silhouette of a Joshua tree dominates the scene. Two smaller Joshua trees can be seen further away. On the right, silhouettes of two people stand on a small hill, looking up at the sky. The sky transitions from deep sea navy blue at the bottom to light blue at the top, filled with stars, and the bright Milky Way streaks across in soft white, purple, and blue tones from the top right corner. The image is covered with five independent rendered Chinese characters. At the top is a large white font that reads '于无垠黑暗中，寻见你的微光'. In the middle, near the character, there is a small black font that reads '心之所向，宇宙回响'.  At the bottom, there is the main title "仰望·逐梦" in large white artistic font, and below it is a slightly smaller white font "心的旅程由此开始". Near the character on the hill, there is a very small, almost hidden black signature called '观星者'. The overall style is graphical and minimalist, combining flat silhouettes with a more detailed and picturesque sky, creating a deep, thought-provoking, and hopeful atmosphere.)}
    
    \item Case \#3: 

    \zh{一张充满活力的视觉作品集平面设计海报，整张图片以非常小的透明棋盘格为背景，展示了一个3D渲染的卡通人物。画面左侧是一位年轻女性的半身像，她皮肤白皙，留着深棕色长卷发，戴着粉色边框的眼镜，眼镜后是棕色的大眼睛。她笑容灿烂，露出牙齿，戴着小巧的银色耳钉。她的着装包括一件浅灰色西装外套、一件白色翻领衬衫和一条红色领带。她手中捧着一束由四朵鲜艳的黄色向日葵组成的花束，花茎为绿色。该角色被一圈粗白的轮廓线包围，使其从背景中凸显出来。海报的右侧主要是大型艺术字。主标题“视觉作品集”采用粗大的黄色笔刷风格字体。其上叠加着一行纤细的红色草书英文“Personalization”。下方是圆润气泡状的黄色小一号字体“VISUAL PORTFOLIO”。
    其下写出了三个亮点：
    “·中英渲染，字字如刻
    ● Bilingual Rendering”
    “·不止真实，更懂美学
    ● Realism \& Aesthetic”
    “· 读懂复杂，生成精妙
    ● Complexity \& Elegance”
    这里中文是白色手写体大字，英文是半透明的印刷体小字。海报包含多个文本块和标志。中上部先是黄色的文字“Z-Image x”，中间是一个戴着耳机的卡通头像的黄色线条画标志，后面跟着文字“x Design”。在右下角有一个可爱的拟人化扩音器，它有两只大大的眼睛，颜色为浅绿色和奶油色，底部有一朵小雏菊。整体风格是3D角色渲染和平面设计的结合，特点是氛围愉快、对比度高，并采用了黄、黑、白为主的配色方案。}

    (Translation: A vibrant visual portfolio graphic design poster, with a very small transparent checkerboard pattern as the background, showcasing a 3D rendered cartoon character. On the left side of the screen is a half body portrait of a young woman with fair skin, long curly dark brown hair, wearing pink framed glasses, and brown big eyes behind the glasses. She had a bright smile, revealing her teeth and wearing small silver earrings. Her attire includes a light gray suit jacket, a white collared shirt, and a red tie. She held a bouquet of four bright yellow sunflowers in her hand, with green stems. The character is surrounded by a thick white outline, making it stand out from the background. On the right side of the poster are mainly large artistic characters. The main title "Visual Works Collection" adopts a thick yellow brush style font. On top of it is a thin line of red cursive English word "Personalization". Below is a round, bubble shaped yellow font with one size smaller reading 'VISUAL PORTFOLIO'.
Below are three highlights:
“\zh{·中英渲染，字字如刻
    ● Bilingual Rendering”
    “·不止真实，更懂美学
    ● Realism \& Aesthetic”
    “· 读懂复杂，生成精妙
    ● Complexity \& Elegance}”
The Chinese characters here are white handwritten large characters, while the English characters are semi transparent printed small characters. The poster contains multiple text blocks and logos. The upper part is first marked with yellow text "Z-Image x", followed by a yellow line drawn logo of a cartoon avatar wearing headphones, and then the text "x Design". In the bottom right corner, there is a cute anthropomorphic amplifier with two big eyes in light green and cream colors, and a small daisy at the bottom. The overall style is a combination of 3D character rendering and graphic design, characterized by a pleasant atmosphere, high contrast, and predominantly yellow, black, and white color schemes.)
\end{itemize}

\textbf{Row \#2}
\begin{itemize}
    \item Case \#1: 

    \zh{一张虚构的英语电影《回忆之味》（The Taste of Memory）的电影海报。场景设置在一个质朴的19世纪风格厨房里。画面中央，一位红棕色头发、留着小胡子的中年男子（演员阿瑟·彭哈利根饰）站在一张木桌后，他身穿白色衬衫、黑色马甲和米色围裙，正看着一位女士，手中拿着一大块生红肉，下方是一个木制切菜板。在他的右边，一位梳着高髻的黑发女子（演员埃莉诺·万斯饰）倚靠在桌子上，温柔地对他微笑。她穿着浅色衬衫和一条上白下蓝的长裙。桌上除了放有切碎的葱和卷心菜丝的切菜板外，还有一个白色陶瓷盘、新鲜香草，左侧一个木箱上放着一串深色葡萄。背景是一面粗糙的灰白色抹灰墙，墙上挂着一幅风景画。最右边的一个台面上放着一盏复古油灯。海报上有大量的文字信息。左上角是白色的无衬线字体“ARTISAN FILMS PRESENTS”，其下方是“ELEANOR VANCE”和“ACADEMY AWARD® WINNER”。右上角写着“ARTHUR PENHALIGON”和“GOLDEN GLOBE® AWARD WINNER”。顶部中央是圣丹斯电影节的桂冠标志，下方写着“SUNDANCE FILM FESTIVAL GRAND JURY PRIZE 2024”。主标题“THE TASTE OF MEMORY”以白色的大号衬线字体醒目地显示在下半部分。标题下方注明了“A FILM BY Tongyi Interaction Lab”。底部区域用白色小字列出了完整的演职员名单，包括“SCREENPLAY BY ANNA REID”、“CULINARY DIRECTION BY JAMES CARTER”以及Artisan Films、Riverstone Pictures和Heritage Media等众多出品公司标志。整体风格是写实主义，采用温暖柔和的灯光方案，营造出一种亲密的氛围。色调以棕色、米色和柔和的绿色等大地色系为主。两位演员的身体都在腰部被截断。}

    (Translation: A movie poster for the fictional English movie 'The Taste of Memory'. The scene is set in a rustic 19th century style kitchen. In the center of the screen, a middle-aged man with reddish brown hair and a small beard (played by actor Arthur Penhaligan) stands behind a wooden table. He is wearing a white shirt, black vest, and beige apron, looking at a woman holding a large piece of raw red meat with a wooden cutting board below. On his right, a black haired woman with a high bun (played by actress Eleanor Vance) leaned against the table and smiled gently at him. She was wearing a light colored shirt and a long skirt with white on top and blue on the bottom. On the table, in addition to a chopping board with chopped onions and shredded cabbage, there is also a white ceramic plate and fresh herbs. On the left side, there is a wooden box with a string of dark grapes. The background is a rough gray white plaster wall with a landscape painting hanging on it. On the far right countertop is a vintage oil lamp.
    There is a lot of textual information on the poster. The white sans serif font "ARTISAN FILMS PRESS" is located in the upper left corner, with "ELEANOR VANCE" and "ACADEMY AWARD" below it ®  WINNER”. In the upper right corner are written "ARTHUR PENHALIGON" and "GOLDEN GLOBE" ®  AWARD WINNER”. At the top center is the crown emblem of Sundance Film Festival, with the words' SUNDANCE FILM FESTIVAL GRAND JURY PRIZE 2024 'written below. The main title "THE TASTE OF Memory" is prominently displayed in large white serif font in the lower half. The title reads 'A FILM BY Tongyi Interaction Lab.'. The bottom area lists the complete cast and crew list in small white font, including "SCREENPLAY BY ANNA REID", "CULINARY Directing BY JAMES CARTER", as well as many production company logos such as Artisan Films, Riverstone Pictures, and Heritage Media. The overall style is realism, using warm and soft lighting schemes to create an intimate atmosphere. The color scheme is dominated by earthy tones such as brown, beige, and soft green. The bodies of both actors were severed at the waist.)
    
    \item Case \#2: 

    \zh{一张竖版日本艺术展海报，背景为深蓝色。设计以醒目的黄色文字和七幅水彩画拼贴为主。顶部是日文和英文标题。日文部分使用大号黄色宋体风格字体，内容为“長谷川 正季 - 水彩画の世界 -”。其下方是较小的黄色无衬线字体“-The world of watercolor-”。主标题“夢中天堂”以非常大的风格化黄色字体突出显示。其下是英文翻译“HEAVEN OF DREAM”，同样为黄色无衬线字体。再下一行是日文副标题“我が心の桂林”，字体较大，后跟其英文翻译“GUILIN IN MY MIND”，字体较小。海报中央是由七幅描绘桂林喀斯特地貌不同场景的水彩画组成的网格。这些画作展示了云雾缭绕的群山、蜿蜒穿过山谷的河流、倒映在水面上的绚丽日落、人们在船上提着灯笼的夜景以及其他富有氛围的风景。海报底部三分之一处用较小的黄色文字列出了活动详情，包括“2025.11.11(六)~17(五) 9:00~20:00”，“阿里巴巴云谷园区”，“(021)-34567890”。整体风格是优雅的平面设计，采用了深蓝色和黄色的高对比度双色调色板。}

    (Translation: \zh{A vertical Japanese art exhibition poster with a dark blue background. The design mainly features eye-catching yellow text and seven watercolor collages. At the top are Japanese and English titles. The Japanese section uses large yellow Song style fonts and the content is "長谷川 正季 - 水彩画の世界 -". Below it is a smaller yellow sans serif font that reads' The world of watercolor - '. The main title "夢中天堂" is highlighted in a very large stylized yellow font. Below is the English translation "HEAVEN OF DREAM", also in yellow sans serif font. The next line is the Japanese subtitle '我が心の桂林', with a larger font size, followed by its English translation 'GUILIN IN MY MIND', with a smaller font size. In the center of the poster is a grid composed of seven watercolor paintings depicting different scenes of Guilin's karst landscape. These paintings showcase misty mountains, winding rivers through valleys, stunning sunsets reflected on the water, night scenes of people carrying lanterns on boats, and other atmospheric landscapes. The activity details are listed in small yellow text at the bottom third of the poster, including "2025.11.11(六)~17(五) 9:00~20:00", "阿里巴巴云谷园区", and "(021) -34567890". The overall style is an elegant graphic design, featuring a high contrast dual tone palette of dark blue and yellow.})
    
    \item Case \#3: 

    \zh{一张方形构图的特写照片，主体是一片巨大的、鲜绿色的植物叶片，并叠加了文字，使其具有海报或杂志封面的外观。主要拍摄对象是一片厚实、有蜡质感的叶子，从左下角到右上角呈对角线弯曲穿过画面。其表面反光性很强，捕捉到一个明亮的直射光源，形成了一道突出的高光，亮面下显露出平行的精细叶脉。背景由其他深绿色的叶子组成，这些叶子轻微失焦，营造出浅景深效果，突出了前景的主叶片。整体风格是写实摄影，明亮的叶片与黑暗的阴影背景之间形成高对比度。图像上有多处渲染文字。左上角是白色的衬线字体文字“PIXEL-PEEPERS GUILD Presents”。右上角同样是白色衬线字体的文字“[Instant Noodle] 泡面调料包”。左侧垂直排列着标题“Render Distance: Max”，为白色衬线字体。左下角是五个硕大的白色宋体汉字“显卡在...燃烧”。右下角是较小的白色衬线字体文字“Leica Glow™ Unobtanium X-1”，其正上方是用白色宋体字书写的名字“蔡几”。}

    (Translation: \zh{A close-up photo with a square composition, featuring a large, bright green plant leaf and overlaid with text to give it the appearance of a poster or magazine cover. The main subject being photographed is a thick, waxy leaf that curves diagonally through the frame from the bottom left corner to the top right corner. Its surface has strong reflectivity, capturing a bright direct light source and forming a prominent highlight, revealing parallel fine leaf veins under the bright surface. The background is composed of other dark green leaves that are slightly out of focus, creating a shallow depth of field effect and highlighting the main leaf of the foreground. The overall style is realistic photography, with high contrast between bright leaves and dark shadow backgrounds. There are multiple rendered texts on the image. In the upper left corner is the white serif font text "PIXEL-PEEPERS GUIDE Gifts". The text in white serif font in the upper right corner reads '[Instant Noodle] 泡面调料包'. The title "Render Distance: Max" is vertically arranged on the left side in white serif font. In the bottom left corner are five large white Song typeface Chinese characters that read '显卡在...燃烧'. The smaller white serif font text "Leica Glow" is located in the bottom right corner ™  Unobtanium X-1”， Above it is the name "蔡几" written in white Song typeface.})
    
\end{itemize}

\textbf{Row \#3}
\begin{itemize}
    \item Case \#1: 

    A vertical digital illustration depicting a serene and majestic Chinese landscape, rendered in a style reminiscent of traditional Shanshui painting but with a modern, clean aesthetic.   The scene is dominated by towering, steep cliffs in various shades of blue and teal, which frame a central valley.   In the distance, layers of mountains fade into a light blue and white mist, creating a strong sense of atmospheric perspective and depth.   A calm, turquoise river flows through the center of the composition, with a small, traditional Chinese boat, possibly a sampan, navigating its waters.   The boat has a bright yellow canopy and a red hull, and it leaves a gentle wake behind it.   It carries several indistinct figures of people.   Sparse vegetation, including green trees and some bare-branched trees, clings to the rocky ledges and peaks.   The overall lighting is soft and diffused, casting a tranquil glow over the entire scene.   Centered in the image is overlaid text.   At the top of the text block is a small, red, circular seal-like logo containing stylized characters.   Below it, in a smaller, black, sans-serif font, are the words 'Zao-Xiang * East Beauty \& West Fashion * Z-Image'.   Directly beneath this, in a  larger, elegant black serif font, is the word 'SHOW \& SHARE CREATIVITY WITH THE WORLD'.  Among them, there are "SHOW \& SHARE", "CREATIVITY", and "WITH THE WORLD"

    \item Case \#2:

    \zh{vertical movie poster for the film "Come Back Home Often." created by Master of Oil painting.
    The artwork is a unified digital painting with a heavy impasto texture, mimicking thick oil paint strokes applied with a palette knife.  The central focus is a massive, abstract figure rendered in thick, textured white paint, resembling a giant bird or a stylized human form.  This white shape is set against a dark navy blue background that is densely covered with small, stylized flowers painted in vibrant red and white, with green stems.  In the bottom right corner, two elderly people are depicted from behind, walking away from the viewer.  One person, slightly ahead, wears a purple jacket and uses a wooden cane.  The other, slightly behind, wears a greyish-blue jacket.  Their bodies are truncated at the ankles by the bottom edge of the frame.  The overall style is surreal and symbolic, with a high-contrast color palette dominated by deep navy, white, and red.
    Text control: all lettering is fully integrated into the painted surface with identical heavy impasto, each character exhibiting raised ridges and knife-scraped edges that catch ambient light.
    In the top left corner, in white sans-serif strokes sculpted with thick, palette-knife ridges, the words “Z-Image” appear, and directly beneath them, still in raised impasto, “Visionary Creator.”
    In the bottom left, the Chinese title is rendered in large, white, cursive calligraphy (草书 style) built up from layered, knife-pressed paint: “常回家看看”, its down-strokes showing visible paint peaks.
    Below this, in a smaller white serif font whose letterforms are similarly embossed with raised impasto, reads the English title: “Come Back Home Often.”}
    
    \item Case \#3:
    
    \zh{传统中国水墨画照片，描绘了萧瑟秋日黄昏景象，位于页面左侧。画作竖向排列，用枯笔勾勒盘曲老藤缠绕古树，浓墨点染栖息昏鸦，淡墨晕染暮色天空。枯藤如龙蛇般攀附在虬曲树干上，三两只乌鸦停驻枝头，剪影般的轮廓。远处隐约可见小桥流水和古朴人家屋舍，炊烟袅袅。近景有一条黄土古道，瘦马低头缓行。画面上方有几行黑色行书书法，题写元曲名句，右上角钤一枚朱红方印。
    在画作右侧，有两列竖排中文文字，纯白色背景，草书字体，第一列写着 "天净沙·秋思"，第二列写着“枯藤老树昏鸦，小桥流水人家”。
    传统中式文人画风格，水墨单色系带飞白枯笔效果，笔触苍劲有力，构图疏密有致，留白处理突出，充满萧瑟苍凉的美学意蕴，氛围孤寂而悠远，具有浓厚的古典诗意和文化韵味。}

    (Translation: \zh{The traditional Chinese ink painting photo depicts a bleak autumn dusk scene, located on the left side of the page. The painting is arranged vertically, using a dry brush to outline the winding old vines around the ancient trees, with thick ink coloring the roosting crows, and light ink blending the twilight sky. The withered vine clung to the winding tree trunk like a dragon or snake, with three or two crows perched on the branches, forming a silhouette like silhouette. In the distance, small bridges, flowing water, and quaint houses can be faintly seen, with smoke rising from cooking. In the close-up, there is a loess ancient road, and thin horses are walking slowly with their heads down. There are several lines of black cursive calligraphy above the screen, inscribed with famous lines of Yuan opera, and a vermilion square seal is stamped in the upper right corner.
On the right side of the painting, there are two vertical columns of Chinese characters with a pure white background and cursive font. The first column reads "天净沙·秋思", and the second column reads "枯藤老树昏鸦，小桥流水人家".
The traditional Chinese literati painting style features a single color ink wash with a flying white and withered pen effect. The brushstrokes are vigorous and powerful, the composition is dense and orderly, and the white space treatment is prominent. It is full of desolate and desolate aesthetic connotations, creating a lonely and distant atmosphere with strong classical poetry and cultural charm.})
\end{itemize}

\textbf{Row \#4}
\begin{itemize}
    \item Case \#1: 

    \zh{竖排行书书法作品特写，以米白色带浅淡肌理的半生熟宣纸为载体，纸张带有自然轻微褶皱，质感温润柔和；黑色墨汁书写的行书字体，笔锋粗细变化灵动，如“人”字捺笔舒展劲挺、“志”字连笔婉转流畅，墨色层次丰富，部分笔画带自然飞白效果，尽显苍劲洒脱的书写张力；文字按传统从右至左竖列排布，可见“無人扶我青雲志”“我自踏雪至山巅”等词句，左侧配有“造相大师”落款小字，纸面点缀多枚朱红方形篆刻印章，印泥色泽饱满、印文线条清晰；多张书法纸呈轻微重叠的错落摆放，背景隐约露出其他纸张的淡色字迹，营造出随性的创作氛围；光线为柔和自然光，均匀铺洒在纸面，凸显墨色的光泽与纸张的纹理褶皱，造梦师的诗意视觉风格，整体氛围雅致古朴，兼具手写书法的灵动随性与传统文房的沉静质感。}

    (Translation: \zh{Close up of vertical cursive calligraphy works, using semi ripe rice paper with a light texture and off white color as the carrier. The paper has natural slight wrinkles and a warm and soft texture; The running script font written in black ink has dynamic changes in stroke thickness, such as the stretching and vigorous strokes of the "人" character and the smooth and graceful strokes of the "志" character. The ink layers are rich, and some strokes have a natural flying white effect, showcasing the vigorous and free spirited writing tension; The text is arranged vertically from right to left according to tradition, with phrases such as "無人扶我青雲志" and "我自踏雪至山巅" visible. On the left side, there is a small signature of "造相大师", and the paper is decorated with multiple vermilion square seal seals. The ink color is full and the lines of the seal are clear; Multiple calligraphy papers are arranged in a slightly overlapping and staggered manner, with the background faintly revealing the light colored handwriting of other papers, creating a casual creative atmosphere; The light is soft natural light, evenly spread on the paper surface, highlighting the luster of ink color and the texture wrinkles of the paper. The poetic visual style of the dream maker creates an elegant and rustic atmosphere, combining the agility and casualness of handwritten calligraphy with the calm texture of traditional study rooms.})

    \item Case \#2: 

    \zh{一张垂直构图的平面设计海报，背景是纯粹而鲜艳的宝蓝色。顶部的巨大无衬线字体主标题，上半部分为浅灰色的“Sofa Montain Slummerfest”，下半部分为白色的“Annual Napping Festival 2025”。其下方是巨大的黑色书法字体中文标题“沙发山打呼节”。海报的下半部分由一幅巨大的、插画风格的老虎插画占据，它正趴着面向观众，眼睛是黄色的。其皮毛由橙、黑、白三色构成。一个包含红色爱心的思想泡泡漂浮在它的头顶。海报上布满了详尽的活动文字。左栏用白色字体列出了以猫为主题的乐队名，如“The Fluffy Paws Grumbers (毛爪咕噜)”、“DJ Meow Mix”、“九命怪猫 (Nine Lives)”、“激光笔追逐者 (The Laser Dots)”、“纸箱爱好者 (Cardbock Box Lovers)”、“呼噜神教 (The Purr-fectionists)”、“猫草成瘾者 (The Catnip Junkies)”、“DJ Chairman Meow (猫主席)”以及像“Varh Radator Fesidenl Paw-Five”这样的无意义短语。右栏列出了活动细节，其中许多都带有滑稽的拼写错误或无意义内容，包括日期“4/1 MONDAY SUNL SUNSET”、地点“上海市浦东新区猫抓板路1号顶楼阳台”以及票务信息如“ADV. 1 CAN OF TUNA, DOOR 2 CANS, KITTENS FREE!”。在最底部是一排虚构的赞助商标志，名称包括“Catberd”、“好主人罐罐有限公司 (Good Oinar Canned Food Ltd)”和“iNONEPAWS”。}

    (Translation: \zh{A vertically composed graphic design poster with a pure and vibrant navy blue background. The main title features a large sans serif font at the top, with the upper half in light gray reading "Sofa Montain Slummerfest" and the lower half in white reading "Annual Napping Festival 2025". Below it is a huge black calligraphy font with the Chinese title "沙发山打呼节". The lower part of the poster is occupied by a huge, illustrated tiger illustration, which is lying face down to the audience with yellow eyes. Its fur is composed of three colors: orange, black, and white. A thought bubble containing a red heart floats above its head. The poster is filled with detailed activity text. The left column lists the names of cat themed bands in white font, such as "The Fluffy Paws Grumbers (毛爪咕噜)", "DJ Meow Mix”, "九命怪猫 (Nine Lives)", "激光笔追逐者 (The Laser Dots)", "纸箱爱好者 (Cardbock Box Lovers)", "呼噜神教 (The Purr-fectionists)", "猫草成瘾者 (The Catnip Junkies)", "DJ Chairman Meow (猫主席)" and meaningless phrases like "Varh Radator Feseidel Paw Five. The right column lists the details of the event, many of which have humorous spelling errors or meaningless content, including the date "4/1 MONDAY SUNL SUNSET", location "上海市浦东新区猫抓板路1号顶楼阳台", and ticketing information such as "ADV. 1 CAN OF TUNA, DOOR 2 CANS, KITTENS FREE. At the bottom is a row of fictional sponsor logos, with names including "Cattered," "好主人罐罐有限公司" and "iNONEPAWS".})
    
    \item Case \#3: 

    \zh{一张植物展览的平面设计海报，背景为素净的米白色。海报上有多幅水彩插画，描绘了各种苔藓和蕨类植物，用色以绿色、棕色和黄色为主，并配有精致的黑色墨水轮廓线。画面中央是一幅巨大而精细的绿色地钱插画，上面有棕色的孢子体。其他较小的插图散布在周围，包括左上角的细叶满江红（Fissidens），顶部中央的巨叶红茎藓（Rhodobryum giganteum），右上角的青苔属（Bryum sp.），以及右下角的凤尾藓（Marchantia formosana）。海报布局干净、简约，文字垂直和水平排列。右上角是纵向排列的黑色宋体大字“苔痕”，下边是横向排列的无衬线字体英文“Moss Exhibition”。左侧是黑色无衬线字体“Elkhorn Fern LifeStyle”。左下角写着“Alishan Moss Ecological”。日期和时间在底部突出显示：“2001”为黑色大号衬线字体，其后是较小的无衬线字体“04.22 [Apr.] am. 09:00”和“05.22 [May] pm. 17:00”。每幅植物插图都附有其学名，使用小号灰色无衬线字体书写，例如“Fissidens”、“Rhodobryum giganteum”、“Bryum sp.”、“Bartramiaceae”、“Alishan Moss Ecological”、“Marchantia formosana”和“Astrocella yoshinagana”。整体风格优雅且具有教育意义，将科学插画与现代排版相结合。}

    (Translation: \zh{A graphic design poster for a plant exhibition, with a plain beige background. The poster features multiple watercolor illustrations depicting various mosses and ferns, primarily using green, brown, and yellow colors, with delicate black ink outlines. In the center is a large and intricate illustration of green liverwort with brown sporophytes on top. Other smaller illustrations are scattered around, including Fissidens in the upper left corner, Rhodobryum giganteum in the top center, Bryum sp. in the upper right corner, and Marchantia formosana in the lower right corner. The poster layout is clean and minimalist, with text arranged both vertically and horizontally. In the upper right corner is the vertically arranged black Song-style characters "苔痕", below which is the horizontally arranged sans-serif English text "Moss Exhibition". On the left side is the black sans-serif text "Elkhorn Fern LifeStyle". The lower left corner reads "Alishan Moss Ecological". The date and time are prominently displayed at the bottom: "2001" in large black serif font, followed by smaller sans-serif text "04.22 [Apr.] am. 09:00" and "05.22 [May] pm. 17:00". Each plant illustration is accompanied by its scientific name, written in small gray sans-serif font, such as "Fissidens", "Rhodobryum giganteum", "Bryum sp.", "Bartramiaceae", "Alishan Moss Ecological", "Marchantia formosana", and "Astrocella yoshinagana". The overall style is elegant and educational, combining scientific illustration with modern typography.})

    \end{itemize}
    
\subsection{Figure 3}
\label{sec:fig_3}
\textbf{Row \#1}
\begin{itemize}
    \item Case \#1: 
    
    \zh{图2中女生的服饰换成图1的服饰，并背上图1的包。}
    
    (Translation: \zh{Replace the girl's outfit in Figure 2 with the outfit from Figure 1, and have her carry the bag from Figure 1 on her back.})
    \item Case \#2: 
    
    \zh{参考图1的风格，修改图2。}
    
    (Translation: \zh{Modify Figure 2 to match the style of Figure 1.})
    \item Case \#3: 
    
    \zh{一位蓝眼古风少女，头戴秋花发饰，颈戴绿粉珠链，温柔抱着一只穿粉蓝背带的白猫；身旁是怒吼的路飞，戴着红草帽和墨镜。背景：樱花与枫叶共舞，阳光温暖。风格：超精细奇幻写实，电影感光影。}
    
    (Translation: \zh{A gentle ancient-style girl with blue eyes, adorned with autumn flower hair accessories and a green-and-pink beaded necklace, tenderly holding a white cat dressed in pink-and-blue overalls. Beside her stands a roaring Luffy, wearing his iconic red straw hat and sunglasses. Background: Cherry blossoms and maple leaves dancing together under warm sunlight. Style: Ultra-fine fantasy realism with cinematic lighting.})
\end{itemize}

\textbf{Row \#2}
\begin{itemize}
    \item Case \#1: 
    
    \zh{给图中女生换一套古风修仙风格的衣服。}
    
    (Translation: \zh{Change the girl's outfit in the image to a set of ancient-style cultivation (Xianxia) attire.})
    \item Case \#2:
    
    \zh{让女生站起来。}
    
    (Translation: \zh{Make the girl stand up.})
    \item Case \#3: 
    
    \zh{修复街道中单人女性的逆光/背光问题，提亮面部和阴影区域，恢复面部细节和肤色。保持光影关系的自然过渡。}
    
    (Translation: \zh{Fix the backlighting issue for the lone woman in the street; brighten her face and shadow areas to restore facial details and skin tone, while maintaining natural transitions in the lighting.})
\end{itemize}

\textbf{Row \#3}
\begin{itemize}
    \item Case \#1: 
    
    \zh{狗替换成柴犬。}
    
    (Translation: \zh{Replace the dog with a Shiba Inu.})
    \item Case \#2: 
    
    \zh{一款以图中女生为主角的暖杏色高级感杂志封面风海报。她身穿黑色吊带裙，神态温婉，海报上排版有飘逸的白色手写字‘Z-Image Edit’和‘MAKE IT SURPRISE’等丰富的设计文案细节。}
    
    (Translation: \zh{A premium magazine cover-style poster in warm apricot tones, featuring the girl from the image as the protagonist. She wears a black slip dress with a gentle expression. The layout includes elegant, flowing white handwritten text reading "Z-Image Edit" and "MAKE IT SURPRISE," along with rich design copy details.})
\end{itemize}

\textbf{Row \#4}
\begin{itemize}
    \item Case \#1: 
    
    \zh{将“艺术家”修改为“文学家”，并保留原图字体。}
    
    (Translation: \zh{Change "artist" to "writer," while preserving the original font.})
    \item Case \#2: 
    
    \zh{用白色小狗替换粉色线条，用粉白条纹抱枕替换绿色线条。真实，自然。}
    
    (Translation: \zh{Replace the pink lines with a white puppy, and replace the green lines with a pink-and-white striped cushion. Make it look real and natural.})

\end{itemize}

\textbf{Row \#5}
\begin{itemize}
    \item Case \#1: 
    
    \zh{移除楼梯上的红色地毯，露出下方的大理石台阶。}
    
    (Translation: \zh{Change "artist" to "writer," while preserving the original font.})
    \item Case \#2: 
    
    \zh{调整白色浴缸内的水位，使水面升高至接近浴缸上边缘，呈现几乎满盈的状态。}
    
    (Translation: \zh{Adjust the water level in the white bathtub, raising the surface to near the top edge to create an almost full appearance.})
    \item Case \#3: 
    
    \zh{在地面和猫身上添加一些透过树叶洒下的斑驳光影。}
    
    (Translation: \zh{Add dappled light and shadows filtering through leaves onto the ground and the cat.})

\end{itemize}

\end{document}